\documentclass{article}

\usepackage{hyperref}

\usepackage{amsmath,amssymb,amsfonts}
\usepackage{graphicx}
\usepackage{graphics}
\usepackage{multirow} 
\usepackage{booktabs} 
\usepackage{hhline}
\usepackage{appendix}
\usepackage{caption}
\usepackage{subcaption}
\usepackage{csquotes}
\usepackage[british]{babel}
\usepackage{dblfloatfix}

\usepackage{enumitem}
\usepackage{fancyhdr}
\usepackage{xcolor}
\setlist[enumerate]{label*=\arabic*.}
\usepackage{mathtools}
\DeclarePairedDelimiterX{\infdivx}[2]{(}{)}{%
  #1\;\delimsize\|\;#2%
}

\usepackage[accepted]{icml2021}

\icmltitlerunning{CLOCS: Contrastive Learning of Cardiac Signals Across Space, Time, and Patients}

\begin{document}

\twocolumn[
\icmltitle{CLOCS: Contrastive Learning of Cardiac Signals Across Space, Time, and Patients}




\begin{icmlauthorlist}
\icmlauthor{Dani Kiyasseh}{ox}
\icmlauthor{Tingting Zhu}{ox}
\icmlauthor{David A. Clifton}{ox,ch}
\end{icmlauthorlist}

\icmlaffiliation{ox}{Department of Engineering Science, University of Oxford, Oxford, United Kingdom}
\icmlaffiliation{ch}{Oxford-Suzhou Centre for Advanced Research, Suzhou, China}

\icmlcorrespondingauthor{Dani Kiyasseh}{dani.kiyasseh@eng.ox.ac.uk}

\icmlkeywords{Contrastive Learning, Cardiac Signals}

\vskip 0.3in
]



\printAffiliationsAndNotice{}  

\begin{abstract}
The healthcare industry generates troves of unlabelled physiological data. This data can be exploited via contrastive learning, a self-supervised pre-training method that encourages representations of instances to be similar to one another. We propose a family of contrastive learning methods, CLOCS, that encourages representations across space, time, \textit{and} patients to be similar to one another. We show that CLOCS consistently outperforms the state-of-the-art methods, BYOL and SimCLR, when performing a linear evaluation of, and fine-tuning on, downstream tasks. We also show that CLOCS achieves strong generalization performance with only 25\% of labelled training data. Furthermore, our training procedure naturally generates patient-specific representations that can be used to quantify patient-similarity.
\end{abstract}

\section{Introduction}

At present, the healthcare system struggles to sufficiently leverage the abundant, unlabelled datasets that it generates on a daily basis. This is partially due to the dependence of deep learning algorithms on high quality labels for strong generalization performance. However, procuring such high quality labels in a clinical setting, where physicians are squeezed for time and attention, is practically infeasible. Self-supervised methods offer a way to overcome such an obstacle. For example, they can exploit unlabelled datasets to formulate pretext tasks such as predicting the rotation of images \citep{Gidaris2018}, their corresponding colourmap \citep{Larsson2017}, and the arrow of time \citep{Wei2018}. More recently, contrastive learning was introduced as a way to learn representations of instances that share some context. By capturing this high-level shared context (e.g., medical diagnosis), representations become invariant to the differences (e.g., input modalities) between the instances. 

Contrastive learning can be characterized by three main components: 1) a positive and negative set of examples, 2) a set of transformation operators, and 3) a variant of the noise contrastive estimation loss. Most research in this domain has focused on curating a positive set of examples by exploiting data temporality \citep{Oord2018}, data augmentations \citep{Chen2020}, and multiple views of the same data instance \citep{Tian2019}. These methods are predominantly catered to the image-domain and central to their implementation is the notion that shared context arises from the same instance. We believe this precludes their applicability to the medical domain where physiological time-series are plentiful. Moreover, their interpretation of shared context is limited to data from a common source, where that source is the individual data instance. In medicine, however, shared context can occur at a higher level, the patient level. This idea is central to our contributions and will encourage the development of representations that are patient-specific. Such representations have the potential to be used in tasks that exploit patient similarity such as disease subgroup clustering and discovery. As a result of the process, medical practitioners may receive more interpretable outputs from networks.  

In this work, we leverage electrocardiogram (ECG) signals to learn patient-specific representations via contrastive learning. In the process, we exploit both temporal and spatial information present in the ECG, with the latter referring to projections of the same electrical signal of the heart onto multiple axes, also known as leads.

\textbf{Contributions.} Our contributions are the following:
\begin{enumerate}[leftmargin=0.5cm]
    \item We propose a family of patient-specific contrastive learning methods, entitled CLOCS, that exploit both temporal and spatial information present in ECG signals.
    
    \item We show that CLOCS outperforms state-of-the-art methods, BYOL and SimCLR, when performing a linear evaluation of, and fine-tuning on, downstream tasks involving cardiac arrhythmia classification. 
\end{enumerate}

\section{Related Work}

\textbf{Contrastive learning.} In contrastive predictive coding, \citet{Oord2018} use representations of current segments to predict those of future segments. More recently, \citet{Tian2019} propose contrastive multi-view coding where multiple views of the same image are treated as \textquote{shared context}. \citet{He2019,Chen2020,Grill2020} exploit the idea of instance discrimination \citep{Wu2018} and interpret multiple views as stochastically augmented forms of the same instance. They explore the benefit of sequential data augmentations and show that cropping and colour distortions are the most important. These augmentations, however, do not trivially extend to the time-series domain. \citet{Shen2020} propose to create mixtures of images to smoothen the output distribution and thus prevent the model from being overly confident. Time Contrastive Learning \citep{Hyvarinen2016} performs contrastive learning over temporal segments in a signal and illustrate the relationship between their approach and ICA. In contrast to our work, they formulate their task as prediction of the segment index within a signal and perform limited experiments that do not exploit the noise contrastive estimation (NCE) loss. \citet{Bachman2019}  Time Contrastive Networks \citep{Sermanet2017} attempt to learn commonalities across views and differences across time. In contrast, our work focuses on identifying commonalities across \textit{both} spatial and temporal components of data. 

\textbf{Self-supervision for medical time-series.} \citet{Miotto2016} propose DeepPatient, a 3-layer stacked denoising autoencoder that attempts to learn a patient representation using electronic health record (EHR) data. Although performed on a large proprietary dataset, their approach is focused on EHRs and does not explore contrastive learning for physiological signals. \citet{Sarkar2020} use ECG signals and define pretext classification tasks in the context of affective computing. These include tasks such as temporal inversion, negation, and time-warping. Their work is limited to affective computing, does not explore contrastive learning, and does not exploit multi-lead data as we do. \citet{Lyu2018,Li2020} explore a sequence to sequence model to learn representations from EHR data in the eICU dataset. In the process, they minimize the reconstruction error of the input time-series. \citet{Cheng2020} explore contrastive learning for biosignals on small-scale datasets. In contrast, we develop a family of patient-specific contrastive learning methods and evaluate them on four distinct datasets. 


\section{Background}

\subsection{Contrastive Learning}
Let us assume the presence of a learner $f_{\theta}: x \in \mathbb{R}^{D} \xrightarrow{} h \in \mathbb{R}^{E}$, parameterized by $\theta$, which maps a $D$-dimensional input, $x$, to an $E$-dimensional representation, $h$. Further assume the presence of an unlabelled dataset, $X \in \mathbb{R}^{N \mathrm{x} D}$, where $N$ is the total number of instances. 

Each unlabelled instance, $x^{i} \in X$, is exposed to a set of transformations, $T_{A}$ and $T_{B}$, such that $x_{A}^{i} = T_{A}(x^{i})$ and $x_{B}^{i} = T_{B}(x^{i})$. Such transformations can consist of two different data augmentation procedures such as random cropping and flipping. These transformed instances now belong to an augmented dataset, $X' \in \mathbb{R}^{N \mathrm{x} D \mathrm{x} V}$, where $V$ is equal to the number of applied transformations. In contrastive learning, representations, $h_{A}^{i} = f_{\theta}(x_{A}^{i})$ and $h_{B}^{i} = f_{\theta}(x_{B}^{i})$, are said to share context. As a result of this shared context, these representations constitute a positive pair because (a) they are derived from the same original instance, $x^{i}$, and (b) the transformations applied to the original instance are class-preserving. Representations within a positive pair are encouraged to be similar to one another and dissimilar to representations of all other instances, $h_{A}^{j}, h_{B}^{j} \: \forall{j} \: j \neq i$. The similarity of these representations, $s(h_{A}^{i},h_{B}^{i})$, is quantified via a metric, $s$, such as cosine similarity. Encouraging a high degree of similarity between representations in the positive pair can result in representations that are invariant to different transformations of the same instance.



\section{Methods}


\subsection{Positive and Negative Pairs of Representations}
Representations that are derived from the same \textit{instance} are typically assumed to share context. This approach, however, fails to capture commonalities present across instances. In the medical domain, for example, multiple physiological recordings from the same patient may share context. It is important to note that if such recordings were collected over large time-scales (e.g., on the order of years) and in drastically different scenarios (e.g., at rest vs. during a stress test), then the shared context across these recordings is likely to diminish. This could be due to changing patient demographics and disease profiles. With the previous caveat in mind, we propose to leverage commonalities present in multiple physiological recordings by redefining a positive pair to refer to representations of transformed instances that belong to the same \textit{patient}. We outline how to arrive at these transformed instances next. 


\subsection{Transformation Operators}
\label{sec:transformation_operators}

\begin{figure*}[!t]
    \centering
    \begin{subfigure}{\textwidth}
    \centering
    \includegraphics[width=1\textwidth]{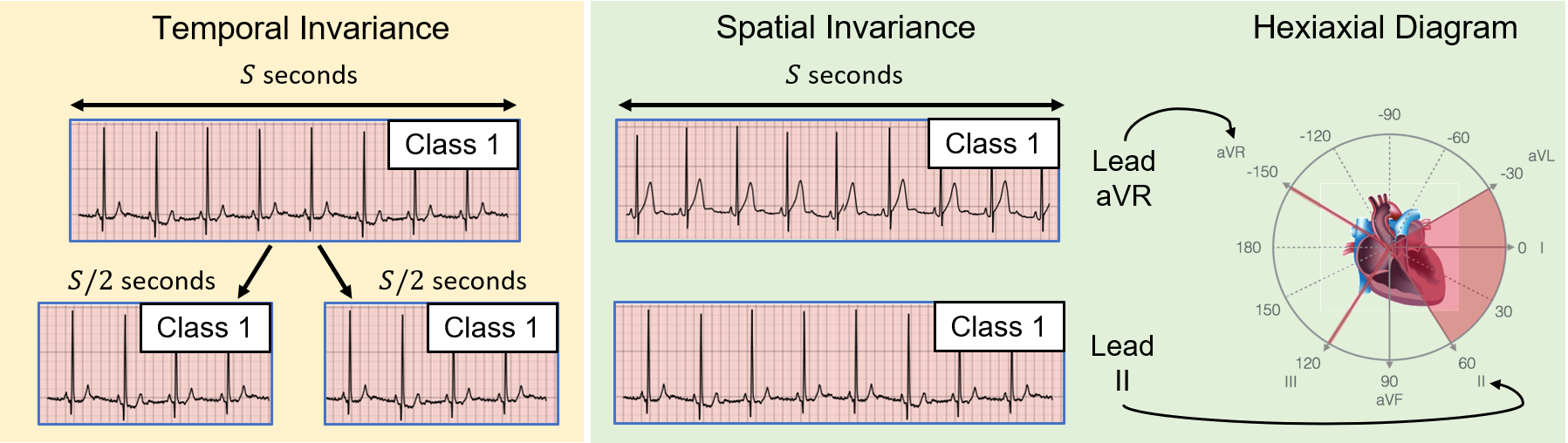}
    \end{subfigure}
    \caption{ECG recordings reflect both temporal and spatial information. This is because they measure the electrical activity of the heart using different leads (views) over time. \textbf{Temporal Invariance.} Abrupt changes to the ECG recording are unlikely to occur on the order of seconds, and therefore adjacent segments of shorter duration will continue to share context. \textbf{Spatial Invariance.} Recordings from different leads (at the same time) will reflect the same cardiac function, and thus share context.}
    \label{fig:invariances}
\end{figure*}

When choosing the transformation operators, $T$, that are applied to each instance, the principal desideratum is that they capture invariances in the ECG recording. Motivated by the observation that ECG recordings reflect both temporal and spatial information, we propose to exploit both temporal and spatial invariances. We provide an intuition for such invariances in Fig.~\ref{fig:invariances}. 

As is pertains to temporal invariance (Fig.~\ref{fig:invariances} left), we assume that upon splitting an ECG recording, associated with $\mathrm{Class} \ 1$, into sub-segments, each of the sub-segments remain associated with $\mathrm{Class} \ 1$. We justify this assumption based on human physiology where abrupt changes in cardiac function (on the order of seconds) are unlikely to occur. If these sub-segments were collected years apart, for example, our assumption may no longer hold. As for spatial invariance (Fig.~\ref{fig:invariances} right), we leverage the hexiaxial diagram which illustrates the location of the leads relative to the heart. We assume that temporally-aligned ECG recordings from different leads (views) are associated with the same class. This is based on the idea that multiple leads (collected at the same time) will reflect the same underlying cardiac function. Occasionally, this assumption may not hold, if, for example, a cardiac condition affects a specific part of the heart, making it detectable by only a few leads. We now describe how to exploit these invariances for contrastive learning. 


\textbf{Contrastive Multi-segment Coding (CMSC).} Given an ECG recording, $x^{i}$, with duration $S$ seconds, we can extract $V$ non-overlapping temporal segments, each with duration $S/V$ seconds. If $V=2$, for example, $x_{t1}^{i} = T_{t1}(x^{i})$ and $x_{t2}^{i} = T_{t2}(x^{i})$ where $t$ indicates the timestamp of the temporal segment (see Fig.~\ref{fig:invariances} left). We exploit temporal invariances in the ECG by defining representations of these adjacent and non-overlapping temporal segments as positive pairs. 

\textbf{Contrastive Multi-lead Coding (CMLC).} Different projections of the same electrical signal emanating from the heart are characterized by different leads, $L$. For example, with two leads, $L1$ and $L2$, then $x_{L1}^{i} = T_{L1}(x^{i})$ and $x_{L2}^{i} = T_{L2}(x^{i})$ (see Fig.~\ref{fig:invariances} right). We exploit spatial invariances in the ECG by defining representations of these different temporally-aligned projections as positive pairs. 


\textbf{Contrastive Multi-segment Multi-lead Coding (CMSMLC).} We simultaneously exploit both temporal and spatial invariances in the ECG by defining representations of non-overlapping temporal segments and different projections as positive pairs. For example, in the presence of two temporal segments with timestamps, $t1$ and $t2$, that belong to two leads, $L1$ and $L2$, then $x_{t1,L1}^{i} = T_{t1,L1}(x^{i})$ and $x_{t2,L2}^{i} = T_{t2,L2}(x^{i})$.


\begin{figure*}[!h]
\centering
\begin{subfigure}{0.8\textwidth}
	\centering
	\includegraphics[width=1\textwidth]{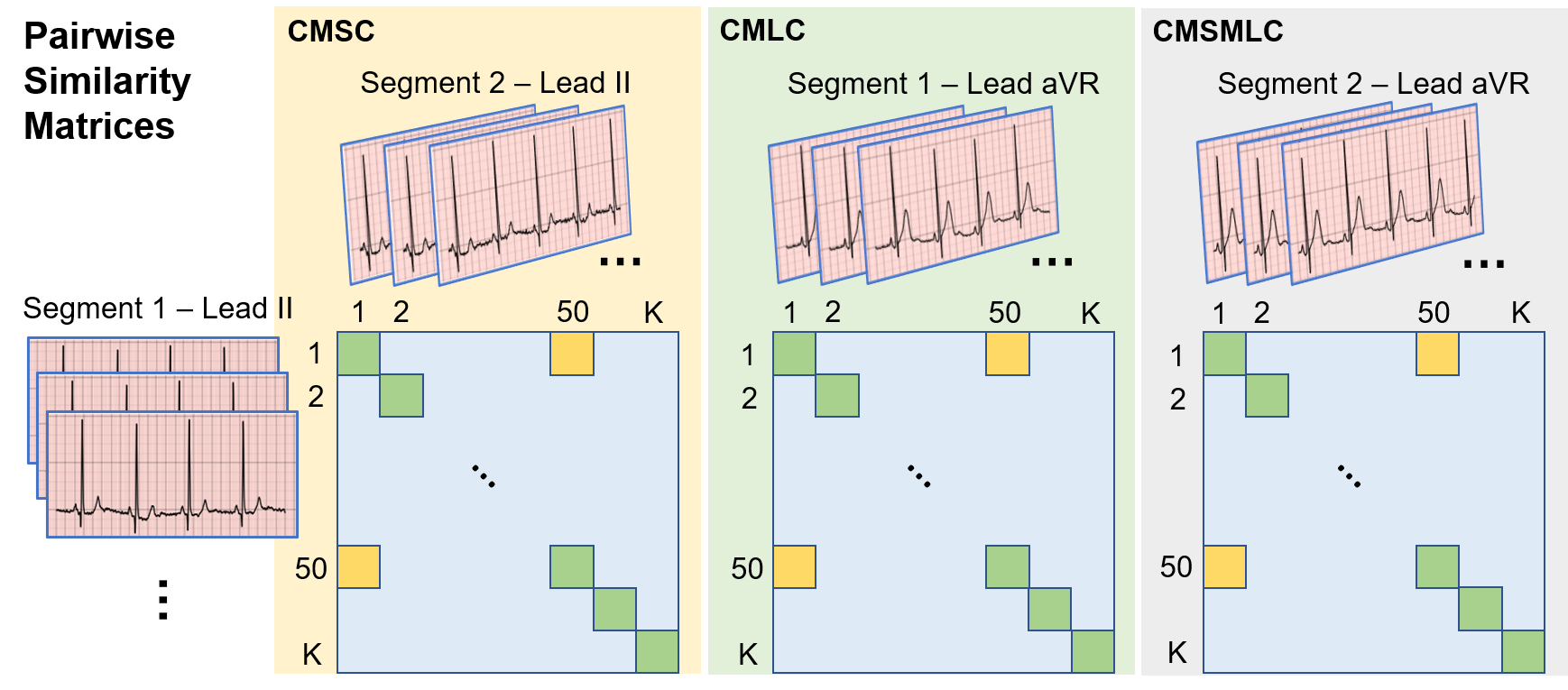}
\end{subfigure}
	\caption{Similarity matrix for a mini-batch of $K$ instances in (Left) \textbf{Contrastive Multi-segment Coding}, (Centre) \textbf{Contrastive Multi-lead Coding}, and (Right) \textbf{Contrastive Multi-segment Multi-lead Coding}. Additional matrices would be generated based on all pairs of applied transformation operators, $T_{A}$ and $T_{B}$. Exemplar transformed ECG instances are illustrated along the edges. To identify positive pairs, we associate each instance with its patient ID. By design, diagonal elements (green) correspond to the same patient, contributing to Eq.~\ref{eq:diagonal}. Similarly, instances 1 and 50 (yellow) belong to the same patient, contributing to Eq.~\ref{eq:off-diagonal}. The blue area corresponds to negative examples as they pertain to instances from different patients.}
	\label{fig:visualization_of_methods}
\end{figure*}

\subsection{Patient-Specific Noise Contrastive Estimation Loss}
Given our patient-centric definition of positive pairs, we propose to optimize a patient-specific noise contrastive estimation loss. More formally, Given a mini-batch of $K$ instances, we apply a pair of transformation operators and generate $2K$ transformed instances (a subset of which is shown in Fig.~\ref{fig:visualization_of_methods}). We encourage a pair of representations, $h_{A}^{i}$ and $h_{B}^{k}$, $i,k \in P$, from the same patient, $P$, to be similar to one another and dissimilar to representations from other patients. We quantify this similarity using the cosine similarity, $s$, with a temperature scaling parameter, $\tau$, (see Eq.~\ref{eq:cosine}) as is performed in \citep{Tian2019,Chen2020}. We extend this to all representations in the mini-batch to form a similarity matrix of dimension $K \times K$. In this matrix, we identify positive pairs by associating each instance with its patient ID. By design, this includes the diagonal elements and results in the loss shown in Eq.~\ref{eq:diagonal}. If the same patient reappears within the mini-batch, then we also consider off-diagonal elements, resulting in the loss shown in Eq.~\ref{eq:off-diagonal}. The frequency of these off-diagonals is inconsistent due to the random shuffling of data. We optimize the objective function in Eq.~\ref{eq:full_loss} for all pairwise combinations of transformation operators, $T_{A}$ and $T_{B}$, where we include Eq.~\ref{eq:diagonal} and Eq.~\ref{eq:off-diagonal} twice to consider negative pairs in both views. 
\begin{equation}
\label{eq:cosine}
    s(h_{A}^{i},h_{B}^{i}) = \frac{f_{\theta}(x_{A}^{i}) \cdot f_{\theta}(x_{B}^{i})}{ \| f_{\theta}(x_{A}^{i}) \| \| f_{\theta}(x_{B}^{i})\|} \frac{1}{\tau}
\end{equation}

\begin{equation}
\label{eq:diagonal}
    \mathcal{L}_{diag}^{h_{A},h_{B}} = -\mathbb{E}_{i \in P} \left[ \log \frac{e^{s(h_{A}^{i},h_{B}^{i})}}{\sum_{j} e^{s(h_{A}^{i},h_{B}^{j})}}\right]
\end{equation}

\begin{equation}
\label{eq:off-diagonal}
    \mathcal{L}_{off-diag}^{h_{A},h_{B}} = - \mathbb{E}_{i,k \in P}  \left[ \log\frac{e^{s(h_{A}^{i},h_{B}^{k})}}{\sum_{j} e^{s(h_{A}^{i},h_{B}^{j})}}\right]
\end{equation}

\begin{equation}
    \mathcal{L} = \mathbb{E}_{T_{A},T_{B}} \left[ \mathcal{L}_{diag}^{h_{A},h_{B}} + \mathcal{L}_{diag}^{h_{B},h_{A}} + \mathcal{L}_{off-diag}^{h_{A},h_{B}} + \mathcal{L}_{off-diag}^{h_{B},h_{A}} \right]
    \label{eq:full_loss}
\end{equation}

\section{Experimental Design}

\subsection{Datasets}

We conduct our experiments\footnote{Code can be accessed at: \url{https://github.com/danikiyasseh/CLOCS}} using PyTorch \citep{Paszke2019} on four ECG datasets that include cardiac arrhythmia labels. \textbf{PhysioNet 2020} \citep{PhysioNet2020} consists of 12-lead ECG recordings from 6,877 patients alongside 9 different classes of cardiac arrhythmia. Each recording can be associated with multiple labels. \textbf{Chapman} \citep{Zheng2020} consists of 12-lead ECG recordings from 10,646 patients alongside 11 different classes of cardiac arrhythmia. As is suggested by \cite{Zheng2020}, we group these labels into 4 major classes. \textbf{PhysioNet 2017} \citep{Clifford2017} consists of 8,528 single-lead ECG recordings alongside 4 different classes. \textbf{Cardiology} \citep{Hannun2019} consists of single-lead ECG recordings from 328 patients alongside 12 different classes of cardiac arrhythmia. An in-depth description of these datasets can be found in Appendix~\ref{appendix:data_description}.

All datasets were split into training, validation, and test sets according to patient ID using a 60, 20, 20 configuration. In other words, patients appeared in only one of the sets. The exact number of instances used during self-supervised pre-training and supervised training can be found in Appendix~\ref{appendix:instances}.  

\subsection{Pre-training Implementation}
We conduct our pre-training experiments on the training set of two of the four datasets: PhysioNet 2020 and Chapman. We chose these datasets as they contain multi-lead data. In \textbf{CMSC}, we extract a pair of non-overlapping temporal segments of $S=2500$ samples. This is equivalent to 10 and 5 seconds worth of ECG data from the Chapman and PhysioNet 2020 datasets, respectively. We choose this segment length so that the assumption of temporal invariance is more likely to hold and because segments $\leq10$ seconds in duration align with in-hospital ECG recording norms. Therefore, our model is presented with a mini-batch of dimension $K \times S \times 2$ where $K$ is the batchsize, and $S$ is the number of samples. In \textbf{CMLC}, we explore two scenarios with a different number of leads corresponding to the same instance. Our mini-batch dimension is $K \times S \times L$, where $L$ is the number of leads. Lastly, in \textbf{CMSMLC}, we incorporate an additional temporal segment in each mini-batch. Therefore, our mini-batch dimension is $K \times 2S \times L$. To ensure a fair comparison between all methods, we expose them to an equal number of patients and instances during training. In CMLC or CMSMLC, we pre-train using either 4 leads (II, V2, aVL, aVR) or all 12 leads. We chose these 4 leads as they cover a large range of axes.

\subsection{Evaluation on Downstream Task}
The downstream task of interest is that of cardiac arrhythmia classification; the diagnosis of abnormalities in the functioning of the heart. Such a procedure can be conducted in hospital and ambulatory settings, has widespread clinical applications from screening and guiding medical treatment to determining patient eligibility for surgery. We focus on this task due to its ubiquity and potential impact on a multitude of clinical workflows.

We evaluate our pre-trained methods in two scenarios. In \textbf{Linear Evaluation of Representations}, we are interested in evaluating the utility of the fixed feature extractor in learning representations. Therefore, the pre-trained parameters are frozen and multinomial logistic regression is performed on the downstream supervised task. In \textbf{Transfer Capabilities of Representations}, we are interested in evaluating the inductive bias introduced by pre-training. Therefore, the pre-trained parameters are used as an initialization for training on the downstream supervised task.

\subsection{Baselines}
We compare our pre-training methods to networks that are initialized randomly (\textbf{Random Init.}), via supervised pre-training (\textbf{Supervised}), or via a multi-task pre-training mechanism introduced specifically for ECG signals (\textbf{MT-SSL}) \citep{Sarkar2020}. We also compare to \textbf{BYOL} \citep{Grill2020} and \textbf{SimCLR} \citep{Chen2020}, which encourage representations of instances and their perturbed counterparts to be similar to one another, with the aim of learning transformation-invariant representations that transfer well. 

As SimCLR has been shown to be highly dependent on the choice of perturbations, we explore the following time-series perturbations (see Appendix~\ref{appendix:visualization_of_data_augmentations} for visualizations). (a) \textbf{Gaussian} \textendash we add $\epsilon \sim \mathcal{N}(0,\sigma)$ to the time-series signal where we chose $\sigma$ based on the amplitude of the signal. This was motivated by the work of \cite{Han2020} who recently showed the effect of additive noise on ECG signals. (b) \textbf{Flip} \textendash we flip the time-series signal temporally (\textbf{$\text{Flip}_{Y}$}), reversing the arrow of time, or we invert the time-series signal along the x-axis (\textbf{$\text{Flip}_{X}$}). (c) \textbf{SpecAugment} \textendash \citep{Park2019}  we take the short-time Fourier transform of the time-series signal, generating a spectrogram. We then mask either temporal (\textbf{$\text{SA}_{t}$}) or spectral (\textbf{$\text{SA}_{f}$}) bins of varying widths before converting the spectrogram to the time domain. We also explore the application of sequential perturbations to the time-series signal. 

\subsection{Hyperparameters}

During self-supervised pre-training, we chose the temperature parameter, $\tau=0.1$, as per \cite{Chen2020}. For BYOL, we chose the decay rate, $\tau_{d}=0.90$, after experimenting with various alternatives (see Appendix~\ref{appendix:effect_of_tau}). For all experiments, we use a neural architecture composed of three 1D convolutional layers followed by two fully connected layers. Further implementation details can be found in Appendix~\ref{appendix:implementation}. 

\section{Experimental Results}

\subsection{Effect of Perturbations on Performance}

Contrastive learning methods, and in particular SimCLR, are notorious for their over-dependence on the choice of perturbations. To begin exploring this dependence, we apply a diverse set of stochastic perturbations, $G$, (see Appendix~\ref{appendix:visualization_of_data_augmentations}) during pre-training and observe its effect on generalization performance. We follow the setup introduced by \cite{Chen2020} and apply either a \textbf{single perturbation} to each instance, $x^{i}$, whereby $x^{i}_{1} = G_{1}(x^{i})$, or \textbf{sequential perturbations} whereby $x^{i}_{1,2} = G_{2}(G_{1}(x^{i}))$. 

\begin{figure*}[t]
    \centering
    \includegraphics[width=0.8\textwidth]{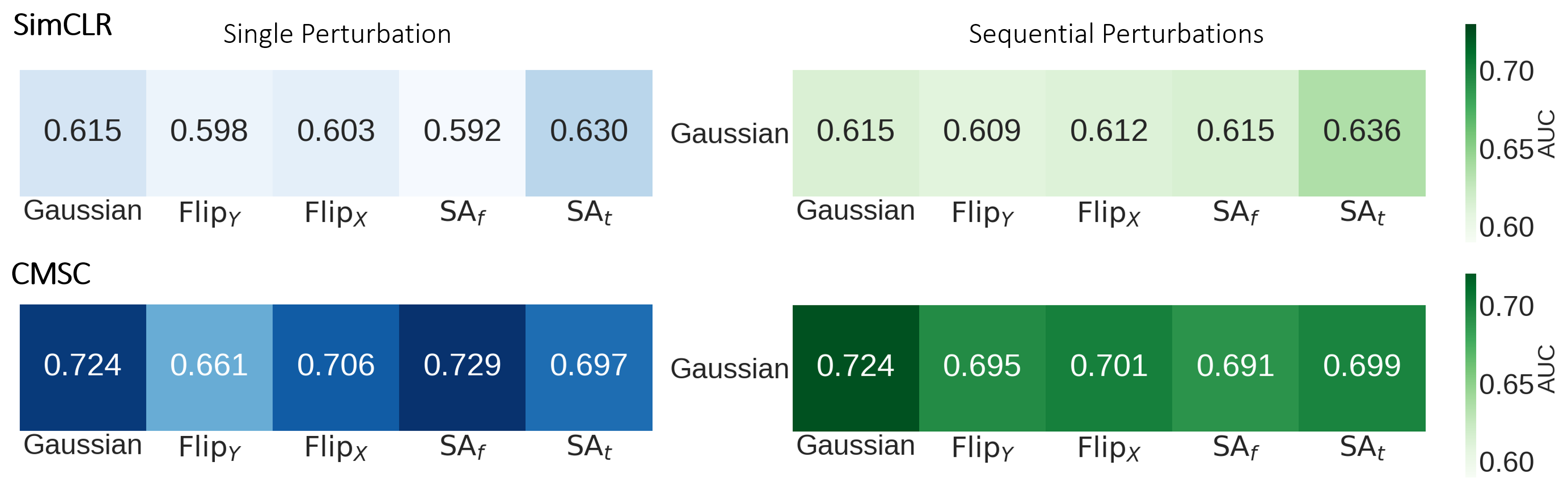}
    \caption{Effect of single (blue) and sequential (green) perturbations applied to the (top) SimCLR and (bottom) CMSC implementations on linear evaluation. Sequential perturbations involve a Gaussian perturbation followed by one of the remaining four types. Pre-training and evaluation was performed on PhysioNet 2020 using 4 leads. Evaluation was performed at $F=0.5$ and results are averaged across 5 seeds. We show that CMSC outperforms SimCLR regardless of the applied perturbation.}
    \label{fig:effect_of_perturbations}
\end{figure*}

We apply such perturbations while pre-training with SimCLR or CMSC on PhysioNet 2020 using 4 leads and, in Fig.~\ref{fig:effect_of_perturbations}, illustrate the test AUC in the linear evaluation scenario. We show that, regardless of the type and number of perturbations, CMSC continues to outperform SimCLR. For example, the \textit{worst-performing} CMSC implementation ($\text{Flip}_{Y}$) results in an $\text{AUC}=0.661$ which is still greater than the \textit{best-performing} SimCLR implementation ($\text{Gaussian} \rightarrow \text{SA}_{t}$) with an $\text{AUC}=0.636$. In fact, we find that pre-training with CMSC \textit{without} applying any perturbations (next section; see Table~\ref{table:linear_evaluation_main}) still outperforms the best-performing SimCLR implementation. Such a finding suggests that CMSC's already strong performance is more likely to stem from its redefinition of the \textquote{shared context} to include both time and patients than from the choice of perturbations. 

\subsection{Linear Evaluation of Representations}
In this section, we evaluate the utility of the self-supervised representations learned using four leads on a downstream linear classification task. In Table~\ref{table:linear_evaluation_main}, we show the test AUC on Chapman and PhysioNet 2020 using 50\% of the labelled data ($F=0.5$) after having learned representations, with dimension $E=128$, using the same two datasets.

\begin{table}[!h]
\footnotesize
\centering
\caption{Test AUC of the linear evaluation of the representations at $F=0.5$, after having pre-trained on Chapman or PhysioNet 2020 with $E=128$. Pre-training and evaluating multi-lead datasets* using 4 leads (II, V2, aVL, aVR). Mean and standard deviation are shown across 5 seeds. Bold reflects the top-performing method.}
\vskip 0.1in 
\label{table:linear_evaluation_main}
\begin{tabular}{c | c c }
\toprule
Dataset&\multicolumn{1}{c}{Chapman*}&\multicolumn{1}{c}{PhysioNet 2020*}\\
\midrule
MT-SSL & 0.677 $\pm$ 0.024 & 0.665 $\pm$ 0.015\\
BYOL &0.643 $\pm$ 0.043 & 0.595 $\pm$ 0.018 \\
SimCLR & \multicolumn{1}{c}{0.738 $\pm$ 0.034} & \multicolumn{1}{c}{0.615 $\pm$ 0.014} \\
CMSC &\multicolumn{1}{c}{\textbf{0.896 $\pm$ 0.005}} & \multicolumn{1}{c}{\textbf{0.715 $\pm$ 0.033}} \\
CMLC &\multicolumn{1}{c}{0.870 $\pm$ 0.022} & \multicolumn{1}{c}{0.596 $\pm$ 0.008} \\
CMSMLC &\multicolumn{1}{c}{0.847 $\pm$ 0.024} & \multicolumn{1}{c}{0.680 $\pm$ 0.008} \\
\bottomrule 
\end{tabular}
\end{table}

We show that CMSC outperforms BYOL and SimCLR on both datasets. On the Chapman dataset, CMSC and SimCLR achieve an $\mathrm{AUC}=0.896$ and $0.738$, respectively, illustrating a $15.8\%$ improvement. Such a finding implies that the representations learned by CMSC are richer and thus allow for improved generalization. We hypothesize that this is due to the setup of CMSC whereby the shared context is across segments (temporally) and patients. Moreover, we show that CLOCS (all 3 proposed methods) outperforms SimCLR in 100\% of all conducted experiments, even when pre-training and evaluating with all 12 leads (see Appendix~\ref{appendix:linear_evaluation}).   


\begin{table*}[!t]
\centering
\caption{Test AUC in the fine-tuning scenario at $F=0.5$, after having pre-trained on Chapman or PhysioNet 2020 with $E=128$. Pre-training, fine-tuning, and evaluating multi-lead datasets* using 4 leads. Mean and standard deviation are shown across 5 seeds. We show that, depending on the downstream dataset, either CMSC or CMSMLC outperform BYOL and SimCLR.}
\label{table:transfer_evaluation_main}
\vskip 0.1in 
\resizebox{\linewidth}{!}{%
\begin{tabular}{c | c c c | c c c }
\toprule
\multirow{1}{*}{Pre-training Dataset}&\multicolumn{3}{c}{Chapman*}&\multicolumn{3}{c}{PhysioNet 2020*}\\
\midrule
Downstream Dataset& Cardiology & PhysioNet 2017 & PhysioNet 2020* &Cardiology & PhysioNet 2017 & Chapman*\\
\midrule
Random Init. &0.678 $\pm$ 0.011 & 0.763 $\pm$ 0.005 & 0.803 $\pm$ 0.008 & 0.678 $\pm$ 0.011 & 0.763 $\pm$ 0.005 & 0.907 $\pm$ 0.006 \\
Supervised &0.684 $\pm$ 0.015 & 0.799 $\pm$ 0.008 & 0.827 $\pm$ 0.001 & 0.730 $\pm$ 0.002 & 0.810 $\pm$ 0.009 & 0.954 $\pm$ 0.003 \\
\midrule
\multicolumn{7}{l}{\textit{Self-supervised Pre-training}} \\
\midrule
MT-SSL & 0.650 $\pm$ 0.009 & 0.741 $\pm$ 0.012 & 0.774 $\pm$ 0.010 & 0.661 $\pm$ 0.011 & 0.746 $\pm$ 0.016 & 0.923 $\pm$ 0.007\\
BYOL &0.678 $\pm$ 0.021 & 0.748 $\pm$ 0.014 & 0.802 $\pm$ 0.013 & 0.674 $\pm$ 0.022 & 0.757 $\pm$ 0.010 & 0.916 $\pm$ 0.009 \\
SimCLR &0.676 $\pm$ 0.011 & 0.772 $\pm$ 0.010 & 0.823 $\pm$ 0.011 & 0.658 $\pm$ 0.027 & 0.762 $\pm$ 0.009 & 0.923 $\pm$ 0.010 \\
CMSC &0.695 $\pm$ 0.024 & 0.773 $\pm$ 0.013 & \textbf{0.830 $\pm$ 0.002} & \textbf{0.714 $\pm$ 0.014} & 0.760 $\pm$ 0.013 & \textbf{0.932 $\pm$ 0.008} \\
CMLC &0.665 $\pm$ 0.016 & 0.767 $\pm$ 0.013 & 0.810 $\pm$ 0.011 & 0.675 $\pm$ 0.013 & 0.762 $\pm$ 0.007 & 0.910 $\pm$ 0.012 \\
CMSMLC &\textbf{0.717 $\pm$ 0.006} & \textbf{0.774 $\pm$ 0.004} & 0.814 $\pm$ 0.009 & 0.698 $\pm$ 0.011 & \textbf{0.774 $\pm$ 0.012} & 0.930 $\pm$ 0.012 \\
\bottomrule 
\end{tabular}}
\end{table*}

\subsection{Transfer Capabilities of Representations}
In this section, we evaluate the utility of initializing a network for a downstream task with parameters learned via self-supervision using four leads. In Table~\ref{table:transfer_evaluation_main}, we show the test AUC on downstream datasets at $F=0.5$ for the various self-supervised methods with $E=128$.

We show that, with a few exceptions, self-supervision is advantageous relative to a Random Initialization. This can be seen by the higher AUC achieved by the former relative to the latter. We also show that, depending on the downstream dataset, either CMSC or CMSMLC outperform BYOL and SimCLR. For example, when pre-training on Chapman and fine-tuning on Cardiology, CMSMLC achieves an $\mathrm{AUC}=0.717$, a $4.1\%$ improvement compared to SimCLR. This implies that by encouraging representations across space, time, and patients to be similar to one another, networks are nudged into a favourable parameter space. 

We also find that CMLC performs consistently worse than CMSC and CMSMLC. We hypothesize that this is because the underlying physiological phenomenon, although the same at a particular time-point, may not manifest equivalently across the leads. For example, an arrhythmia in the right ventricle of the heart may be more apparent in Lead V1 than in Lead I. Therefore, attracting representations of such leads, as is done with CMLC, may confuse the network. In Appendix~\ref{sec:finetune_4_leads}, we extend these findings and illustrate that CLOCS outperforms SimCLR in at least $75\%$ of all experiments conducted, on average. When pre-training, fine-tuning, and evaluating using all 12 leads, we show that CMSC outperforms all other methods in at least $90\%$ of all experiments conducted (see Appendix~\ref{sec:finetune_12_leads}).

\subsection{Doing More With Less Labelled Data} 
\begin{figure}[!h]
\centering
\begin{subfigure}{0.9\columnwidth}
	\centering
	\includegraphics[width=\columnwidth]{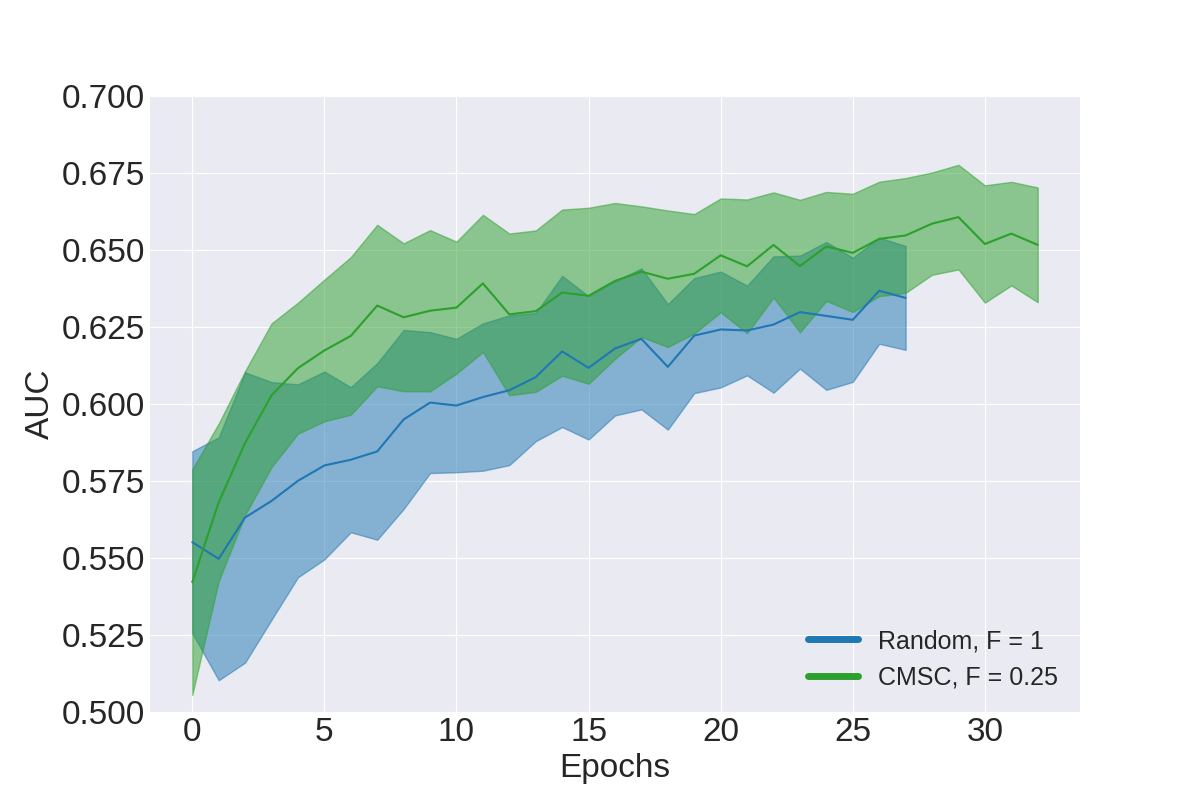}
	\caption{PhysioNet 2020 $\xrightarrow{}$ Cardiology}
	\label{fig:more_with_less_1}
\end{subfigure}
\begin{subfigure}{0.9\columnwidth}
	\centering
	\includegraphics[width=\columnwidth]{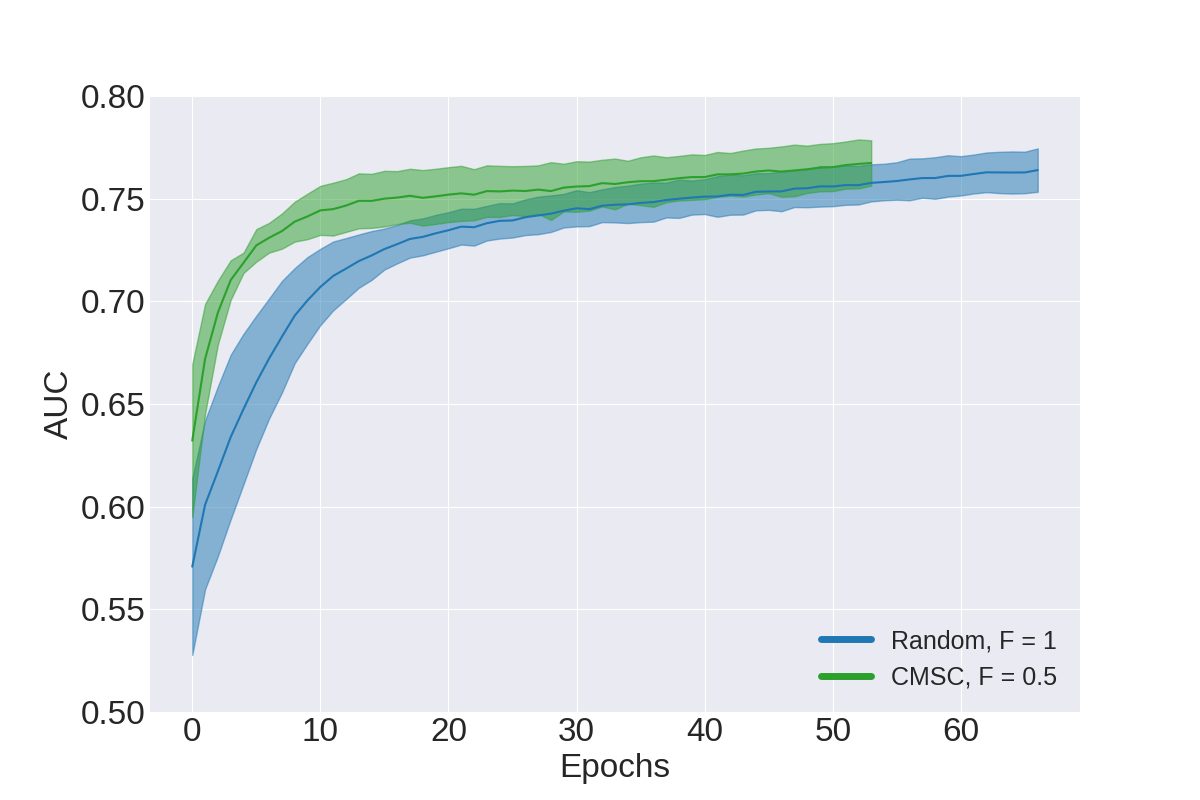}
	\caption{Chapman $\xrightarrow{}$ PhysioNet 2017}
	\label{fig:more_with_less_2}
\end{subfigure}
	\caption{Validation AUC of a network initialized randomly or via CMSC and which is exposed to different amounts of labelled training data, $F$. Results are averaged across 5 seeds. Shaded area represents one standard deviation. We show that a network initialized with CMSC and exposed to less data ($F<1$) outperforms one randomly initialized and exposed to all data ($F=1$).}
	\label{fig:more_with_less}
\end{figure}

Having established that self-supervision can nudge networks to a favourable parameter space, we set out to investigate whether such a space can lead to strong generalization with less labelled data in the downstream task. In Fig.~\ref{fig:more_with_less}, we illustrate the validation AUC of networks initialized randomly, with access to 100\% of the labels, or via CMSC, with access to fewer labels, and fine-tuned on two different datasets. 

We find that fine-tuning a network based on a CMSC initialization drastically improves data-efficiency. In Fig.~\ref{fig:more_with_less_1}, we show that a network initialized with CMSC and exposed to only $25\%$ of the labelled data outperforms one that is initialized randomly and exposed to $100\%$ of the labelled data. This can be seen by the consistently higher AUC during, and at the end of, training. A similar outcome can be seen in Fig.~\ref{fig:more_with_less_2}. This suggests that self-supervised pre-training exploits data efficiently such that networks can do more with less on downstream classification tasks.

\subsection{Effect of Embedding Dimension, $E$, and Availability of Labelled Data, $F$}

\begin{figure}[!h]
\centering
\begin{subfigure}{0.9\columnwidth} 
	\centering
	\includegraphics[width=\columnwidth]{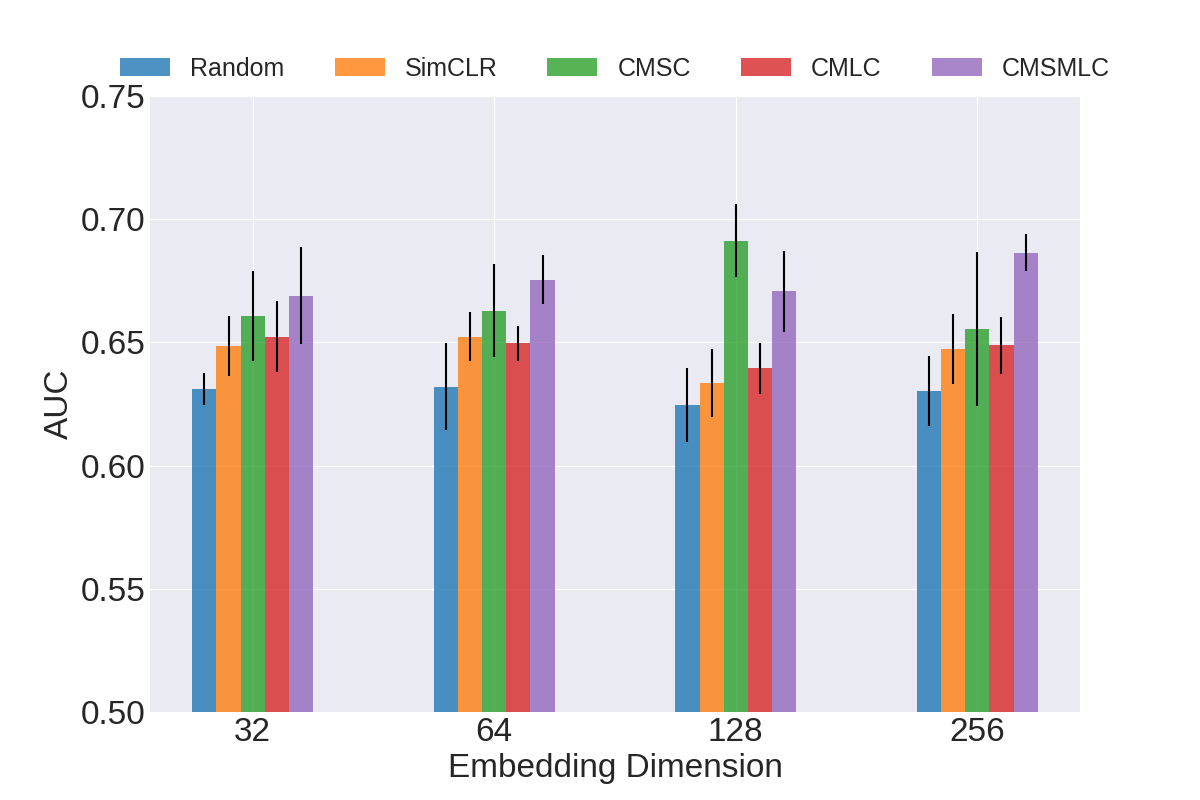}
	\caption{Chapman $\xrightarrow{}$ Cardiology, $F=0.25$}
	\label{fig:effect_of_ED}
\end{subfigure}
\begin{subfigure}{0.9\columnwidth}
	\centering
	\includegraphics[width=\columnwidth]{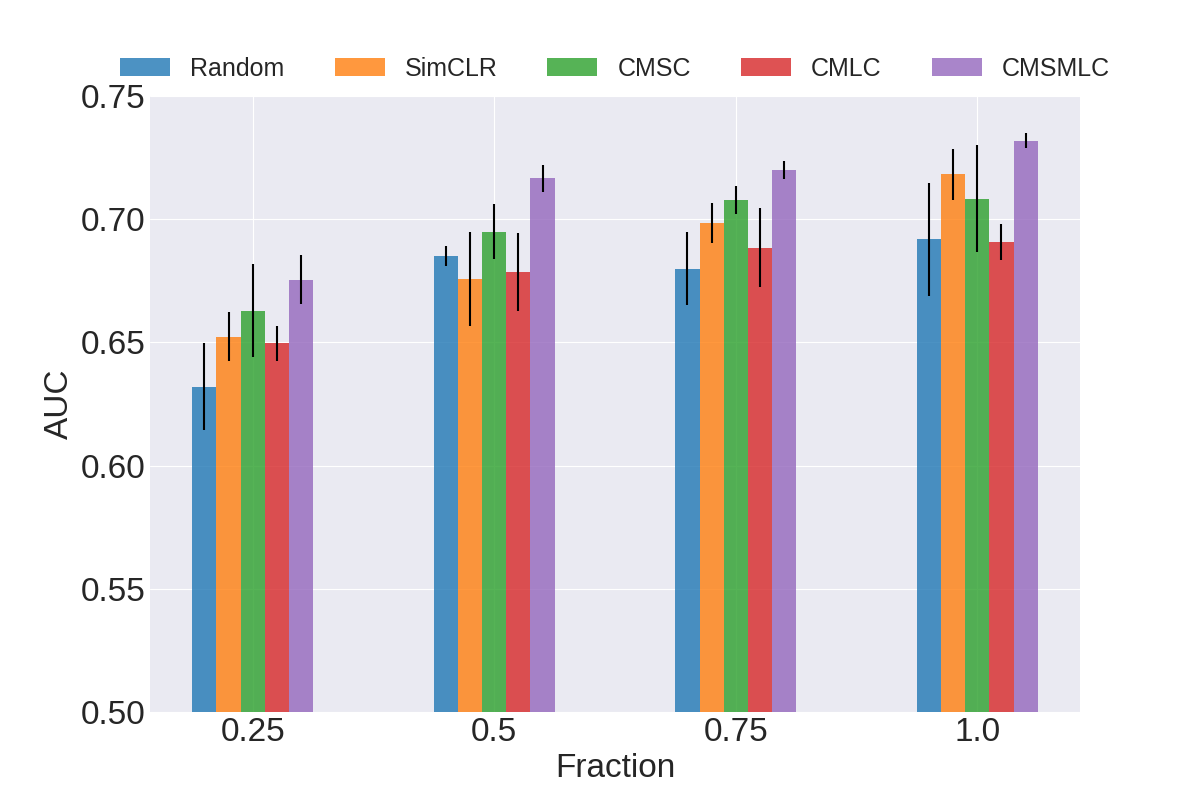}
	\caption{Chapman $\xrightarrow{}$ Cardiology, $E=64$}
	\label{fig:effect_of_LF}
\end{subfigure}
	\caption{Effect of (a) embedding dimension, $E$, and (b) labelled fraction, $F$, on the test AUC when pre-training on Chapman and fine-tuning on Cardiology. Results are averaged across 5 seeds. Error bars represent one standard deviation.}
	\label{fig:parameter_characterizations}
\end{figure}

The dimension of the representation learned during self-supervision and the availability of labelled training data can both have an effect on model performance. In this section, we investigate these claims. In Figs.~\ref{fig:effect_of_ED} and \ref{fig:effect_of_LF}, we illustrate the test AUC for all pre-training methods as a function of $E=(32,64,128,256)$ and $F=(0.25,0.50,0.75,1)$. 

In Fig.~\ref{fig:effect_of_ED}, we show that networks initialized randomly or via SimCLR are not significantly affected by the embedding dimension. This can be seen by the $\mathrm{AUC} \approx 0.63$ and $\approx 0.65$, for these two methods across all values of $E$. In contrast, the embedding dimension has a greater effect on CMSC where $\mathrm{AUC} \approx 0.66 \xrightarrow{} 0.69$ as $E = 32 \xrightarrow{} 128$. This implies that CMSC is still capable of achieving strong generalization performance despite the presence of few labelled data ($F=0.25$). We hypothesize that the strong performance of CMSC, particularly at $E=128$, is driven by its learning of patient-specific representations (see Appendix~\ref{appendix:patient_distances}) that cluster tightly around one another, a positive characteristic especially when such representations are class-discriminative. 

In Fig.~\ref{fig:effect_of_LF}, we show that increasing the amount of labelled training data benefits the generalization performance of all methods. This can be seen by the increasing AUC values as $F=0.25 \xrightarrow{} 1$. We also show that at all fraction values, CMSMLC outperforms its counterparts. For example, at $F=1$, CMSMLC achieves an $\mathrm{AUC}=0.732$ whereas SimCLR achieves an $\mathrm{AUC}=0.718$. Such behaviour still holds at $F=0.25$ where the two methods achieve an $\mathrm{AUC}=0.675$ and $0.652$, respectively. This outcome emphasizes the robustness of CMSMLC to scarce, labelled training data. 

\subsection{CLOCS Learns Patient-Specific Representations}
\label{sec:patient_specific_reps}

\begin{figure}[!t]
\centering
\begin{subfigure}{0.9\columnwidth}
	\centering
	\includegraphics[width=\columnwidth]{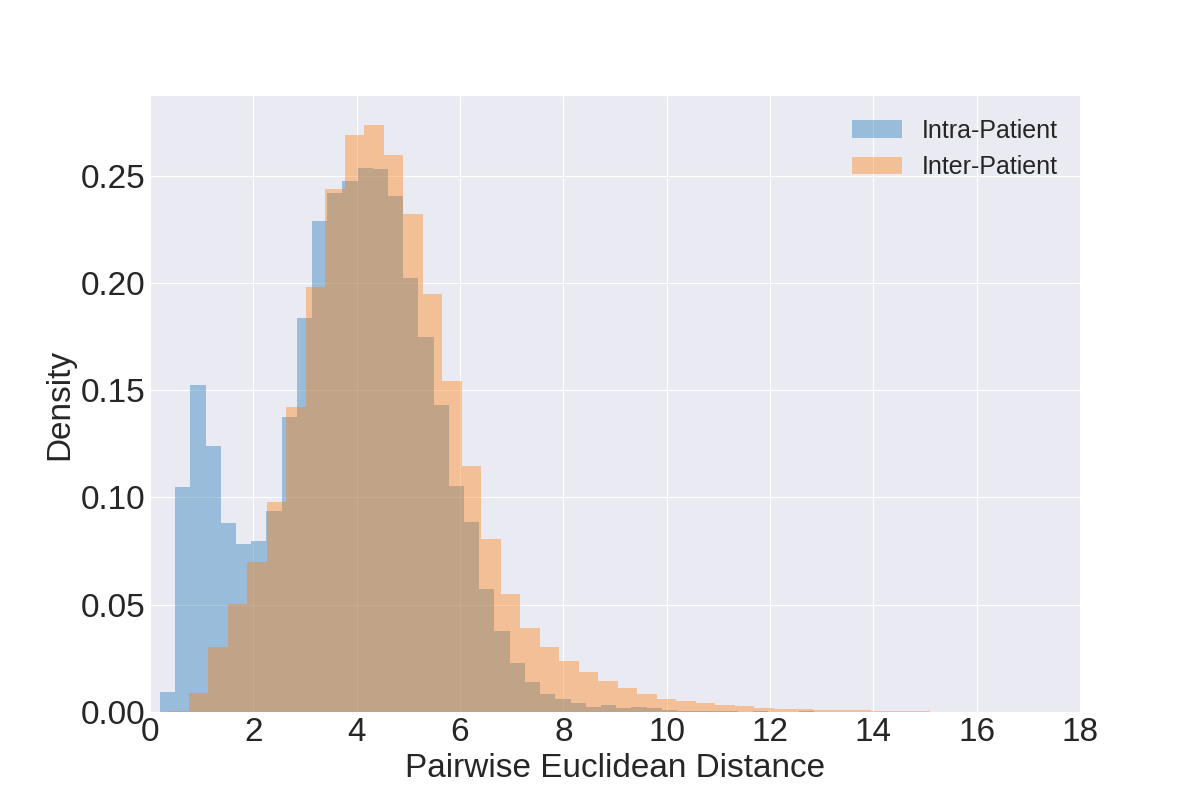}
	\caption{SimCLR}
	\label{fig:patient_distances_SimCLR_128}
\end{subfigure}
\begin{subfigure}{0.9\columnwidth}
	\centering
	\includegraphics[width=\columnwidth]{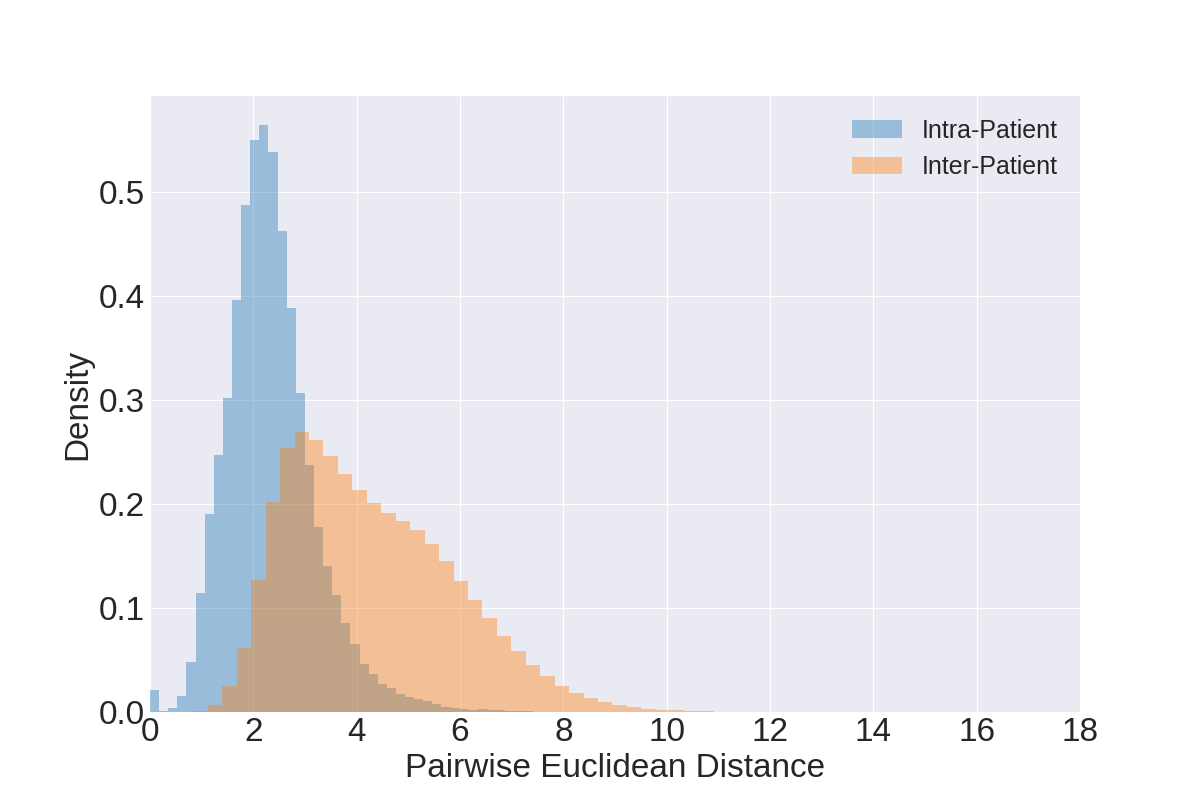}
	\caption{CMSC}
	\label{fig:patient_distances_CMSC_128}
\end{subfigure}
	\caption{Distribution of pairwise Euclidean distance between representations ($E=128$) belonging to the same patient (Intra-Patient) and those belonging to different patients (Inter-Patient). Self-supervision was performed on PhysioNet 2020. Notice the lower average intra-patient distance and improved separability between the two distributions with CMSC than with SimCLR.}
	\label{fig:patient_distances_main}
\end{figure}

We redefined \textquote{shared context} to refer to representations from the same patient, which in turn should produce patient-specific representations. To validate this hypothesis, we calculate the pairwise Euclidean distance between representations of the same patient (Intra-Patient) and those of different patients (Inter-Patient). On average, the former should be smaller than the latter. In Fig.~\ref{fig:patient_distances_main}, we illustrate the two distributions associated with the intra and inter-patient distances at $E=128$. At higher embedding dimensions, we find that these distributions are simply shifted to higher values (see Appendix~\ref{appendix:patient_distances}).

We show that these two distributions have large mean values and overlap significantly when implementing SimCLR, as seen in Fig.~\ref{fig:patient_distances_SimCLR_128}. This is expected as SimCLR is blind to the notion of a patient. In contrast, when implementing CMSC, the intra-patient distances are lower than those found in SimCLR, as seen in Fig.~\ref{fig:patient_distances_CMSC_128}. Moreover, the intra and inter-patient distributions are more separable. This implies that pre-training with CMSC leads to patient-specific representations. We note that this phenomenon takes place while concomitantly learning better representations, as observed in previous sections. 

\section{Discussion and Future Work}
In this paper, we proposed a family of self-supervised pre-training mechanisms, entitled CLOCS, based on contrastive learning for physiological signals. In the process, we encouraged representations across segments (temporally) and leads (spatially) that correspond to instances from the same patient to be similar to one another. We showed that our methods outperform the state-of-the-art methods, BYOL and SimCLR, when performing a linear evaluation of, and fine-tuning on, downstream tasks. This conclusion also holds when applying a range of perturbations and when pre-training and evaluating with a different number of leads. We now elucidate several avenues worth exploring.

\textbf{Quantifying patient similarity.} We have managed to learn patient-specific representations. These representations can be used to quantify patient-similarity in order to assist with diagnosis or gain a better understanding of a medical condition. Validation of these representations can be performed by comparing patients known to be similar. 

\textbf{Multi-modal transfer.} We transferred parameters from one task to another that shared the same input modality, the ECG. Such data may not always be available for self-supervision. An interesting path would be to explore whether contrastive self-supervision on one modality can transfer well to another modality. 

\textbf{Multi-modal contrastive learning.} We exploited the temporal and spatial invariance of a single modality, the ECG, for contrastive learning. However, we also envision CLOCS being applied to other modalities (e.g., cardiac ultrasound, brain MRI, and ICU vitals data). In the last case, for example, oxygen saturation and respiratory rate recorded over time can be interpreted as two spatial \enquote{views} of the same physiological phenomenon and can thus be exploited by CMLC. Furthermore, in a hospital setting, ECG signals are typically recorded alongside other modalities, such as the photoplethysmogram (PPG). One potential multi-modal approach would be to attract representations of temporally-aligned ECG and PPG segments.

\section*{Acknowledgements}
We thank the anonymous reviewers for their insightful feedback. We also thank Fairuz and Asmahan for lending us their voice. David Clifton was supported by the EPSRC under Grants EP/P009824/1and EP/N020774/1, and by the National Institute for Health Research (NIHR) Oxford Biomedical Research Centre (BRC). The views expressed are those of the authors and not necessarily those of the NHS, the NIHR or the Department of Health. Tingting Zhu was supported by the Engineering for Development Research Fellowship provided by the Royal Academy of Engineering. 

\bibliography{example_paper}
\bibliographystyle{icml2021}

\clearpage

\appendix

\onecolumn
\begin{subappendices}

\renewcommand{\thesubsection}{\Alph{section}.\arabic{subsection}}
\section{Datasets}
\label{appendix:datasets}

\subsection{Data Preprocessing}
\label{appendix:data_description}

For all of the datasets, frames consisted of 2500 samples and consecutive frames had no overlap with one another. Data splits were always performed at the patient-level.

\textbf{PhysioNet 2020} \citep{PhysioNet2020}. Each ECG recording varied in duration from 6 seconds to 60 seconds with a sampling rate of 500Hz. Each ECG frame in our setup consisted of 2500 samples (5 seconds). We assign multiple labels to each ECG recording as provided by the original authors. These labels are: AF, I-AVB, LBBB, Normal, PAC, PVC, RBBB, STD, and STE. The ECG frames were normalized in amplitude between the values of 0 and 1.

\textbf{Chapman} \citep{Zheng2020}. Each ECG recording was originally 10 seconds with a sampling rate of 500Hz. We downsample the recording to 250Hz and therefore each ECG frame in our setup consisted of 2500 samples. We follow the labelling setup suggested by \cite{Zheng2020} which resulted in four classes: Atrial Fibrillation, GSVT, Sudden Bradychardia, Sinus Rhythm. The ECG frames were normalized in amplitude between the values of 0 and 1. 

\textbf{Cardiology} \citep{Hannun2019}. Each ECG recording was originally 30 seconds with a sampling rate of 200Hz. Each ECG frame in our setup consisted of 256 samples resampled to 2500 samples. Labels made by a group of physicians were used to assign classes to each ECG frame depending on whether that label coincided in time with the ECG frame. These labels are: AFIB, AVB, BIGEMINY, EAR, IVR, JUNCTIONAL, NOISE, NSR, SVT, TRIGEMINY, VT, and WENCKEBACH. Sudden bradycardia cases were excluded from the data as they were not included in the original formulation by the authors. The ECG frames were not normalized.

\textbf{PhysioNet 2017} \citep{Clifford2017}. Each ECG recording originally varied in length between 9 and 30 seconds with a sampling rate of 300Hz. Each ECG frames in our setup consisted of 2500 samples. We use the original labels, resulting in four classes: Normal, AF, Other, and Noisy. The ECG frames were not normalized. 

\clearpage
\subsection{Data Samples}
\label{appendix:instances}

\subsubsection{Self-supervised Pre-training}

In this section, we outline the dimension of the inputs used for the various pre-training methods. They are expressed in the form of $N \times S \times L$ where $N$ is the total number of instances, $S$ is the frame length of each instance, and $L$ (if applicable) is the number of leads used. Where $L$ is not explicitly mentioned, we report values with four leads as this was primarily used for all experiments conducted. 

\begin{table}[h]
\small
\centering
\caption{Dimension of the input data, $N \times S \times L$, used during the training and validation phases of the various self-supervised pre-training methods. $S=2500$ is the number of samples in each instance fed to the network. $L$ is the number of leads (projections) used during pre-training.}
\vskip 0.1in
\label{table:data_splits}
\begin{tabular}{c c | c c}
\toprule
Dataset & Method & Train & Validation\\
\midrule
\multirow{4}{*}{PhysioNet 2020}& 
BYOL & 51,880 $\times S$ & 12,948 $\times S$ \\
& SimCLR & 51,880 $\times S$ & 12,948 $\times S$ \\
                                & CMSC & 24,080 $\times 2S$ & 6,076 $\times 2S$ \\
                                & CMLC & 24,080 $\times S \times L$ & 6,076 $\times S \times L$\\
                                & CMSMLC & 6,020 $\times 2S \times L$& 1,519 $\times 2S \times L$ \\
\midrule
\multirow{4}{*}{Chapman}&
BYOL & 25,543 $\times S$ & 8,512 $\times S$ \\
& SimCLR & 25,543 $\times S$ & 8,512 $\times S$ \\
                                & CMSC & 25,543 $\times 2S$ & 8,512 $\times 2S$ \\
                                & CMLC & 25,543 $\times S \times L$ & 8,512  $\times S \times L$\\
                                & CMSMLC & 6,382 $\times 2S \times L$& 2125 $\times 2S \times L$\\

\bottomrule
\end{tabular}
\end{table}

\subsubsection{Supervised Training}

In this section, we outline the number of instances used during supervised training on the downstream tasks. For multi-lead datasets, we report these values having used four leads. A simple multiplicative factor can be used to deduce the number of instances used with a different number of leads. 

\begin{table}[h]
\small
\centering
\caption{Number of instances (number of patients) used during the supervised training of the downstream tasks. For multi-lead datasets*, these represent sample sizes for the four leads (II, V2, aVL, aVR).}
\vskip 0.1in
\label{table:data_splits}
\begin{tabular}{c | c c c}
\toprule
Dataset & Train & Validation & Test\\
\midrule
\multirow{1}{*}{PhysioNet 2020*}&51,880 (4,402)&12,948 (1,100)&15,820 (1,375)\\
\multirow{1}{*}{Chapman*}&25,543 (6,387)&8,512 (2,129)&8,520 (2,130)\\
\multirow{1}{*}{Cardiology}&4,584 (201)& 1,109 (50)& 1,386 (62)\\
\multirow{1}{*}{PhysioNet 2017}&18,256 (5,459)& 4,581 (1,364)& 5,824 (1,705)\\
\bottomrule \end{tabular}
\end{table}

\end{subappendices}

\clearpage
\begin{subappendices}
\renewcommand{\thesubsection}{\Alph{section}.\arabic{subsection}}
\section{Visualization of Data Augmentations}
\label{appendix:visualization_of_data_augmentations}

\begin{figure}[!h]
    \centering
    \begin{subfigure}{0.3\textwidth}
    \centering
    \includegraphics[width=\textwidth]{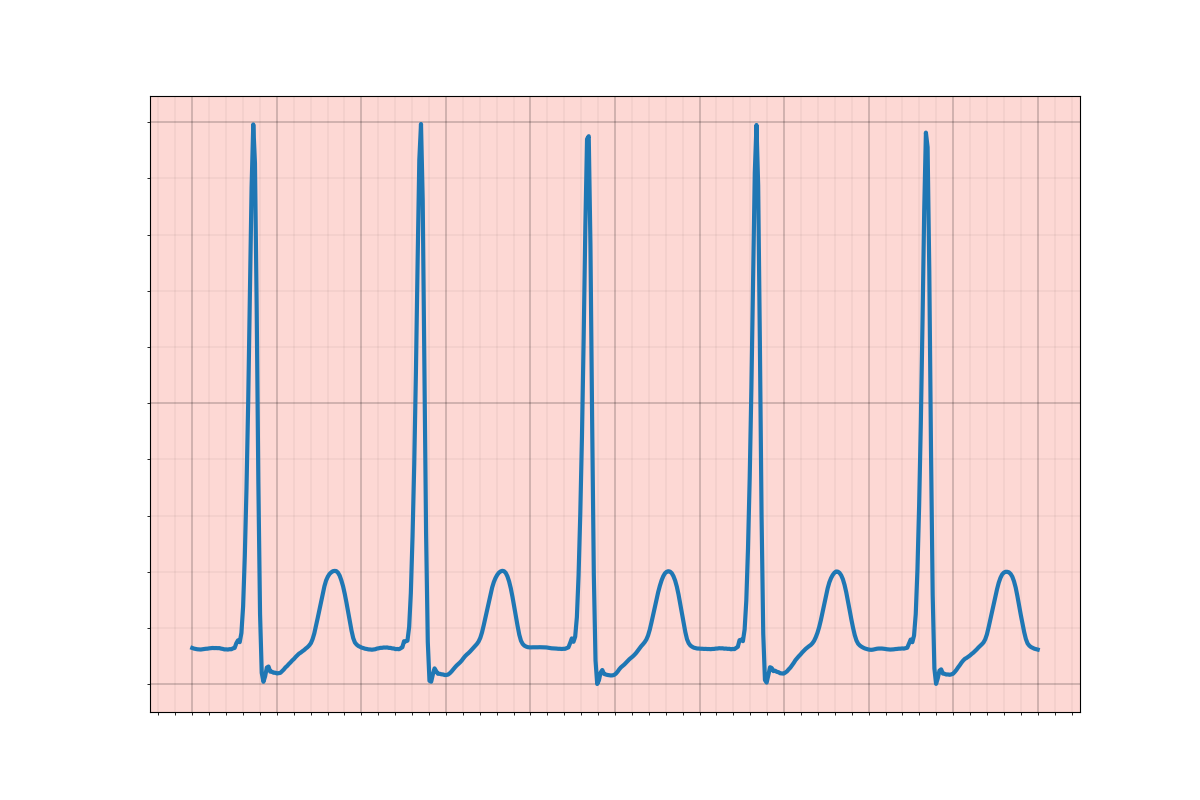}
    \caption{Original}
    \label{fig:frame}
    \end{subfigure}
    ~
    \begin{subfigure}{0.3\textwidth}
    \centering
    \includegraphics[width=\textwidth]{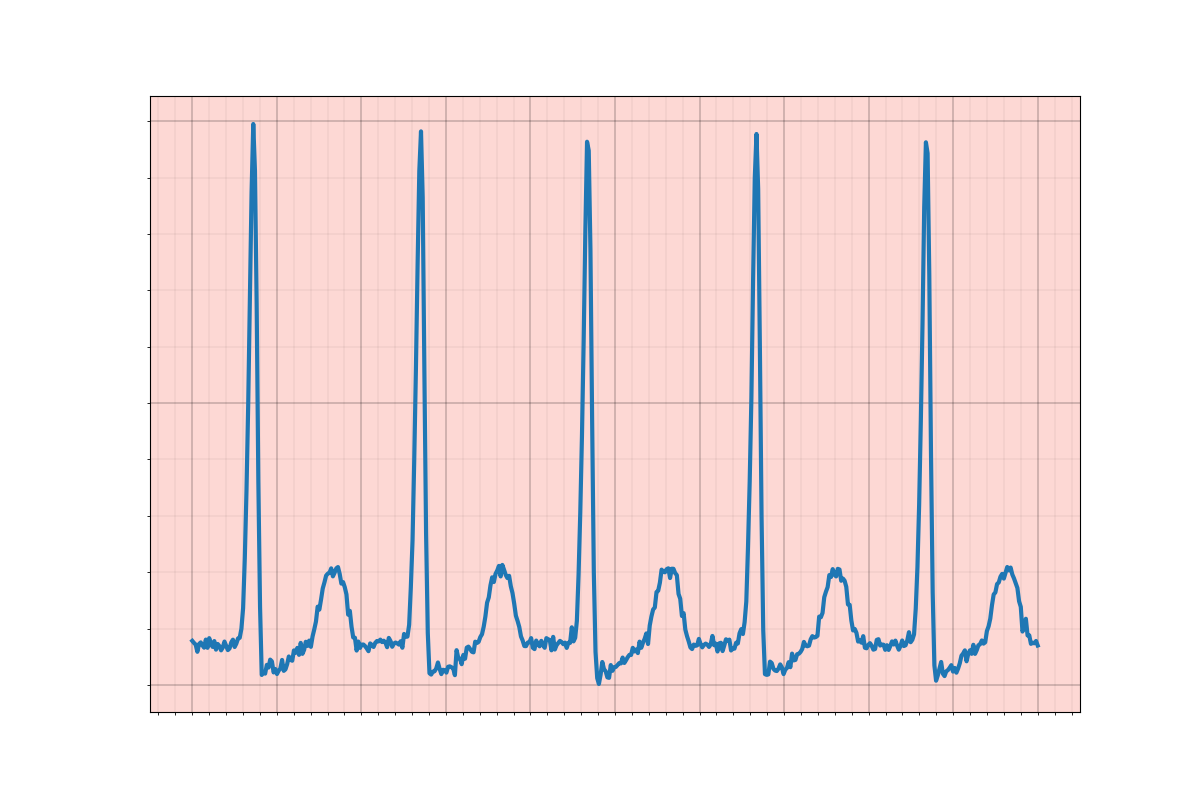}
    \caption{Gaussian Noise}
    \label{fig:frame_noise}
    \end{subfigure}
    ~
    \begin{subfigure}{0.3\textwidth}
    \centering
    \includegraphics[width=\textwidth]{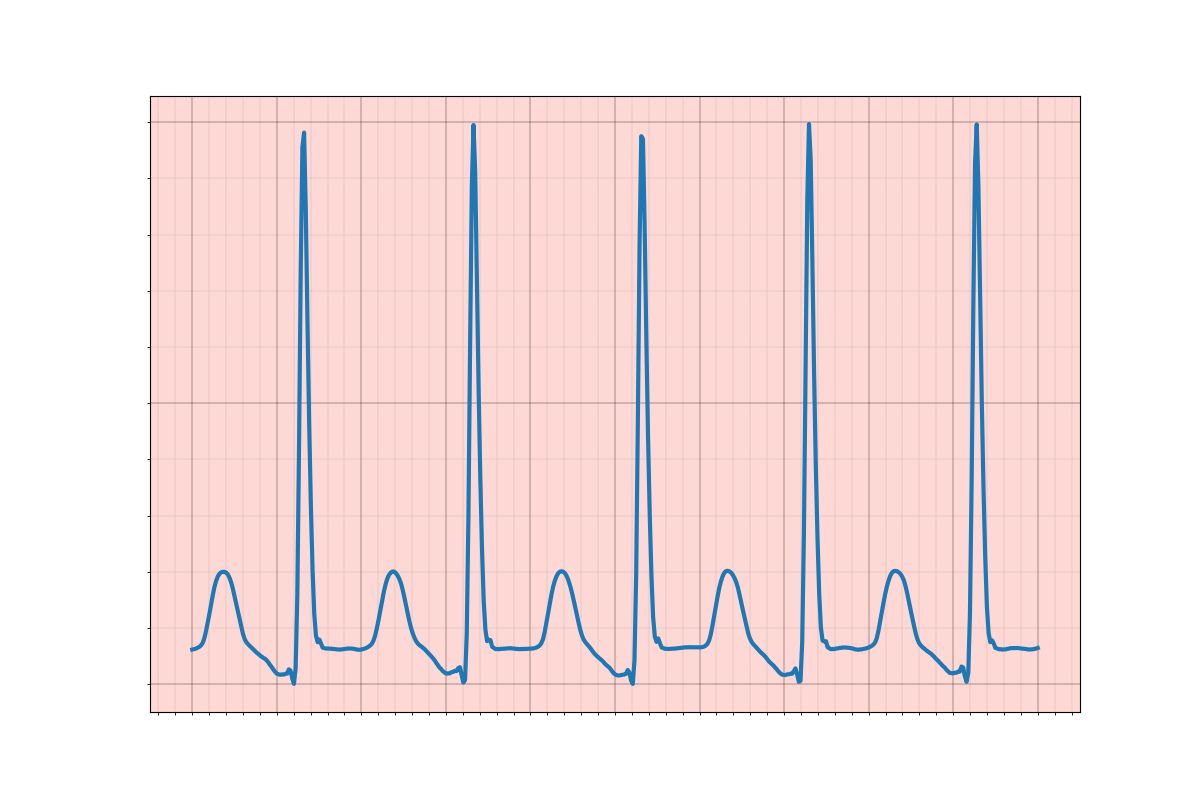}
    \caption{$\text{Flip}_{Y}$}
    \label{fig:frame_hflip}
    \end{subfigure}
    ~
    \begin{subfigure}{0.3\textwidth}
    \centering
    \includegraphics[width=\textwidth]{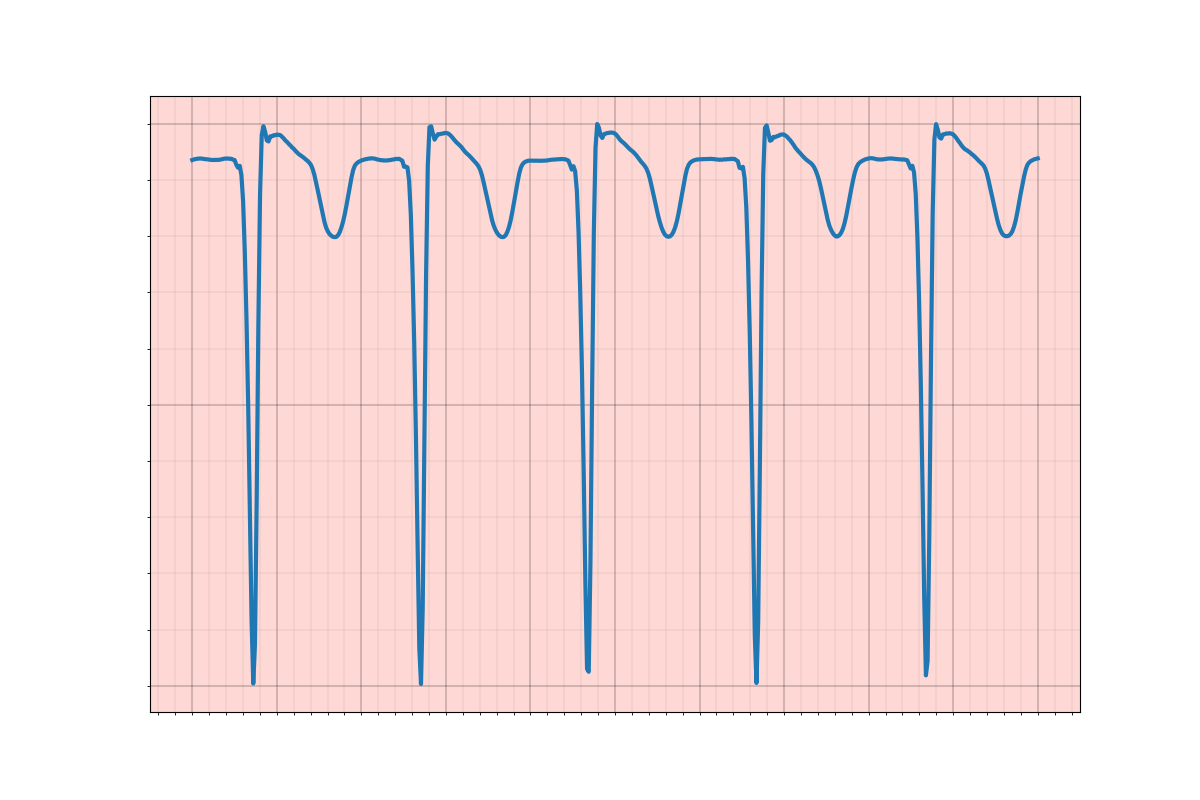}
    \caption{$\text{Flip}_{X}$}
    \label{fig:frame_vflip}
    \end{subfigure}
    \caption{ECG segment a) without any perturbations, b) with additive Gaussian noise, c) after being flipped temporally, $\text{Flip}_{Y}$, and d) after being flipped along the x-axis, $\text{Flip}_{X}$.}
    \label{fig:frame_perturbations}
\end{figure}

In this section, we outline the various data augmentations applied to the time-series signals and provide exemplar visualizations for the reader. In Fig.~\ref{fig:frame}, We present a single, unperturbed ECG frame for illustration purposes. To that original frame, we apply the following transformations:

\begin{enumerate}
    \item \textbf{Gaussian Noise}: Gaussian noise, $\epsilon \sim \mathcal{N}(0,\sigma)$ is added to the original frame. We chose the value of $\sigma$ in order to preserve the class of the original frame. More concretely, for the Chapman dataset, $\sigma=10$, whereas for the PTB-XL dataset, $\sigma=0.01$. The difference in the magnitude of $\sigma$ across datasets is attributed to the difference in the magnitude of the original signals from each dataset. Although it can be argued that such noise is trivial for a contrastive learning setup, recent work has shown the effect of additive noise on ECG signals \citep{Han2020}. 
    \item \textbf{$\text{Flip}_{Y}$}: we perturb the original frame by flipping it along the temporal dimension. In other words, the signal is read in reverse. In designing this perturbation, we were motivated by the self-supervision task proposed by \cite{} that revolves around reversing the 'arrow of time'.
    \item \textbf{$\text{Flip}_{X}$}: we perturb the original frame by negating the magnitude of the signal. This perturbation was motivated by the fact that ECG recordings made by physical leads that are connected to the patient's body incorrectly could lead to such 'inverted' signals. 
    \item \textbf{SpecAugment}: we perturb the original frame by masking spectral or temporal components of the signal. To do so, we follow a similar setup to that introduced in SpecAugment \citep{Park2019}. We take the Short-time Fourier transform (STFT) of the signal, randomly choose the number \textit{and} width of spectral or temporal bins to mask. As the resultant STFT is a matrix of complex numbers, we set masked values to zero. Lastly, we perform the Inverse STFT (ISTFT) to obtain the signal in the time-domain. We provide step-by-step instructions on how to apply this perturbation in the next section.
\end{enumerate}

\subsection{Perturbations}

\clearpage

\subsection{SpecAugment}

To apply the SpecAugment perturbations, we followed these steps:
\begin{enumerate}
    \item Apply the Short-time Fourier transform (STFT) to the time-series signal. This splits the signal into $N_{f}$ spectral and $N_{t}$ temporal bins. 
    \item Depending on whether a spectral or temporal mask is desired, the bin width, $w \in [0,1]$, defines the fraction of the total number of bins to mask. For example, $w=0.5$ means that 50\% of the bins are masked. The total number of bins to mask is thus $N_{m} = w \times N_{f}$ or $N_{m} = w \times N_{t}$. 
    \item Now that we have the number of bins to mask, we need to identify \textit{which} bins to mask. We formulate this as identifying the bin to start the masking and do so by uniformly sampling a number, $\mathrm{start}$, from $0$ to $N_{f} - N_{m}$. The masked bins range from $\mathrm{start}$ to $\mathrm{start} + N_{m}$.
    \item As the STFT of a time-series signal is a complex number, masking involves setting the complex-valued entries to zero, i.e., $0 + 0j$. 
    \item This process is repeated $R$ times until all desired components are masked.
    \item We convert the masked STFT back to the time-domain by taking its inverse (ISTFT). 
\end{enumerate}

In following the aforementioned steps, several hyperparameters exist. In our implementation, we chose $w=0.2$ and $R=1$ to balance between masking too many components which might violate the assumption of a shared context and masking too few components which would make the contrastive learning task quite trivial. 

\begin{figure}[!h]
    \centering
    \begin{subfigure}{0.45\textwidth}
    \centering
    \includegraphics[width=0.7\textwidth]{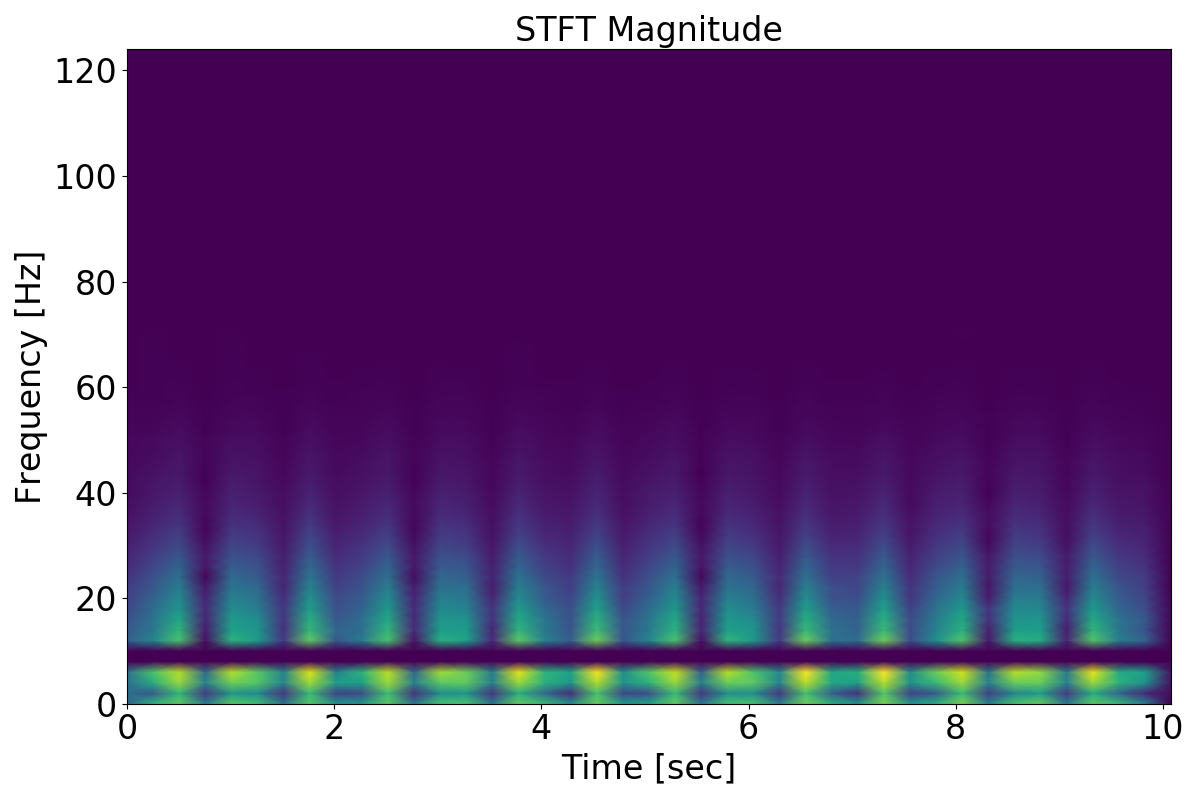}
    \caption{Spectrogram with spectral mask, $\text{SA}_{f}$}
    \label{fig:my_label}
    \end{subfigure}
    ~
    \begin{subfigure}{0.45\textwidth}
    \centering
    \includegraphics[width=0.7\textwidth]{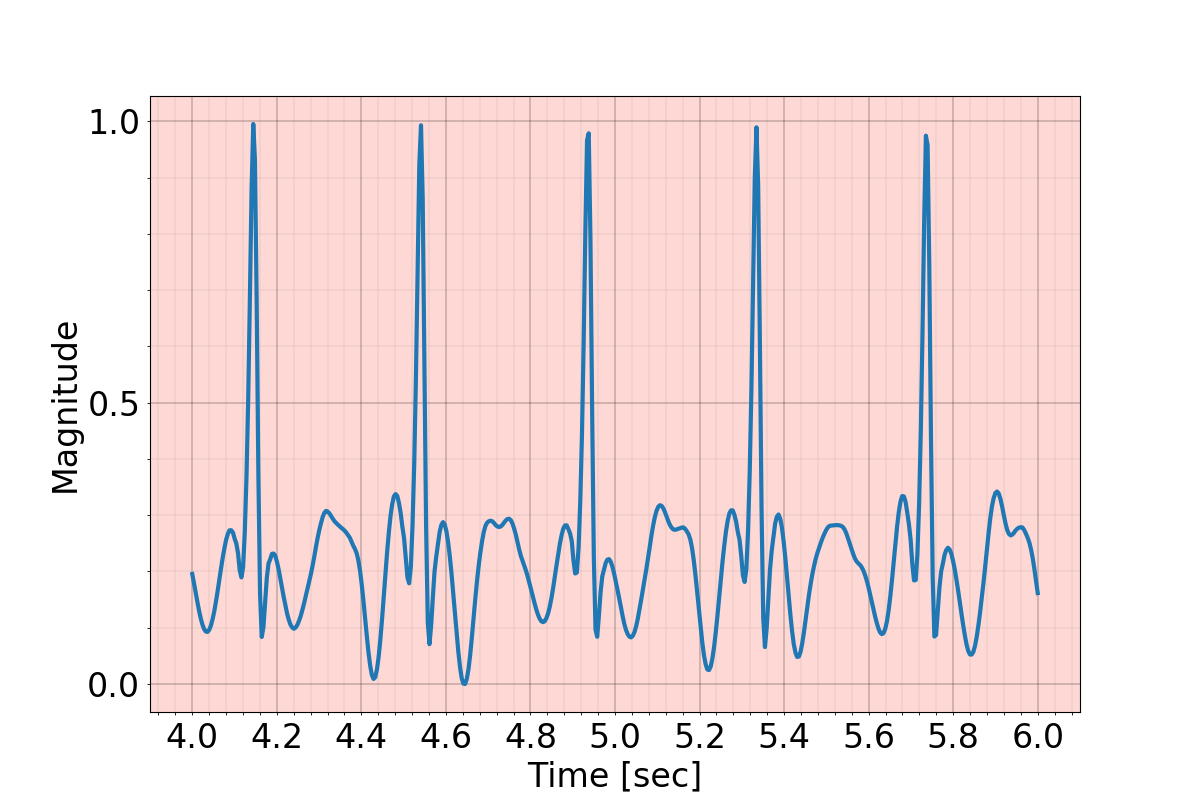}
    \caption{Frame with spectral mask}
    \label{fig:my_label}
    \end{subfigure}
    ~
    \begin{subfigure}{0.45\textwidth}
    \centering
    \includegraphics[width=0.7\textwidth]{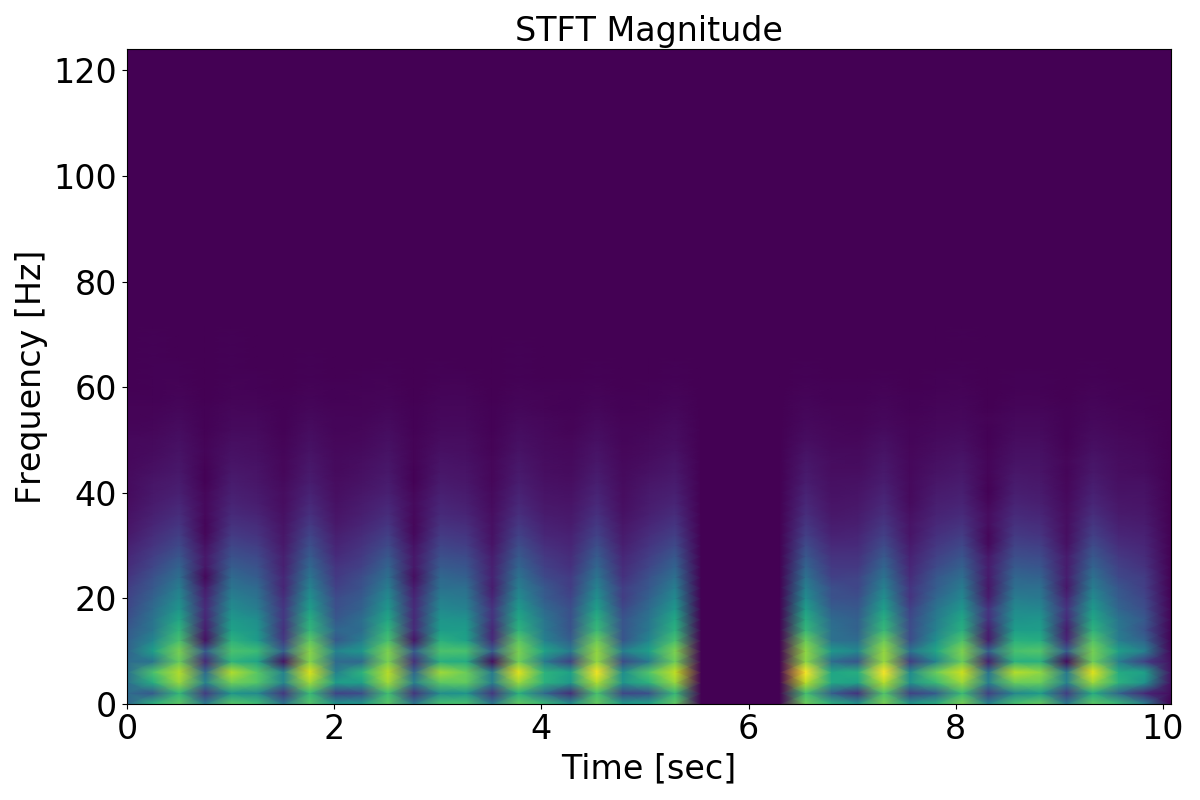}
    \caption{Sepctrogram with temporal mask, $\text{SA}_{t}$}
    \label{fig:my_label}
    \end{subfigure}
    ~
    \begin{subfigure}{0.45\textwidth}
    \centering
    \includegraphics[width=0.7\textwidth]{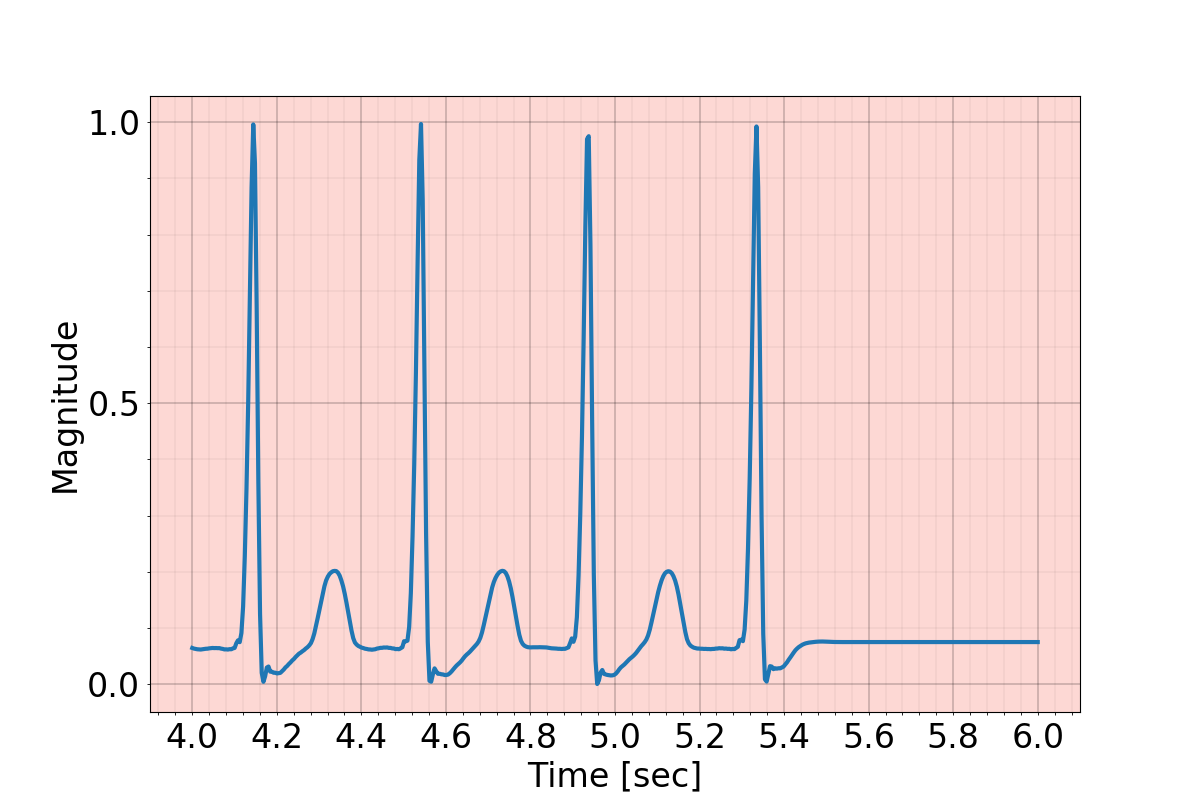}
    \caption{Frame with temporal mask}
    \label{fig:my_label}
    \end{subfigure}
    \caption{Illustration of SpecAugment perturbations applied to the original ECG segment shown in Fig.~\ref{fig:frame_perturbations}. (a), (c) spectrograms with spectral and temporal masks, respectively. (b), (d) time-series representations of the masked spectrograms. Note that the time-series segments only span seconds 4-6.}
    \label{fig:spec_augment_perturbations}
\end{figure}

\end{subappendices}

\clearpage
\begin{subappendices}
\renewcommand{\thesubsection}{\Alph{section}.\arabic{subsection}}
\section{Implementation Details}
\label{appendix:implementation}

\subsection{Network Architecture}
\label{appendix:network}

In this section, we outline the architecture of the neural network used for all experiments. For pre-training, the final layer (Layer 5) was removed and representations with dimension $E$ were learned. During training on the downstream tasks, the final layer was introduced. 

\begin{table}[h]
\small
\centering
\caption{Network architecture used for all experiments. \textit{K}, \textit{C}\textsubscript{in}, and \textit{C}\textsubscript{out} represent the kernel size, number of input channels, and number of output channels, respectively. A stride of 3 was used for all convolutional layers. $E$ represents the dimension of the final representation.}
\vskip 0.1in
\label{table:network_architecture}
\begin{tabular}{c c c}
\toprule
Layer Number &Layer Components&Kernel Dimension\\
\midrule
	\multirow{5}{*}{1}&Conv 1D & 7 x 1 x 4 (\textit{K} x \textit{C}\textsubscript{in} x \textit{C}\textsubscript{out})\\
										& BatchNorm &\\
										& ReLU& \\
										& MaxPool(2)& \\
										& Dropout(0.1) &\\
	\midrule
	\multirow{5}{*}{2}&Conv 1D & 7 x 4 x 16\\
										& BatchNorm& \\
										& ReLU &\\
										& MaxPool(2) &\\
										& Dropout(0.1)& \\
	\midrule
	\multirow{5}{*}{3}&Conv 1D & 7 x 16 x 32 \\
										& BatchNorm &\\
										& ReLU &\\
										& MaxPool(2) &\\
										& Dropout(0.1) &\\
	\midrule
	\multirow{2}{*}{4}&Linear&320 x $E$ \\
										& ReLU &\\
	\midrule
	\multirow{1}{*}{5}&Linear &$E$ x C (classes) \\
\bottomrule \end{tabular}
\end{table}

\subsection{Experiment Details}

\begin{table}[h]
\small
\centering
\caption{Batchsize and learning rates used for training with different datasets. The Adam optimizer was used for all experiments.}
\vskip 0.1in
\label{table:batchsize}
\begin{tabular}{c | c c }
\toprule
Dataset&Batchsize&Learning Rate\\
\midrule
PhysioNet 2020 & 256 & 10\textsuperscript{-4}\\
Chapman & 256 & 10\textsuperscript{-4}\\
Cardiology & 16 & 10\textsuperscript{-4}\\
PhysioNet 2017 & 256 & 10\textsuperscript{-4}\\
\bottomrule \end{tabular}
\end{table}

\clearpage

\subsection{Baseline Implementations}

\subsubsection{Supervised Pre-training}
In this implementation, we pre-train on the specified dataset under the assumption that 100\% of the data is labelled and available for training (i.e., $F=1$). Given the presence of labels, pre-training involves solving a supervised classification task to diagnose the cardiac arrhythmia that corresponds to each ECG recording. In our context, supervised pre-training is expected to generate the best downstream generalization performance due to the availability of labels \textit{and} the high similarity between the upstream and downstream tasks, namely cardiac arrhythmia classification. 

\subsubsection{MT-SSL}
In this implementation, we introduce six different pre-text tasks that are used for pre-training a network. We follow the multi-task pre-training setup proposed by \citep{Sarkar2020} where six different classification heads are used to solve each of the six tasks. These tasks comprise binary classification where the network is asked to discriminate between ECG instances and their perturbed counterpart. Such perturbations take on the form of 1) Gaussian noise addition, 2) scaling, 3) negation, 4) temporal inversion, 5) permutation, and 6) time-warping. For the Chapman dataset, we only pre-train using scaling, negation, and temporal inversion since additional tasks prevented the network from converging. On the PhysioNet2020 dataset, however, we pre-train using all of the aforementioned tasks.

\subsubsection{BYOL}
In this implementation, an instance is perturbed by applying two stochastic transformations. In our setup, these transformations can include any of those outlined in Appendix~\ref{appendix:visualization_of_data_augmentations}. This process results in two views of the same instance, each of which is passed through an online network and a target network. The target network is an exponential moving average of the online network, and is thus a delayed version of the online network. This delay is dictated by the decay rate, $\tau_{d}$. We chose $\tau_{d}=0.9$ with experiments to validate this decision in Appendix~\ref{appendix:effect_of_tau}. A key difference between the two networks is that they are \textit{asymmetric}, with the online network consisting of an additional prediction head. The goal is for the representation from the online network to predict that from the target network. This is done by minimizing the mean squared error of the two representations. In our setup, we introduce asymmetry by repeating Layer 4 shown in Appendix~\ref{appendix:network}. This is similar to what was performed by \cite{Grill2020}. 

\subsubsection{SimCLR}
In this implementation, an instance is perturbed by applying two stochastic transformations. In our setup, these transformations can include any of those outlined in Appendix~\ref{appendix:visualization_of_data_augmentations}. This process results in two views of the same instance, each of which is passed through the same network. The InfoNCE loss is used to attract representations that are similar to one another and repel those that are different. Whether representations should be attracted to one another depends on whether they belong to the same original instance.

\end{subappendices}

\clearpage

\begin{subappendices}
\renewcommand{\thesubsection}{\Alph{section}.\arabic{subsection}}
\section{Linear Evaluation of Representations}
\label{appendix:linear_evaluation}

In this section, we evaluate the utility of the representations learned as a result of self-supervised pre-training. We pre-train on two different datasets, freeze the network parameters, and transfer them to a downstream task whereby a linear multinomial logistic regression (MLR) model is trained. In doing so, we are evaluating the richness of the representations learned. We perform these experiments under two scenarios. The first involves pre-training and evaluating using 4 leads (II, V2, aVL, aVR) (see Sec.~\ref{sec:linear_4_leads}). The second involves pre-training and evaluating using all 12 leads (see Sec.~\ref{sec:linear_12_leads}). We chose these two scenarios to help determine whether our findings generalize to domains where a different number of leads is available.

\subsection{Pre-training and Evaluating using 4 leads}
\label{sec:linear_4_leads}

We present Tables~\ref{table:test_auc_linear_E_32} - \ref{table:test_auc_linear_E_256} which illustrate the test AUC of an MLR evaluated on Chapman and PhysioNet 2020 after having pre-trained on these two datasets using only 4 of the 12 leads, respectively, These are presented for a range of embedding dimensions, $E=(32,64,128,256)$, and available labelled training data, $F=(0.25,0.50,0.75,1)$.  

\subsubsection{Embedding Dimension, $E=32$}

We show that CMSMLC outperforms all other methods when evaluating on Chapman, regardless of the available labelled training data. This can be seen by the higher AUC achieved by this method relative to the remaining methods. For instance, at $F=0.25$, CMSMLC achieves an $\mathrm{AUC}=0.844$ compared to $0.665$ for SimCLR. When evaluating on PhysioNet 2020, we find that CMSC consistently outperforms the remaining methods, as seen by its higher test AUC values. 

\begin{table}[!h]
\centering
\caption{Comparison of self-supervised methods when using networks as feature extractors and performing linear evaluation on downstream datasets. Pre-training and evaluating multi-lead datasets* using 4 leads. Mean and standard deviation are shown across 5 seeds.}
\label{table:test_auc_linear_E_32}
\vskip 0.1in 
\begin{subtable}{\textwidth}
\centering
\caption{$F=0.25$}
\label{table:linear_E_32}
\begin{tabular}{c | c c c | c c c }
\toprule
Dataset&\multicolumn{1}{c}{Chapman*}&\multicolumn{1}{c}{PhysioNet 2020*}\\
\midrule
SimCLR & \multicolumn{1}{c}{0.665 $\pm$ 0.014} & \multicolumn{1}{c}{0.564 $\pm$ 0.009} \\
CMSC & \multicolumn{1}{c}{0.831 $\pm$ 0.131} & \multicolumn{1}{c}{\textbf{0.701 $\pm$ 0.046}} \\
CMLC & \multicolumn{1}{c}{0.789 $\pm$ 0.020} & \multicolumn{1}{c}{0.563 $\pm$ 0.008} \\
CMSMLC & \multicolumn{1}{c}{\textbf{0.844 $\pm$ 0.023}} & \multicolumn{1}{c}{0.619 $\pm$ 0.019} \\
\bottomrule 
\end{tabular}
\end{subtable}

\vskip 0.1in \begin{subtable}{\textwidth}
\centering
\caption{$F=0.5$}
\label{table:linear_32_0.5}
\begin{tabular}{c | c c c | c c c }
\toprule
Dataset&\multicolumn{1}{c}{Chapman*}&\multicolumn{1}{c}{PhysioNet 2020*}\\
\midrule
SimCLR & \multicolumn{1}{c}{0.666 $\pm$ 0.015} & \multicolumn{1}{c}{0.587 $\pm$ 0.009} \\
CMSC &\multicolumn{1}{c}{0.831 $\pm$ 0.131} & \multicolumn{1}{c}{\textbf{0.707 $\pm$ 0.038}} \\
CMLC &\multicolumn{1}{c}{0.801 $\pm$ 0.016} & \multicolumn{1}{c}{0.572 $\pm$ 0.008} \\
CMSMLC &\multicolumn{1}{c}{\textbf{0.850 $\pm$ 0.022}} & \multicolumn{1}{c}{0.636 $\pm$ 0.020} \\
\bottomrule \end{tabular}
\end{subtable}

\vskip 0.1in \begin{subtable}{\textwidth}
\centering
\caption{$F=0.75$}
\label{table:linear_32_0.75}
\begin{tabular}{c | c c c | c c c }
\toprule
Dataset&\multicolumn{1}{c}{Chapman*}&\multicolumn{1}{c}{PhysioNet 2020*}\\
\midrule
SimCLR &\multicolumn{1}{c}{0.670 $\pm$ 0.013} & \multicolumn{1}{c}{0.591 $\pm$ 0.010} \\
CMSC &\multicolumn{1}{c}{0.833 $\pm$ 0.129}& \multicolumn{1}{c}{\textbf{0.709 $\pm$ 0.039}} \\
CMLC &\multicolumn{1}{c}{0.805 $\pm$ 0.018} & \multicolumn{1}{c}{0.585 $\pm$ 0.009} \\
CMSMLC &\multicolumn{1}{c}{\textbf{0.850 $\pm$ 0.021}} & \multicolumn{1}{c}{0.643 $\pm$ 0.020} \\
\bottomrule \end{tabular}
\end{subtable}

\vskip 0.1in \begin{subtable}{\textwidth}
\centering
\caption{$F=1$}
\label{table:linear_32_1}
\begin{tabular}{c | c c c | c c c }
\toprule
Dataset&\multicolumn{1}{c}{Chapman*}&\multicolumn{1}{c}{PhysioNet 2020*}\\
\midrule
SimCLR &\multicolumn{1}{c}{0.670 $\pm$ 0.013} & \multicolumn{1}{c}{0.594 $\pm$ 0.010} \\
CMSC &\multicolumn{1}{c}{0.831 $\pm$ 0.131} & \multicolumn{1}{c}{\textbf{0.709 $\pm$ 0.038}} \\
CMLC &\multicolumn{1}{c}{0.807 $\pm$ 0.017} & \multicolumn{1}{c}{0.593 $\pm$ 0.009} \\
CMSMLC &\multicolumn{1}{c}{\textbf{0.852 $\pm$ 0.021}} & \multicolumn{1}{c}{0.645 $\pm$ 0.021} \\
\bottomrule \end{tabular}
\end{subtable}
\end{table}

\clearpage

\subsubsection{Embedding Dimension, $E=64$}

We find that the conclusions arrived at with $E=32$ are similar to those in this scenario. Namely, CMSMLC outperforms all remaining methods when evaluating on Chapman. On the other hand, CMSC outperforms all methods when evaluating on PhysioNet 2020. This can be seen by the bold test AUC values in Table~\ref{table:test_auc_linear_E_64}. 

\begin{table}[!h]
\centering
\caption{Comparison of self-supervised methods when using networks as feature extractors and performing linear evaluation on downstream datasets. Pre-training and evaluating multi-lead datasets* using 4 leads. Mean and standard deviation are shown across 5 seeds.}
\label{table:test_auc_linear_E_64}
\vskip 0.1in 
\begin{subtable}{\textwidth}
\centering
\caption{$F=0.25$}
\label{table:linear_64_0.25}
\begin{tabular}{c | c c c | c c c }
\toprule
Dataset&\multicolumn{1}{c}{Chapman*}&\multicolumn{1}{c}{PhysioNet 2020*}\\
\midrule
SimCLR &\multicolumn{1}{c}{0.709 $\pm$ 0.019} & \multicolumn{1}{c}{0.574 $\pm$ 0.005} \\
CMSC &\multicolumn{1}{c}{0.829 $\pm$ 0.130} & \multicolumn{1}{c}{\textbf{0.720 $\pm$ 0.012}} \\
CMLC &\multicolumn{1}{c}{0.842 $\pm$ 0.020} & \multicolumn{1}{c}{0.592 $\pm$ 0.019} \\
CMSMLC &\multicolumn{1}{c}{\textbf{0.856 $\pm$ 0.022}} & \multicolumn{1}{c}{0.641 $\pm$ 0.023} \\
\bottomrule \end{tabular}
\end{subtable}

\vskip 0.1in \begin{subtable}{\textwidth}
\centering
\caption{$F=0.5$}
\label{table:linear_64_0.5}
\begin{tabular}{c | c c c | c c c }
\toprule
Dataset&\multicolumn{1}{c}{Chapman*}&\multicolumn{1}{c}{PhysioNet 2020*}\\
\midrule
SimCLR &\multicolumn{1}{c}{0.722 $\pm$ 0.025} & \multicolumn{1}{c}{0.599 $\pm$ 0.010} \\
CMSC &\multicolumn{1}{c}{0.830 $\pm$ 0.132} & \multicolumn{1}{c}{\textbf{0.721 $\pm$ 0.013}} \\
CMLC &\multicolumn{1}{c}{0.850 $\pm$ 0.02} & \multicolumn{1}{c}{0.607 $\pm$ 0.018} \\
CMSMLC &\multicolumn{1}{c}{\textbf{0.861 $\pm$ 0.02}} & \multicolumn{1}{c}{0.662 $\pm$ 0.020} \\
\bottomrule \end{tabular}
\end{subtable}

\vskip 0.1in \begin{subtable}{\textwidth}
\centering
\caption{$F=0.75$}
\label{table:linear_64_0.75}
\begin{tabular}{c | c c c | c c c }
\toprule
Dataset&\multicolumn{1}{c}{Chapman*}&\multicolumn{1}{c}{PhysioNet 2020*}\\
\midrule
SimCLR &\multicolumn{1}{c}{0.726 $\pm$ 0.023} & \multicolumn{1}{c}{0.604 $\pm$ 0.010} \\
CMSC &\multicolumn{1}{c}{0.831 $\pm$ 0.126} & \multicolumn{1}{c}{\textbf{0.725 $\pm$ 0.009}} \\
CMLC &\multicolumn{1}{c}{0.854 $\pm$ 0.021} & \multicolumn{1}{c}{0.619 $\pm$ 0.017} \\
CMSMLC &\multicolumn{1}{c}{\textbf{0.861 $\pm$ 0.021}} & \multicolumn{1}{c}{0.671 $\pm$ 0.018} \\
\bottomrule \end{tabular}
\end{subtable}

\vskip 0.1in \begin{subtable}{\textwidth}
\centering
\caption{$F=1$}
\label{table:linear_64_1}
\begin{tabular}{c | c c c | c c c }
\toprule
Dataset&\multicolumn{1}{c}{Chapman*}&\multicolumn{1}{c}{PhysioNet 2020*}\\
\midrule
SimCLR &\multicolumn{1}{c}{0.727 $\pm$ 0.025} & \multicolumn{1}{c}{0.608 $\pm$ 0.010} \\
CMSC &\multicolumn{1}{c}{0.832 $\pm$ 0.126} & \multicolumn{1}{c}{\textbf{0.726 $\pm$ 0.009}} \\
CMLC &\multicolumn{1}{c}{0.855 $\pm$ 0.020} & \multicolumn{1}{c}{0.627 $\pm$ 0.015} \\
CMSMLC &\multicolumn{1}{c}{\textbf{0.862 $\pm$ 0.020}} & \multicolumn{1}{c}{0.673 $\pm$ 0.018} \\
\bottomrule \end{tabular}
\end{subtable}
\end{table}

\clearpage

\subsubsection{Embedding Dimension = 128}

In this scenario and in contrast to conclusions arrived at with $E=32$ and $64$, we find that CMSC outperforms all methods when evaluated on both datasets, Chapman and PhysioNet 2020. This can be seen by the bold test AUC values in Table~\ref{table:test_auc_linear_E_128}. For instance, at $F=0.25$, CMSC achieves an $\mathrm{AUC}=0.895$ compared to $0.727$ achieved by SimCLR. That is a 16.8\% improvement relative to the state-of-the-art. 

\begin{table}[!h]
\centering
\caption{Comparison of self-supervised methods when using networks as feature extractors and performing linear evaluation on downstream datasets. Pre-training and evaluating multi-lead datasets* using 4 leads. Mean and standard deviation are shown across 5 seeds.}
\label{table:test_auc_linear_E_128}
\vskip 0.1in 
\begin{subtable}{\textwidth}
\centering
\caption{$F=0.25$}
\label{table:linear_32_0.25}
\begin{tabular}{c | c c c | c c c }
\toprule
Dataset&\multicolumn{1}{c}{Chapman*}&\multicolumn{1}{c}{PhysioNet 2020*}\\
\midrule
BYOL &0.671 $\pm$ 0.042 & 0.587 $\pm$ 0.021 \\ 
SimCLR &\multicolumn{1}{c}{0.727 $\pm$ 0.032} & \multicolumn{1}{c}{0.585 $\pm$ 0.016} \\
CMSC &\multicolumn{1}{c}{\textbf{0.895 $\pm$ 0.004}} & \multicolumn{1}{c}{\textbf{0.713 $\pm$ 0.032}} \\
CMLC &\multicolumn{1}{c}{0.863 $\pm$ 0.026} & \multicolumn{1}{c}{0.580 $\pm$ 0.007} \\
CMSMLC &\multicolumn{1}{c}{0.842 $\pm$ 0.021} & \multicolumn{1}{c}{0.661 $\pm$ 0.010} \\
\bottomrule \end{tabular}
\end{subtable}

\vskip 0.1in \begin{subtable}{\textwidth}
\centering
\caption{$F=0.5$}
\label{table:linear_32_0.25}
\begin{tabular}{c | c c c | c c c }
\toprule
Dataset&\multicolumn{1}{c}{Chapman*}&\multicolumn{1}{c}{PhysioNet 2020*}\\
\midrule
BYOL &0.643 $\pm$ 0.043 & 0.595 $\pm$ 0.018 \\
SimCLR &\multicolumn{1}{c}{0.738 $\pm$ 0.034} & \multicolumn{1}{c}{0.615 $\pm$ 0.014} \\
CMSC &\multicolumn{1}{c}{\textbf{0.896 $\pm$ 0.005}} & \multicolumn{1}{c}{\textbf{0.715 $\pm$ 0.033}} \\
CMLC &\multicolumn{1}{c}{0.870 $\pm$ 0.022} & \multicolumn{1}{c}{0.596 $\pm$ 0.008} \\
CMSMLC &\multicolumn{1}{c}{0.847 $\pm$ 0.024} & \multicolumn{1}{c}{0.680 $\pm$ 0.008} \\
\bottomrule \end{tabular}
\end{subtable}

\vskip 0.1in \begin{subtable}{\textwidth}
\centering
\caption{$F=0.75$}
\label{table:linear_32_0.25}
\begin{tabular}{c | c c c | c c c }
\toprule
Dataset&\multicolumn{1}{c}{Chapman*}&\multicolumn{1}{c}{PhysioNet 2020*}\\
\midrule
BYOL &0.666 $\pm$ 0.032 & 0.598 $\pm$ 0.022 \\
SimCLR &\multicolumn{1}{c}{0.742 $\pm$ 0.033} & \multicolumn{1}{c}{0.620 $\pm$ 0.015} \\
CMSC &\multicolumn{1}{c}{\textbf{0.898 $\pm$ 0.002}} & \multicolumn{1}{c}{\textbf{0.717 $\pm$ 0.033}} \\
CMLC &\multicolumn{1}{c}{0.872 $\pm$ 0.022} & \multicolumn{1}{c}{0.606 $\pm$ 0.008} \\
CMSMLC &\multicolumn{1}{c}{0.848 $\pm$ 0.023} & \multicolumn{1}{c}{0.685 $\pm$ 0.008} \\
\bottomrule \end{tabular}
\end{subtable}

\vskip 0.1in \begin{subtable}{\textwidth}
\centering
\caption{$F=1$}
\label{table:linear_32_0.25}
\begin{tabular}{c | c c c | c c c }
\toprule
Dataset&\multicolumn{1}{c}{Chapman*}&\multicolumn{1}{c}{PhysioNet 2020*}\\
\midrule
BYOL &0.653 $\pm$ 0.026 & 0.602 $\pm$ 0.015 \\
SimCLR &\multicolumn{1}{c}{0.742 $\pm$ 0.033} &\multicolumn{1}{c}{0.623 $\pm$ 0.014} \\
CMSC &\multicolumn{1}{c}{\textbf{0.897 $\pm$ 0.003}} & \multicolumn{1}{c}{\textbf{0.718 $\pm$ 0.033}} \\
CMLC &\multicolumn{1}{c}{0.873 $\pm$ 0.021} & \multicolumn{1}{c}{0.612 $\pm$ 0.010} \\
CMSMLC &\multicolumn{1}{c}{0.849 $\pm$ 0.022} & \multicolumn{1}{c}{0.686 $\pm$ 0.008} \\
\bottomrule \end{tabular}
\end{subtable}
\end{table}

\clearpage

\subsubsection{Embedding Dimension, $E=256$}

In this scenario, we find that CMLC consistently outperforms all methods when evaluated on Chapman. Similar to findings at lower embedding dimensions, CMSC outperforms all methods on PhysioNet 2020. These claims are supported by the bold test AUC values in Table~\ref{table:test_auc_linear_E_256}. 

\begin{table}[!h]
\centering
\caption{Comparison of self-supervised methods when using networks as feature extractors and performing linear evaluation on downstream datasets. Pre-training and evaluating multi-lead datasets* using 4 leads. Mean and standard deviation are shown across 5 seeds.}
\label{table:test_auc_linear_E_256}
\vskip 0.1in 
\begin{subtable}{\textwidth}
\centering
\caption{$F=0.25$}
\label{table:linear_64_0.25}
\begin{tabular}{c | c c c | c c c }
\toprule
Dataset&\multicolumn{1}{c}{Chapman*}&\multicolumn{1}{c}{PhysioNet 2020*}\\
\midrule
SimCLR &\multicolumn{1}{c}{0.742 $\pm$ 0.031} & \multicolumn{1}{c}{0.591 $\pm$ 0.007} \\
CMSC &\multicolumn{1}{c}{0.832 $\pm$ 0.128} & \multicolumn{1}{c}{\textbf{0.721 $\pm$ 0.016}} \\
CMLC &\multicolumn{1}{c}{\textbf{0.883 $\pm$ 0.009}} & \multicolumn{1}{c}{0.607 $\pm$ 0.027} \\
CMSMLC &\multicolumn{1}{c}{0.828 $\pm$ 0.040} & \multicolumn{1}{c}{0.652 $\pm$ 0.023} \\
\bottomrule \end{tabular}
\end{subtable}

\vskip 0.1in \begin{subtable}{\textwidth}
\centering
\caption{$F=0.5$}
\label{table:linear_64_0.25}
\begin{tabular}{c | c c c | c c c }
\toprule

Dataset&\multicolumn{1}{c}{Chapman*}&\multicolumn{1}{c}{PhysioNet 2020*}\\
\midrule
SimCLR &\multicolumn{1}{c}{0.749 $\pm$ 0.032} & \multicolumn{1}{c}{0.615 $\pm$ 0.010} \\
CMSC &\multicolumn{1}{c}{0.833 $\pm$ 0.130} & \multicolumn{1}{c}{\textbf{0.722 $\pm$ 0.017}} \\
CMLC &\multicolumn{1}{c}{\textbf{0.887 $\pm$ 0.008}} & \multicolumn{1}{c}{0.619 $\pm$ 0.026} \\
CMSMLC &\multicolumn{1}{c}{0.831 $\pm$ 0.042} & \multicolumn{1}{c}{0.670 $\pm$ 0.018} \\
\bottomrule \end{tabular}
\end{subtable}

\vskip 0.1in \begin{subtable}{\textwidth}
\centering
\caption{$F=0.75$}
\label{table:linear_64_0.25}
\begin{tabular}{c | c c c | c c c }
\toprule
Dataset&\multicolumn{1}{c}{Chapman*}&\multicolumn{1}{c}{PhysioNet 2020*}\\
\midrule
SimCLR &\multicolumn{1}{c}{0.752 $\pm$ 0.033} & \multicolumn{1}{c}{0.619 $\pm$ 0.010} \\
CMSC &\multicolumn{1}{c}{0.833 $\pm$ 0.130} & \multicolumn{1}{c}{\textbf{0.723 $\pm$ 0.017}} \\
CMLC &\multicolumn{1}{c}{\textbf{0.889 $\pm$ 0.007}} & \multicolumn{1}{c}{0.626 $\pm$ 0.026} \\
CMSMLC &\multicolumn{1}{c}{0.831 $\pm$ 0.040} & \multicolumn{1}{c}{0.675 $\pm$ 0.018} \\
\bottomrule \end{tabular}
\end{subtable}

\vskip 0.1in \begin{subtable}{\textwidth}
\centering
\caption{$F=1$}
\label{table:linear_64_0.25}
\begin{tabular}{c | c c c | c c c }
\toprule

Dataset&\multicolumn{1}{c}{Chapman*}&\multicolumn{1}{c}{PhysioNet 2020*}\\
\midrule
SimCLR &\multicolumn{1}{c}{0.753 $\pm$ 0.033} & \multicolumn{1}{c}{0.621 $\pm$ 0.010} \\
CMSC &\multicolumn{1}{c}{0.833 $\pm$ 0.132} & \multicolumn{1}{c}{\textbf{0.724 $\pm$ 0.017}} \\
CMLC &\multicolumn{1}{c}{\textbf{0.890 $\pm$ 0.006}} & \multicolumn{1}{c}{0.633 $\pm$ 0.026} \\
CMSMLC &\multicolumn{1}{c}{0.832 $\pm$ 0.039} & \multicolumn{1}{c}{0.677 $\pm$ 0.018} \\
\bottomrule \end{tabular}
\end{subtable}
\end{table}

\clearpage
\subsection{Pre-training and Evaluating using 12 leads}
\label{sec:linear_12_leads}

We present Tables~\ref{table:test_auc_linear_L_12_E_32} - \ref{table:test_auc_linear_L_12_E_256} which illustrate the test AUC of an MLR evaluated on Chapman and PhysioNet 2020 after having pre-trained on these two datasets using all 12 leads, respectively, These are presented for a range of embedding dimensions, $E=(32,64,128,256)$, and available labelled training data, $F=(0.25,0.50,0.75,1)$. Overall, we find that pre-training and evaluating with all 12 leads results in a clearly superior self-supervised method, CMSC. We support this claim with the results presented in the subsequent sections. This finding is in contrast to what we observed when pre-training and evaluating on only 4 of the 12 leads. In that scenario, although our proposed methods outperform SimCLR, CMSC does not consistently outperform the other methods. 

\subsubsection{Embedding Dimension, $E=32$}

We show that CMSC consistently outperforms all other pre-training methods when evaluated on both the Chapman and PhysioNet 2020 dataset. This can be seen by the higher AUC achieved by this method relative to the remaining methods. For instance, when evaluating on the Chapman dataset using only 25\% of the labels ($F=0.25$) during training, CMSC achieves an $\mathrm{AUC}=0.899$ compared to $0.667$ for SimCLR.

\begin{table}[!h]
\centering
\caption{Comparison of self-supervised methods when using networks as feature extractors and performing linear evaluation on downstream datasets. Pre-training and evaluating multi-lead datasets* using all 12 leads. Mean and standard deviation are shown across 5 seeds.}
\label{table:test_auc_linear_L_12_E_32}
\vskip 0.1in 
\begin{subtable}{\textwidth}
\centering
\caption{$F=0.25$}
\label{table:linear_E_32}
\begin{tabular}{c | c c c | c c c }
\toprule
Dataset&\multicolumn{1}{c}{Chapman*}&\multicolumn{1}{c}{PhysioNet 2020*}\\
\midrule
SimCLR &0.667 $\pm$ 0.019 & 0.585 $\pm$ 0.013 \\
CMSC &\textbf{0.899 $\pm$ 0.003} & \textbf{0.744 $\pm$ 0.011} \\
CMLC &0.728 $\pm$ 0.021 & 0.627 $\pm$ 0.037 \\
CMSMLC &0.838 $\pm$ 0.015 & 0.644 $\pm$ 0.026 \\
\bottomrule \end{tabular}
\end{subtable}

\vskip 0.1in \begin{subtable}{\textwidth}
\centering
\caption{$F=0.5$}
\label{table:linear_32_0.5}
\begin{tabular}{c | c c c | c c c }
\toprule

Dataset&\multicolumn{1}{c}{Chapman*}&\multicolumn{1}{c}{PhysioNet 2020*}\\
\midrule
SimCLR &0.667 $\pm$ 0.021 & 0.585 $\pm$ 0.015 \\
CMSC & \textbf{0.898 $\pm$ 0.004} & \textbf{0.741 $\pm$ 0.011} \\
CMLC &0.729 $\pm$ 0.019 & 0.627 $\pm$ 0.038 \\
CMSMLC &0.838 $\pm$ 0.018 & 0.641 $\pm$ 0.030 \\
\bottomrule \end{tabular}
\end{subtable}

\vskip 0.1in \begin{subtable}{\textwidth}
\centering
\caption{$F=0.75$}
\label{table:linear_32_0.75}
\begin{tabular}{c | c c c | c c c }
\toprule

Dataset&\multicolumn{1}{c}{Chapman*}&\multicolumn{1}{c}{PhysioNet 2020*}\\
\midrule
SimCLR &0.667 $\pm$ 0.020 & 0.589 $\pm$ 0.014 \\
CMSC &\textbf{0.897 $\pm$ 0.004} & \textbf{0.747 $\pm$ 0.012} \\
CMLC &0.730 $\pm$ 0.019 & 0.635 $\pm$ 0.034 \\
CMSMLC &0.843 $\pm$ 0.018 & 0.652 $\pm$ 0.025 \\
\bottomrule \end{tabular}
\end{subtable}

\vskip 0.1in \begin{subtable}{\textwidth}
\centering
\caption{$F=1$}
\label{table:linear_32_1}
\begin{tabular}{c | c c c | c c c }
\toprule

Dataset&\multicolumn{1}{c}{Chapman*}&\multicolumn{1}{c}{PhysioNet 2020*}\\
\midrule
SimCLR &0.667 $\pm$ 0.019 & 0.589 $\pm$ 0.013 \\
CMSC &\textbf{0.897 $\pm$ 0.004} & \textbf{0.745 $\pm$ 0.013} \\
CMLC &0.726 $\pm$ 0.019 & 0.632 $\pm$ 0.038 \\
CMSMLC &0.844 $\pm$ 0.015 & 0.649 $\pm$ 0.027 \\
\bottomrule \end{tabular}
\end{subtable}
\end{table}

\clearpage

\subsubsection{Embedding Dimension, $E=64$}

We find that the conclusions arrived at with $E=32$ are similar to those in this scenario. Namely, CMSC outperforms all remaining methods when evaluating on both Chapman and PhysioNet 2020. Moreover, our other proposed pre-training methods (CMLC and CMSMLC) also outperform the state-of-the-art method, SimCLR. This can be seen by the bold test AUC values in Table~\ref{table:test_auc_linear_L_12_E_64}. 

\begin{table}[!h]
\centering
\caption{Comparison of self-supervised methods when using networks as feature extractors and performing linear evaluation on downstream datasets. Pre-training and evaluating multi-lead datasets* using all 12 leads. Mean and standard deviation are shown across 5 seeds.}
\label{table:test_auc_linear_L_12_E_64}
\vskip 0.1in 
\begin{subtable}{\textwidth}
\centering
\caption{$F=0.25$}
\label{table:linear_64_0.25}
\begin{tabular}{c | c c c | c c c }
\toprule

Dataset&\multicolumn{1}{c}{Chapman*}&\multicolumn{1}{c}{PhysioNet 2020*}\\
\midrule
SimCLR &0.752 $\pm$ 0.045 & 0.611 $\pm$ 0.009 \\
CMSC & \textbf{0.904 $\pm$ 0.005} & \textbf{0.764 $\pm$ 0.022} \\
CMLC &0.734 $\pm$ 0.021 & 0.650 $\pm$ 0.039 \\
CMSMLC &0.852 $\pm$ 0.024 & 0.669 $\pm$ 0.016 \\
\bottomrule \end{tabular}
\end{subtable}

\vskip 0.1in \begin{subtable}{\textwidth}
\centering
\caption{$F=0.5$}
\label{table:linear_64_0.5}
\begin{tabular}{c | c c c | c c c }

\midrule
Dataset&\multicolumn{1}{c}{Chapman*}&\multicolumn{1}{c}{PhysioNet 2020*}\\
\midrule
SimCLR &0.753 $\pm$ 0.046 & 0.613 $\pm$ 0.011 \\
CMSC &\textbf{0.905 $\pm$ 0.005} & \textbf{0.759 $\pm$ 0.024} \\
CMLC &0.735 $\pm$ 0.020 & 0.650 $\pm$ 0.039 \\
CMSMLC &0.851 $\pm$ 0.023 & 0.665 $\pm$ 0.016 \\
\bottomrule \end{tabular}
\end{subtable}

\vskip 0.1in \begin{subtable}{\textwidth}
\centering
\caption{$F=0.75$}
\label{table:linear_64_0.75}
\begin{tabular}{c | c c c | c c c }
\toprule

Dataset&\multicolumn{1}{c}{Chapman*}&\multicolumn{1}{c}{PhysioNet 2020*}\\
\midrule
SimCLR &0.753 $\pm$ 0.046 & 0.617 $\pm$ 0.008 \\
CMSC &\textbf{0.904 $\pm$ 0.005} & \textbf{0.770 $\pm$ 0.019} \\
CMLC &0.735 $\pm$ 0.020 & 0.661 $\pm$ 0.035 \\
CMSMLC &0.854 $\pm$ 0.018 & 0.675 $\pm$ 0.016 \\
\bottomrule \end{tabular}
\end{subtable}

\vskip 0.1in \begin{subtable}{\textwidth}
\centering
\caption{$F=1$}
\label{table:linear_64_1}
\begin{tabular}{c | c c c | c c c }
\toprule

Dataset&\multicolumn{1}{c}{Chapman*}&\multicolumn{1}{c}{PhysioNet 2020*}\\
\midrule
SimCLR &0.754 $\pm$ 0.045 & 0.616 $\pm$ 0.009 \\
CMSC &\textbf{0.905 $\pm$ 0.005} & \textbf{0.770 $\pm$ 0.019} \\
CMLC &0.735 $\pm$ 0.020 & 0.654 $\pm$ 0.040 \\
CMSMLC &0.853 $\pm$ 0.017 & 0.674 $\pm$ 0.016 \\
\bottomrule \end{tabular}
\end{subtable}
\end{table}

\clearpage

\subsubsection{Embedding Dimension, $E=128$}

In this scenario, the same conclusions as those arrived at with smaller embedding dimensions still hold. Although CMSC continues to outperform all other methods, the performance gap between such methods decreases when compared to results obtained at smaller embedding dimensions. For instance, when evaluating on the Chapman dataset at $F=0.25$ with $E=128$, CMSC achieves an $\mathrm{AUC}=0.903$ whereas SimCLR achieves an $\mathrm{AUC}=0.771$, a performance gap of 13.2\%. In contrast, at $E=64$, the performance gap between these two methods was 15.2\%. 

\begin{table}[!h]
\centering
\caption{Comparison of self-supervised methods when using networks as feature extractors and performing linear evaluation on downstream datasets. Pre-training and evaluating multi-lead datasets* using all 12 leads. Mean and standard deviation are shown across 5 seeds.}
\label{table:test_auc_linear_E_128}
\vskip 0.1in 
\begin{subtable}{\textwidth}
\centering
\caption{$F=0.25$}
\label{table:linear_32_0.25}
\begin{tabular}{c | c c c | c c c }
\toprule
Dataset&\multicolumn{1}{c}{Chapman*}&\multicolumn{1}{c}{PhysioNet 2020*}\\
\midrule
SimCLR &0.771 $\pm$ 0.012 & 0.605 $\pm$ 0.013 \\
CMSC &\textbf{0.903 $\pm$ 0.002} & \textbf{0.7600 $\pm$ 0.019} \\
CMLC &0.779 $\pm$ 0.018 & 0.667 $\pm$ 0.030 \\
CMSMLC &0.846 $\pm$ 0.024 & 0.659 $\pm$ 0.016 \\
\bottomrule \end{tabular}
\end{subtable}

\vskip 0.1in \begin{subtable}{\textwidth}
\centering
\caption{$F=0.5$}
\label{table:linear_32_0.25}
\begin{tabular}{c | c c c | c c c }
\toprule

Dataset&\multicolumn{1}{c}{Chapman*}&\multicolumn{1}{c}{PhysioNet 2020*}\\
\midrule
SimCLR &0.773 $\pm$ 0.012 & 0.606 $\pm$ 0.013 \\
CMSC &\textbf{0.902 $\pm$ 0.003} & \textbf{0.758 $\pm$ 0.019} \\
CMLC &0.783 $\pm$ 0.020 & 0.665 $\pm$ 0.032 \\
CMSMLC &0.850 $\pm$ 0.022 & 0.659 $\pm$ 0.016 \\
\bottomrule \end{tabular}
\end{subtable}

\vskip 0.1in \begin{subtable}{\textwidth}
\centering
\caption{$F=0.75$}
\label{table:linear_32_0.25}
\begin{tabular}{c | c c c | c c c }
\toprule

Dataset&\multicolumn{1}{c}{Chapman*}&\multicolumn{1}{c}{PhysioNet 2020*}\\
\midrule
SimCLR &0.774 $\pm$ 0.012 & 0.611 $\pm$ 0.012 \\
CMSC &\textbf{0.902 $\pm$ 0.003} & \textbf{0.763 $\pm$ 0.019} \\
CMLC &0.788 $\pm$ 0.018 & 0.671 $\pm$ 0.032 \\
CMSMLC &0.851 $\pm$ 0.019 & 0.669 $\pm$ 0.013 \\
\bottomrule \end{tabular}
\end{subtable}

\vskip 0.1in \begin{subtable}{\textwidth}
\centering
\caption{$F=1$}
\label{table:linear_32_0.25}
\begin{tabular}{c | c c c | c c c }
\toprule

Dataset&\multicolumn{1}{c}{Chapman*}&\multicolumn{1}{c}{PhysioNet 2020*}\\
\midrule
SimCLR &0.775 $\pm$ 0.012 & 0.610 $\pm$ 0.013 \\
CMSC &\textbf{0.902 $\pm$ 0.003} & \textbf{0.761 $\pm$ 0.019} \\
CMLC &0.787 $\pm$ 0.020 & 0.672 $\pm$ 0.028 \\
CMSMLC &0.853 $\pm$ 0.017 & 0.669 $\pm$ 0.013 \\
\bottomrule \end{tabular}
\end{subtable}
\end{table}

\clearpage

\subsubsection{Embedding Dimension, $E=256$}

In this scenario, and similar to findings at lower embedding dimensions, CMSC outperforms all methods on both the Chapman and PhysioNet 2020 datasets. Pre-training and evaluating multi-lead datasets* using all 12 leads. These claims are supported by the bold test AUC values in Table~\ref{table:test_auc_linear_L_12_E_256}. 

\begin{table}[!h]
\centering
\caption{Comparison of self-supervised methods when using networks as feature extractors and performing linear evaluation on downstream datasets. Pre-training and evaluating multi-lead datasets* using all 12 leads. Mean and standard deviation are shown across 5 seeds.}
\label{table:test_auc_linear_L_12_E_256}
\vskip 0.1in 
\begin{subtable}{\textwidth}
\centering
\caption{$F=0.25$}
\label{table:linear_64_0.25}
\begin{tabular}{c | c c c | c c c }
\toprule

Dataset&\multicolumn{1}{c}{Chapman*}&\multicolumn{1}{c}{PhysioNet 2020*}\\
\midrule
SimCLR &0.769 $\pm$ 0.028 & 0.614 $\pm$ 0.007 \\
CMSC &\textbf{0.904 $\pm$ 0.002} & \textbf{0.761 $\pm$ 0.011} \\
CMLC &0.784 $\pm$ 0.013 & 0.672 $\pm$ 0.033 \\
CMSMLC &0.852 $\pm$ 0.013 & 0.672 $\pm$ 0.013 \\
\bottomrule \end{tabular}
\end{subtable}

\vskip 0.1in \begin{subtable}{\textwidth}
\centering
\caption{$F=0.5$}
\label{table:linear_64_0.25}
\begin{tabular}{c | c c c | c c c }
\toprule

Dataset&\multicolumn{1}{c}{Chapman*}&\multicolumn{1}{c}{PhysioNet 2020*}\\
\midrule
SimCLR &0.770 $\pm$ 0.027 & 0.617 $\pm$ 0.008 \\
CMSC &\textbf{0.906 $\pm$ 0.002} & \textbf{0.756 $\pm$ 0.010} \\
CMLC &0.790 $\pm$ 0.016 & 0.668 $\pm$ 0.036 \\
CMSMLC &0.852 $\pm$ 0.010& 0.672 $\pm$ 0.012 \\
\bottomrule \end{tabular}
\end{subtable}

\vskip 0.1in \begin{subtable}{\textwidth}
\centering
\caption{$F=0.75$}
\label{table:linear_64_0.25}
\begin{tabular}{c | c c c | c c c }
\toprule

Dataset&\multicolumn{1}{c}{Chapman*}&\multicolumn{1}{c}{PhysioNet 2020*}\\
\midrule
SimCLR &0.770 $\pm$ 0.026 & 0.619 $\pm$ 0.008 \\
CMSC &\textbf{0.905 $\pm$ 0.003} & \textbf{0.764 $\pm$ 0.011} \\
CMLC &0.793 $\pm$ 0.019 & 0.680 $\pm$ 0.029 \\
CMSMLC &0.854 $\pm$ 0.012 & 0.680 $\pm$ 0.012 \\
\bottomrule \end{tabular}
\end{subtable}

\vskip 0.1in \begin{subtable}{\textwidth}
\centering
\caption{$F=1$}
\label{table:linear_64_0.25}
\begin{tabular}{c | c c c | c c c }
\toprule

Dataset&\multicolumn{1}{c}{Chapman*}&\multicolumn{1}{c}{PhysioNet 2020*}\\
\midrule
SimCLR &0.771 $\pm$ 0.027 & 0.619 $\pm$ 0.008 \\
CMSC &\textbf{0.906 $\pm$ 0.003} & \textbf{0.764 $\pm$ 0.010} \\
CMLC &0.797 $\pm$ 0.016 & 0.677 $\pm$ 0.029 \\
CMSMLC &0.858 $\pm$ 0.011 & 0.679 $\pm$ 0.011 \\
\bottomrule \end{tabular}
\end{subtable}
\end{table}

\clearpage

\section{Transfer Capabilities of Representations}
\label{appendix:fine_tuning}

In this section, we evaluate the utility of self-supervised pre-training in generating a favourable parameter initialization for a downstream task. After pre-training, we transfer the parameters to a downstream task and allow all parameters to be updated. In doing so, we are evaluating the benefit brought about by the inductive bias of self-supervised pre-training. 

We perform these experiments under two scenarios. The first involves pre-training, fine-tuning, and evaluating using 4 leads (II, V2, aVL, aVR) (see Sec.~\ref{sec:finetune_4_leads}). The second involves pre-training, fine-tuning, and evaluating using all 12 leads (see Sec.~\ref{sec:finetune_12_leads}). We chose these two scenarios for several reasons. Firstly, they will help determine whether our findings generalize to domains where a different number of leads is available. For example, expensive hospital equipment may record all 12 leads of an ECG, whereas low-cost wearable sensors may only collect data from a subset of leads. Secondly, we wanted to evaluate whether or not contrastive-learning with more views (leads) would improve generalization performance on the downstream task. Previous studies in computer vision have shown this to be the case.

\clearpage

\subsection{Pre-training, Fine-Tuning, and Evaluating using 4 Leads}
\label{sec:finetune_4_leads}

We present Tables~\ref{table:test_auc_finetuning_E_32} - \ref{table:test_auc_finetuning_E_256} which illustrate the test AUC on downstream datasets after having pre-trained on Chapman or PhysioNet 2020. These are shown for a range of embedding dimensions, $E=(32,64,128,256)$, and available labelled training data, $F=(0.25,0.50,0.75,1.00)$. 

\subsubsection{Embedding Dimension, $E=32$}

In this section, we show that in 18/24 (75\%) of all experiments conducted, our family of contrastive learning methods outperforms the state-of-the-art method, SimCLR. This can be seen by the bold test AUC results in Table~\ref{table:test_auc_finetuning_E_32}. The majority of these positive results can be attributed to CMSC. Such a finding illustrates the robustness of our methods to the pre-training and downstream dataset used for evaluation, especially given the diversity of the tasks at hand. 

\begin{table}[!h]
\centering
\caption{Comparison of self-supervised methods when used as parameter initializations before fine-tuning on downstream datasets. Pre-training, fine-tuning, and evaluating multi-lead datasets* using 4 leads. Mean and standard deviation are shown across 5 seeds.}
\label{table:test_auc_finetuning_E_32}
\vskip 0.1in 
\begin{subtable}{\textwidth}
\centering
\caption{$F=0.25$}
\label{table:linear_64_0.25}
\resizebox{\linewidth}{!}{%
\begin{tabular}{c | c c c | c c c }
\toprule
\multirow{1}{*}{Pre-training Dataset}&\multicolumn{3}{c}{Chapman*}&\multicolumn{3}{c}{PhysioNet 2020*}\\
\midrule
Downstream Dataset& Cardiology & PhysioNet 2017 & PhysioNet 2020* &Cardiology & PhysioNet 2017 & Chapman*\\
\midrule
Random Init. &0.631 $\pm$ 0.006 & 0.738 $\pm$ 0.014 & 0.766 $\pm$ 0.005 & 0.631 $\pm$ 0.006 & 0.738 $\pm$ 0.014 & 0.898 $\pm$ 0.002 \\
 SimCLR &0.649 $\pm$ 0.012 & 0.731 $\pm$ 0.017 & 0.790$\pm$ 0.008 & 0.642 $\pm$ 0.020 & 0.738 $\pm$ 0.009 & 0.907 $\pm$ 0.013 \\
 CMSC &0.661 $\pm$ 0.018 & \textbf{0.770 $\pm$ 0.012} & \textbf{0.801 $\pm$ 0.013} & \textbf{0.658 $\pm$ 0.018} & 0.748 $\pm$ 0.027 & \textbf{0.908 $\pm$ 0.011} \\
 CMLC &0.652 $\pm$ 0.014 & 0.767 $\pm$ 0.012 & 0.768 $\pm$ 0.004 & 0.635 $\pm$ 0.017 & \textbf{0.753 $\pm$ 0.013} & 0.906 $\pm$ 0.009 \\
 CMSMLC &\textbf{0.669 $\pm$ 0.020} & 0.758 $\pm$ 0.008 & 0.761 $\pm$ 0.015 & 0.652 $\pm$ 0.013 & 0.733 $\pm$ 0.004 & 0.900 $\pm$ 0.009 \\
\bottomrule \end{tabular}}
\end{subtable}

\vskip 0.1in \begin{subtable}{\textwidth}
\centering
\caption{$F=0.5$}
\label{table:linear_64_0.25}
\resizebox{\linewidth}{!}{%
\begin{tabular}{c | c c c | c c c }
\toprule

\multirow{1}{*}{Pre-training Dataset}&\multicolumn{3}{c}{Chapman*}&\multicolumn{3}{c}{PhysioNet 2020*}\\
\midrule
Downstream Dataset& Cardiology & PhysioNet 2017 & PhysioNet 2020* &Cardiology & PhysioNet 2017 & Chapman*\\
\midrule
Random Init. &0.669 $\pm$ 0.007 & 0.782 $\pm$ 0.011 & 0.811 $\pm$ 0.011 & 0.669 $\pm$ 0.007 & 0.782 $\pm$ 0.011 & 0.907 $\pm$ 0.011 \\
SimCLR &0.691 $\pm$ 0.008 & 0.748 $\pm$ 0.018 & \textbf{0.829 $\pm$ 0.003} & 0.679 $\pm$ 0.012 & 0.767 $\pm$ 0.012 & \textbf{0.933 $\pm$ 0.010} \\
CMSC &0.687 $\pm$ 0.018 & \textbf{0.771 $\pm$ 0.030} & 0.822 $\pm$ 0.011 & \textbf{0.689 $\pm$ 0.025} & \textbf{0.769 $\pm$ 0.010} & 0.926 $\pm$ 0.010 \\
CMLC &0.680 $\pm$ 0.003 & 0.772 $\pm$ 0.007 & 0.812 $\pm$ 0.013 & 0.677 $\pm$ 0.013 & 0.764 $\pm$ 0.027 & 0.918 $\pm$ 0.008 \\
CMSMLC &\textbf{0.708 $\pm$ 0.017} & 0.769 $\pm$ 0.015 & 0.799 $\pm$ 0.011 & 0.684 $\pm$ 0.011 & 0.761 $\pm$ 0.022 & 0.923 $\pm$ 0.012 \\
\bottomrule \end{tabular}}
\end{subtable}

\vskip 0.1in \begin{subtable}{\textwidth}
\centering
\caption{$F=0.75$}
\label{table:linear_64_0.25}
\resizebox{\linewidth}{!}{%
\begin{tabular}{c | c c c | c c c }
\toprule

\multirow{1}{*}{Pre-training Dataset}&\multicolumn{3}{c}{Chapman*}&\multicolumn{3}{c}{PhysioNet 2020*}\\
\midrule
Downstream Dataset& Cardiology & PhysioNet 2017 & PhysioNet 2020* &Cardiology & PhysioNet 2017 & Chapman*\\
\midrule
Random Init. &0.682 $\pm$ 0.016 & 0.764 $\pm$ 0.011 & 0.824 $\pm$ 0.013 & 0.682 $\pm$ 0.016 & 0.764 $\pm$ 0.011 & 0.925 $\pm$ 0.009 \\
SimCLR &0.699 $\pm$ 0.010 & 0.782 $\pm$ 0.015 & \textbf{0.839 $\pm$ 0.003} & 0.691 $\pm$ 0.009 & \textbf{0.795 $\pm$ 0.017} & 0.938 $\pm$ 0.012 \\
CMSC &0.712 $\pm$ 0.011 & 0.760 $\pm$ 0.032 & 0.835 $\pm$ 0.006 & \textbf{0.704 $\pm$ 0.024} & 0.780 $\pm$ 0.015 & \textbf{0.941 $\pm$ 0.006} \\
CMLC &0.682 $\pm$ 0.011 & 0.769 $\pm$ 0.020 & 0.826 $\pm$ 0.014 & 0.665 $\pm$ 0.016 & 0.764 $\pm$ 0.020& 0.930 $\pm$ 0.013 \\
CMSMLC &\textbf{0.715 $\pm$ 0.009} & \textbf{0.789 $\pm$ 0.014} & 0.820 $\pm$ 0.004 & 0.703 $\pm$ 0.010 & 0.778 $\pm$ 0.019 & 0.936 $\pm$ 0.007 \\
\bottomrule \end{tabular}}
\end{subtable}

\vskip 0.1in \begin{subtable}{\textwidth}
\centering
\caption{$F=1$}
\label{table:linear_64_0.25}
\resizebox{\linewidth}{!}{%
\begin{tabular}{c | c c c | c c c }
\toprule

\multirow{1}{*}{Pre-training Dataset}&\multicolumn{3}{c}{Chapman*}&\multicolumn{3}{c}{PhysioNet 2020*}\\
\midrule
Downstream Dataset& Cardiology & PhysioNet 2017 & PhysioNet 2020* &Cardiology & PhysioNet 2017 & Chapman*\\
\midrule
Random Init. &0.700 $\pm$ 0.019 & 0.771 $\pm$ 0.018 & 0.832 $\pm$ 0.006 & 0.700 $\pm$ 0.019 & 0.771 $\pm$ 0.018 & 0.937 $\pm$ 0.005 \\
SimCLR &0.715 $\pm$ 0.005 & \textbf{0.804 $\pm$ 0.020} & \textbf{0.844 $\pm$ 0.001} & 0.704 $\pm$ 0.009 & 0.785 $\pm$ 0.025 & 0.938 $\pm$ 0.011 \\
CMSC &\textbf{0.723 $\pm$ 0.004} & 0.803 $\pm$ 0.033 & 0.841 $\pm$ 0.007 & \textbf{0.724 $\pm$ 0.015} & \textbf{0.795 $\pm$ 0.009} & \textbf{0.945 $\pm$ 0.004} \\
CMLC &0.699 $\pm$ 0.025 & 0.778 $\pm$ 0.015 & 0.837 $\pm$ 0.006 & 0.707 $\pm$ 0.014 & 0.780 $\pm$ 0.023 & 0.933 $\pm$ 0.014 \\
CMSMLC &0.719 $\pm$ 0.016 & 0.798 $\pm$ 0.018 & 0.830 $\pm$ 0.006 & 0.715 $\pm$ 0.014 & 0.775 $\pm$ 0.009 & 0.940 $\pm$ 0.006 \\
\bottomrule \end{tabular}}
\end{subtable}
\end{table}

\clearpage

\subsubsection{Embedding Dimension, $E=64$}

In this section, we show that in 20/24 (83\%) of all experiments conducted, our family of contrastive learning methods outperforms the state-of-the-art method, SimCLR. This can be seen by the bold test AUC results in Table~\ref{table:test_auc_finetuning_E_64}. 

\begin{table}[!h]
\centering
\caption{Comparison of self-supervised methods when used as parameter initializations before fine-tuning on downstream datasets. Pre-training, fine-tuning, and evaluating multi-lead datasets* using 4 leads. Mean and standard deviation are shown across 5 seeds.}
\label{table:test_auc_finetuning_E_64}
\vskip 0.1in 
\begin{subtable}{\textwidth}
\centering
\caption{$F=0.25$}
\label{table:linear_64_0.25}
\resizebox{\linewidth}{!}{%
\begin{tabular}{c | c c c | c c c }
\toprule

\multirow{1}{*}{Pre-training Dataset}&\multicolumn{3}{c}{Chapman*}&\multicolumn{3}{c}{PhysioNet 2020*}\\
\midrule
Downstream Dataset& Cardiology & PhysioNet 2017 & PhysioNet 2020* &Cardiology & PhysioNet 2017 & Chapman*\\
\midrule
Random Init. &0.632 $\pm$ 0.018 & 0.746 $\pm$ 0.009 & 0.775 $\pm$ 0.010& 0.632 $\pm$ 0.018 & 0.746 $\pm$ 0.009 & 0.895 $\pm$ 0.001 \\
SimCLR &0.652 $\pm$ 0.010 & 0.744 $\pm$ 0.013 & 0.784 $\pm$ 0.022 & 0.641 $\pm$ 0.006 & 0.739 $\pm$ 0.019 & \textbf{0.911 $\pm$ 0.007} \\
CMSC &0.663 $\pm$ 0.019 & \textbf{0.765 $\pm$ 0.023} & \textbf{0.794 $\pm$ 0.022} & 0.659 $\pm$ 0.032 & 0.755 $\pm$ 0.018 & 0.910 $\pm$ 0.007 \\
CMLC &0.650 $\pm$ 0.007 & 0.753 $\pm$ 0.012 & 0.786 $\pm$ 0.008 & 0.644 $\pm$ 0.014 & \textbf{0.762 $\pm$ 0.009} & 0.904 $\pm$ 0.007 \\
CMSMLC &\textbf{0.675 $\pm$ 0.010} & 0.755 $\pm$ 0.009 & 0.767 $\pm$ 0.008 & \textbf{0.660 $\pm$ 0.012} & 0.743 $\pm$ 0.016 & 0.901 $\pm$ 0.003 \\
\bottomrule \end{tabular}}
\end{subtable}

\vskip 0.1in \begin{subtable}{\textwidth}
\centering
\caption{$F=0.5$}
\label{table:linear_64_0.25}
\resizebox{\linewidth}{!}{%
\begin{tabular}{c | c c c | c c c }
\toprule

\multirow{1}{*}{Pre-training Dataset}&\multicolumn{3}{c}{Chapman*}&\multicolumn{3}{c}{PhysioNet 2020*}\\
\midrule
Downstream Dataset& Cardiology & PhysioNet 2017 & PhysioNet 2020* &Cardiology & PhysioNet 2017 & Chapman*\\
\midrule
Random Init. &0.685 $\pm$ 0.004 & 0.768 $\pm$ 0.010& 0.817 $\pm$ 0.009 & 0.685 $\pm$ 0.004 & 0.768 $\pm$ 0.010& 0.906 $\pm$ 0.003 \\
SimCLR &0.676 $\pm$ 0.019 & 0.778 $\pm$ 0.008 & 0.822 $\pm$ 0.011 & 0.678 $\pm$ 0.009 & 0.771 $\pm$ 0.018 & 0.927 $\pm$ 0.009 \\
CMSC &0.695 $\pm$ 0.011 & \textbf{0.786 $\pm$ 0.017} & 0.816 $\pm$ 0.016 & \textbf{0.701 $\pm$ 0.023} & 0.772 $\pm$ 0.009 & 0.928 $\pm$ 0.004 \\
CMLC &0.679 $\pm$ 0.016 & \textbf{0.775 $\pm$ 0.010} & \textbf{0.824 $\pm$ 0.004} & 0.677 $\pm$ 0.021 & \textbf{0.775 $\pm$ 0.010} & 0.918 $\pm$ 0.014 \\
CMSMLC &\textbf{0.717 $\pm$ 0.005} & 0.773 $\pm$ 0.011 & 0.808 $\pm$ 0.009 & 0.699 $\pm$ 0.011 & 0.769 $\pm$ 0.009 & \textbf{0.934 $\pm$ 0.004} \\
\bottomrule \end{tabular}}
\end{subtable}

\vskip 0.1in \begin{subtable}{\textwidth}
\centering
\caption{$F=0.75$}
\label{table:linear_64_0.25}
\resizebox{\linewidth}{!}{%
\begin{tabular}{c | c c c | c c c }
\toprule

\multirow{1}{*}{Pre-training Dataset}&\multicolumn{3}{c}{Chapman*}&\multicolumn{3}{c}{PhysioNet 2020*}\\
\midrule
Downstream Dataset& Cardiology & PhysioNet 2017 & PhysioNet 2020* &Cardiology & PhysioNet 2017 & Chapman*\\
\midrule
Random Init. &0.680 $\pm$ 0.015 & 0.7600 $\pm$ 0.011 & 0.830 $\pm$ 0.007 & 0.680 $\pm$ 0.015 & 0.7600 $\pm$ 0.011 & 0.916 $\pm$ 0.011 \\
SimCLR &0.698 $\pm$ 0.008 & 0.790 $\pm$ 0.010& 0.832 $\pm$ 0.007 & 0.689 $\pm$ 0.016 & \textbf{0.783 $\pm$ 0.015} & 0.934 $\pm$ 0.006 \\
CMSC &0.708 $\pm$ 0.006 & 0.790 $\pm$ 0.027 & 0.834 $\pm$ 0.004 & \textbf{0.715 $\pm$ 0.008} & 0.779 $\pm$ 0.013 & \textbf{0.940 $\pm$ 0.007} \\
CMLC &0.688 $\pm$ 0.016 & 0.777 $\pm$ 0.017 & \textbf{0.837 $\pm$ 0.003} & 0.678 $\pm$ 0.019 & 0.777 $\pm$ 0.011 & 0.926 $\pm$ 0.016 \\
CMSMLC &\textbf{0.720 $\pm$ 0.004} & \textbf{0.795 $\pm$ 0.008} & 0.829 $\pm$ 0.009 & 0.704 $\pm$ 0.007 & 0.775 $\pm$ 0.013 & 0.932 $\pm$ 0.005 \\
\bottomrule \end{tabular}}
\end{subtable}

\vskip 0.1in \begin{subtable}{\textwidth}
\centering
\caption{$F=1$}
\label{table:linear_64_0.25}
\resizebox{\linewidth}{!}{%
\begin{tabular}{c | c c c | c c c }
\toprule

\multirow{1}{*}{Pre-training Dataset}&\multicolumn{3}{c}{Chapman*}&\multicolumn{3}{c}{PhysioNet 2020*}\\
\midrule
Downstream Dataset& Cardiology & PhysioNet 2017 & PhysioNet 2020* &Cardiology & PhysioNet 2017 & Chapman*\\
\midrule
Random Init. &0.692 $\pm$ 0.023 & 0.778 $\pm$ 0.017 & 0.840 $\pm$ 0.004 & 0.692 $\pm$ 0.023 & 0.778 $\pm$ 0.017 & 0.932 $\pm$ 0.008 \\
SimCLR &0.718 $\pm$ 0.010 & 0.797 $\pm$ 0.014 & 0.837 $\pm$ 0.009 & 0.706 $\pm$ 0.008 & 0.789 $\pm$ 0.023 & \textbf{0.944 $\pm$ 0.004} \\
CMSC &0.708 $\pm$ 0.022 & 0.800 $\pm$ 0.020 & \textbf{0.842 $\pm$ 0.003} & \textbf{0.726 $\pm$ 0.007} & \textbf{0.797 $\pm$ 0.009} & 0.941 $\pm$ 0.005 \\
CMLC &0.691 $\pm$ 0.007 & 0.780 $\pm$ 0.013 & 0.841 $\pm$ 0.003 & 0.696 $\pm$ 0.022 & 0.786 $\pm$ 0.016 & 0.931 $\pm$ 0.018 \\
CMSMLC &\textbf{0.732 $\pm$ 0.003} & \textbf{0.810 $\pm$ 0.012} & 0.837 $\pm$ 0.010 & 0.722 $\pm$ 0.010& 0.792 $\pm$ 0.006 & 0.940 $\pm$ 0.007 \\
\bottomrule \end{tabular}}
\end{subtable}
\end{table}

\clearpage

\subsubsection{Embedding Dimension, $E=128$}

In this section, we show that in 21/24 (88\%) of all experiments conducted, our family of contrastive learning methods outperforms the state-of-the-art method, SimCLR. This can be seen by the bold test AUC results in Table~\ref{table:test_auc_finetuning_E_128}.

\begin{table}[!h]
\centering
\caption{Comparison of self-supervised methods when used as parameter initializations before fine-tuning on downstream datasets. Pre-training, fine-tuning, and evaluating multi-lead datasets* using 4 leads. Mean and standard deviation are shown across 5 seeds.}
\label{table:test_auc_finetuning_E_128}
\vskip 0.1in 
\begin{subtable}{\textwidth}
\centering
\caption{$F=0.25$}
\label{table:linear_64_0.25}
\resizebox{\linewidth}{!}{%
\begin{tabular}{c | c c c | c c c }
\toprule

\multirow{1}{*}{Pre-training Dataset}&\multicolumn{3}{c}{Chapman*}&\multicolumn{3}{c}{PhysioNet 2020*}\\
\midrule
Downstream Dataset& Cardiology & PhysioNet 2017 & PhysioNet 2020* &Cardiology & PhysioNet 2017 & Chapman*\\
\midrule
Random Init. &0.625 $\pm$ 0.015 & 0.746 $\pm$ 0.006 & 0.764 $\pm$ 0.016 & 0.625 $\pm$ 0.015 & 0.746 $\pm$ 0.006 & 0.894 $\pm$ 0.002 \\
Supervised &0.671 $\pm$ 0.009 & 0.786 $\pm$ 0.012 & 0.804 $\pm$ 0.005 & 0.679 $\pm$ 0.011 & 0.805 $\pm$ 0.005 & 0.942 $\pm$ 0.011 \\
\midrule
\multicolumn{7}{l}{\textit{Self-supervised Pre-training}} \\
\midrule
BYOL &0.620 $\pm$ 0.013 & 0.726 $\pm$ 0.013 & 0.764 $\pm$ 0.013 & 0.624 $\pm$ 0.021 & 0.752 $\pm$ 0.011 & 0.904 $\pm$ 0.006 \\
SimCLR &0.634 $\pm$ 0.014 & 0.738 $\pm$ 0.006 & 0.777 $\pm$ 0.015 & 0.631 $\pm$ 0.022 & 0.727 $\pm$ 0.014 & 0.903 $\pm$ 0.007 \\
CMSC &\textbf{0.691 $\pm$ 0.015} & \textbf{0.768 $\pm$ 0.005} & \textbf{0.813 $\pm$ 0.007} & \textbf{0.671 $\pm$ 0.018} & \textbf{0.756 $\pm$ 0.009} & \textbf{0.911 $\pm$ 0.016} \\
CMLC &0.639 $\pm$ 0.010 & 0.745 $\pm$ 0.012 & 0.770 $\pm$ 0.006 & 0.641 $\pm$ 0.014 & 0.746 $\pm$ 0.014 & 0.897 $\pm$ 0.003 \\
CMSMLC &0.671 $\pm$ 0.016 & 0.755 $\pm$ 0.011 & 0.781 $\pm$ 0.012 & 0.668 $\pm$ 0.011 & 0.751 $\pm$ 0.007 & 0.903 $\pm$ 0.009 \\
\bottomrule \end{tabular}}
\end{subtable}

\vskip 0.1in \begin{subtable}{\textwidth}
\centering
\caption{$F=0.5$}
\label{table:linear_64_0.25}
\resizebox{\linewidth}{!}{%
\begin{tabular}{c | c c c | c c c }
\toprule

\multirow{1}{*}{Pre-training Dataset}&\multicolumn{3}{c}{Chapman*}&\multicolumn{3}{c}{PhysioNet 2020*}\\
\midrule
Downstream Dataset& Cardiology & PhysioNet 2017 & PhysioNet 2020* &Cardiology & PhysioNet 2017 & Chapman*\\
\midrule
Random Init. &0.678 $\pm$ 0.011 & 0.763 $\pm$ 0.005 & 0.803 $\pm$ 0.008 & 0.678 $\pm$ 0.011 & 0.763 $\pm$ 0.005 & 0.907 $\pm$ 0.006 \\
Supervised &0.684 $\pm$ 0.015 & 0.799 $\pm$ 0.008 & 0.827 $\pm$ 0.001 & 0.730 $\pm$ 0.002 & 0.810 $\pm$ 0.009 & 0.954 $\pm$ 0.003 \\
\midrule
\multicolumn{7}{l}{\textit{Self-supervised Pre-training}} \\
\midrule
BYOL &0.678 $\pm$ 0.021 & 0.748 $\pm$ 0.014 & 0.802 $\pm$ 0.013 & 0.674 $\pm$ 0.022 & 0.757 $\pm$ 0.01 & 0.916 $\pm$ 0.009 \\
SimCLR &0.676 $\pm$ 0.011 & 0.772 $\pm$ 0.010& 0.823 $\pm$ 0.011 & 0.658 $\pm$ 0.027 & 0.762 $\pm$ 0.009 & 0.923 $\pm$ 0.010\\
CMSC &0.695 $\pm$ 0.024 & 0.773 $\pm$ 0.013 & \textbf{0.830 $\pm$ 0.002} & \textbf{0.714 $\pm$ 0.014} & 0.760 $\pm$ 0.013 & \textbf{0.932 $\pm$ 0.008} \\
CMLC &0.665 $\pm$ 0.016 & 0.767 $\pm$ 0.013 & 0.810 $\pm$ 0.011 & 0.675 $\pm$ 0.013 & 0.762 $\pm$ 0.007 & 0.910 $\pm$ 0.012 \\
CMSMLC &\textbf{0.717 $\pm$ 0.006} & \textbf{0.774 $\pm$ 0.004} & 0.814 $\pm$ 0.009 & 0.698 $\pm$ 0.011 & \textbf{0.774 $\pm$ 0.012} & 0.930 $\pm$ 0.012 \\
\bottomrule \end{tabular}}
\end{subtable}

\vskip 0.1in \begin{subtable}{\textwidth}
\centering
\caption{$F=0.75$}
\label{table:linear_64_0.25}
\resizebox{\linewidth}{!}{%
\begin{tabular}{c | c c c | c c c }
\toprule

\multirow{1}{*}{Pre-training Dataset}&\multicolumn{3}{c}{Chapman*}&\multicolumn{3}{c}{PhysioNet 2020*}\\
\midrule
Downstream Dataset& Cardiology & PhysioNet 2017 & PhysioNet 2020* &Cardiology & PhysioNet 2017 & Chapman*\\
\midrule
Random Init. &0.675 $\pm$ 0.020& 0.775 $\pm$ 0.005 & 0.831 $\pm$ 0.011 & 0.675 $\pm$ 0.020& 0.775 $\pm$ 0.005 & 0.937 $\pm$ 0.008 \\
Supervised &0.712 $\pm$ 0.017 & 0.799 $\pm$ 0.014 & 0.837 $\pm$ 0.005 & 0.731 $\pm$ 0.007 & 0.815 $\pm$ 0.007 & 0.958 $\pm$ 0.004 \\
\midrule
\multicolumn{7}{l}{\textit{Self-supervised Pre-training}} \\
\midrule
BYOL &0.671 $\pm$ 0.022 & 0.754 $\pm$ 0.009 & 0.825 $\pm$ 0.009 & 0.700 $\pm$ 0.02 & 0.751 $\pm$ 0.033 & 0.930 $\pm$ 0.005 \\
SimCLR &0.694 $\pm$ 0.019 & 0.776 $\pm$ 0.013 & 0.834 $\pm$ 0.009 & 0.686 $\pm$ 0.019 & \textbf{0.785 $\pm$ 0.011} & 0.931 $\pm$ 0.013 \\
CMSC &0.700 $\pm$ 0.012 & \textbf{0.801 $\pm$ 0.013} & \textbf{0.840 $\pm$ 0.004} & 0.707 $\pm$ 0.015 & 0.777 $\pm$ 0.016 & \textbf{0.942 $\pm$ 0.012} \\
CMLC &0.670 $\pm$ 0.019 & 0.771 $\pm$ 0.010 & 0.831 $\pm$ 0.004 & 0.682 $\pm$ 0.005 & 0.772 $\pm$ 0.009 & 0.917 $\pm$ 0.011 \\
CMSMLC &\textbf{0.719 $\pm$ 0.011} & 0.792 $\pm$ 0.014 & 0.837 $\pm$ 0.008 & \textbf{0.711 $\pm$ 0.011} & 0.777 $\pm$ 0.017 & 0.938 $\pm$ 0.010\\
\bottomrule \end{tabular}}
\end{subtable}

\vskip 0.1in \begin{subtable}{\textwidth}
\centering
\caption{$F=1$}
\label{table:linear_64_0.25}
\resizebox{\linewidth}{!}{%
\begin{tabular}{c | c c c | c c c }
\toprule

\multirow{1}{*}{Pre-training Dataset}&\multicolumn{3}{c}{Chapman*}&\multicolumn{3}{c}{PhysioNet 2020*}\\
\midrule
Downstream Dataset& Cardiology & PhysioNet 2017 & PhysioNet 2020* &Cardiology & PhysioNet 2017 & Chapman*\\
\midrule
Random Init. &0.702 $\pm$ 0.016 & 0.773 $\pm$ 0.010 & 0.843 $\pm$ 0.002 & 0.702 $\pm$ 0.016 & 0.773 $\pm$ 0.010 & 0.930 $\pm$ 0.013 \\
Supervised &0.712 $\pm$ 0.017 & 0.799 $\pm$ 0.011 & 0.844 $\pm$ 0.003 & 0.732 $\pm$ 0.008 & 0.821 $\pm$ 0.006 & 0.961 $\pm$ 0.004 \\
\midrule
\multicolumn{7}{l}{\textit{Self-supervised Pre-training}} \\
\midrule
BYOL &0.697 $\pm$ 0.006 & 0.774 $\pm$ 0.017 & 0.834 $\pm$ 0.011 & 0.709 $\pm$ 0.017 & 0.771 $\pm$ 0.022 & 0.935 $\pm$ 0.008 \\
SimCLR &0.705 $\pm$ 0.008 & \textbf{0.810 $\pm$ 0.016} & 0.844 $\pm$ 0.005 & 0.700 $\pm$ 0.012 & \textbf{0.795 $\pm$ 0.021} & 0.941 $\pm$ 0.006 \\
CMSC &0.715 $\pm$ 0.018 & 0.804 $\pm$ 0.018 & \textbf{0.846 $\pm$ 0.002} & \textbf{0.725 $\pm$ 0.020} & 0.779 $\pm$ 0.024 & 0.942 $\pm$ 0.009 \\
CMLC &0.698 $\pm$ 0.007 & 0.781 $\pm$ 0.014 & 0.836 $\pm$ 0.003 & 0.681 $\pm$ 0.005 & 0.785 $\pm$ 0.011 & 0.933 $\pm$ 0.014 \\
CMSMLC &\textbf{0.732 $\pm$ 0.003} & 0.793 $\pm$ 0.012 & 0.844 $\pm$ 0.005 & 0.716 $\pm$ 0.010 & 0.778 $\pm$ 0.025 & \textbf{0.945 $\pm$ 0.005} \\
\bottomrule \end{tabular}}
\end{subtable}
\end{table}

\clearpage

\subsubsection{Embedding Dimension, $E=256$}

In this section, we show that in 16/24 (66\%) of all experiments conducted, our family of contrastive learning methods outperforms the state-of-the-art method, SimCLR. This can be seen by the bold test AUC results in Table~\ref{table:test_auc_finetuning_E_256}.

\begin{table}[!h]
\centering
\caption{Comparison of self-supervised methods when used as parameter initializations before fine-tuning on downstream datasets. Pre-training, fine-tuning, and evaluating multi-lead datasets* using 4 leads. Mean and standard deviation are shown across 5 seeds.}
\label{table:test_auc_finetuning_E_256}
\vskip 0.1in 
\begin{subtable}{\textwidth}
\centering
\caption{$F=0.25$}
\label{table:linear_64_0.25}
\resizebox{\linewidth}{!}{%
\begin{tabular}{c | c c c | c c c }
\toprule

\multirow{1}{*}{Pre-training Dataset}&\multicolumn{3}{c}{Chapman*}&\multicolumn{3}{c}{PhysioNet 2020*}\\
\midrule
Downstream Dataset& Cardiology & PhysioNet 2017 & PhysioNet 2020* &Cardiology & PhysioNet 2017 & Chapman*\\
\midrule
Random Init. &0.630 $\pm$ 0.014 & 0.737 $\pm$ 0.008 & 0.765 $\pm$ 0.004 & 0.630 $\pm$ 0.014 & 0.737 $\pm$ 0.008 & 0.896 $\pm$ 0.002 \\
SimCLR &0.647 $\pm$ 0.014 & 0.727 $\pm$ 0.007 & \textbf{0.791 $\pm$ 0.014} & 0.636 $\pm$ 0.009 & 0.736 $\pm$ 0.008 & 0.902 $\pm$ 0.006 \\
CMSC &0.656 $\pm$ 0.031 & \textbf{0.756 $\pm$ 0.011} & 0.789 $\pm$ 0.019 & \textbf{0.682 $\pm$ 0.024} & 0.750 $\pm$ 0.014 & \textbf{0.905 $\pm$ 0.009} \\
CMLC &0.649 $\pm$ 0.012 & 0.743 $\pm$ 0.005 & 0.784 $\pm$ 0.009 & 0.645 $\pm$ 0.017 & 0.741 $\pm$ 0.008 & 0.898 $\pm$ 0.004 \\
CMSMLC &\textbf{0.686 $\pm$ 0.008} & 0.752 $\pm$ 0.010 & 0.768 $\pm$ 0.017 & 0.652 $\pm$ 0.023 & \textbf{0.758 $\pm$ 0.014} & 0.896 $\pm$ 0.002 \\
\bottomrule \end{tabular}}
\end{subtable}

\vskip 0.1in \begin{subtable}{\textwidth}
\centering
\caption{$F=0.5$}
\label{table:linear_64_0.25}
\resizebox{\linewidth}{!}{%
\begin{tabular}{c | c c c | c c c }
\toprule

\multirow{1}{*}{Pre-training Dataset}&\multicolumn{3}{c}{Chapman*}&\multicolumn{3}{c}{PhysioNet 2020*}\\
\midrule
Downstream Dataset& Cardiology & PhysioNet 2017 & PhysioNet 2020* &Cardiology & PhysioNet 2017 & Chapman*\\
\midrule
Random Init. &0.659 $\pm$ 0.012 & 0.758 $\pm$ 0.021 & 0.817 $\pm$ 0.008 & 0.659 $\pm$ 0.012 & 0.758 $\pm$ 0.021 & 0.901 $\pm$ 0.003 \\
SimCLR &0.667 $\pm$ 0.019 & 0.758 $\pm$ 0.002 & 0.825 $\pm$ 0.014 & 0.659 $\pm$ 0.010 & \textbf{0.769 $\pm$ 0.017} & \textbf{0.924 $\pm$ 0.012} \\
CMSC &0.667 $\pm$ 0.030 & 0.765 $\pm$ 0.003 & 0.819 $\pm$ 0.002 & \textbf{0.709 $\pm$ 0.028} & 0.762 $\pm$ 0.015 & 0.914 $\pm$ 0.011 \\
CMLC &0.679 $\pm$ 0.014 & 0.768 $\pm$ 0.006 & \textbf{0.826 $\pm$ 0.005} & 0.669 $\pm$ 0.027 & 0.768 $\pm$ 0.012 & 0.906 $\pm$ 0.007 \\
CMSMLC &\textbf{0.702 $\pm$ 0.017} & \textbf{0.776 $\pm$ 0.011} & 0.812 $\pm$ 0.014 & 0.694 $\pm$ 0.011 & 0.762 $\pm$ 0.009 & 0.917 $\pm$ 0.011 \\
\bottomrule \end{tabular}}
\end{subtable}

\vskip 0.1in \begin{subtable}{\textwidth}
\centering
\caption{$F=0.75$}
\label{table:linear_64_0.25}
\resizebox{\linewidth}{!}{%
\begin{tabular}{c | c c c | c c c }
\toprule

\multirow{1}{*}{Pre-training Dataset}&\multicolumn{3}{c}{Chapman*}&\multicolumn{3}{c}{PhysioNet 2020*}\\
\midrule
Downstream Dataset& Cardiology & PhysioNet 2017 & PhysioNet 2020* &Cardiology & PhysioNet 2017 & Chapman*\\
\midrule
Random Init. &0.680 $\pm$ 0.018 & 0.764 $\pm$ 0.006 & 0.834 $\pm$ 0.004 & 0.680$\pm$ 0.018 & 0.764 $\pm$ 0.006 & 0.916 $\pm$ 0.015 \\
SimCLR &0.677 $\pm$ 0.016 & \textbf{0.790 $\pm$ 0.015} & 0.834 $\pm$ 0.011 & 0.684 $\pm$ 0.008 & \textbf{0.787 $\pm$ 0.015} & 0.933 $\pm$ 0.013 \\
CMSC &0.698 $\pm$ 0.015 & 0.784 $\pm$ 0.015 & 0.827 $\pm$ 0.014 & \textbf{0.717 $\pm$ 0.010} & 0.780 $\pm$ 0.018 & \textbf{0.935 $\pm$ 0.007} \\
CMLC &0.677 $\pm$ 0.023 & 0.773 $\pm$ 0.006 & \textbf{0.841 $\pm$ 0.001} & 0.681 $\pm$ 0.012 & 0.779 $\pm$ 0.012 & 0.917 $\pm$ 0.012 \\
CMSMLC &\textbf{0.715 $\pm$ 0.008} & 0.785 $\pm$ 0.004 & 0.827 $\pm$ 0.015 & 0.697 $\pm$ 0.016 & 0.784 $\pm$ 0.010& 0.930 $\pm$ 0.004 \\
\bottomrule \end{tabular}}
\end{subtable}

\vskip 0.1in \begin{subtable}{\textwidth}
\centering
\caption{$F=1$}
\label{table:linear_64_0.25}
\resizebox{\linewidth}{!}{%
\begin{tabular}{c | c c c | c c c }

\toprule
\multirow{1}{*}{Pre-training Dataset}&\multicolumn{3}{c}{Chapman*}&\multicolumn{3}{c}{PhysioNet 2020*}\\
\midrule
Downstream Dataset& Cardiology & PhysioNet 2017 & PhysioNet 2020* &Cardiology & PhysioNet 2017 & Chapman*\\
\midrule
Random Init. &0.696 $\pm$ 0.015 & 0.763 $\pm$ 0.012 & 0.842 $\pm$ 0.005 & 0.696 $\pm$ 0.015 & 0.763 $\pm$ 0.012 & 0.918 $\pm$ 0.015 \\
SimCLR &0.711 $\pm$ 0.008 & \textbf{0.798 $\pm$ 0.014} & 0.841 $\pm$ 0.006 & 0.703 $\pm$ 0.007 & \textbf{0.806 $\pm$ 0.012} & 0.943 $\pm$ 0.004 \\
CMSC &0.704 $\pm$ 0.023 & 0.794 $\pm$ 0.018 & 0.840 $\pm$ 0.007 & \textbf{0.718 $\pm$ 0.012} & 0.792 $\pm$ 0.016 & \textbf{0.944 $\pm$ 0.007} \\
CMLC &0.705 $\pm$ 0.009 & 0.781 $\pm$ 0.005 & \textbf{0.844 $\pm$ 0.002} & 0.690 $\pm$ 0.020& 0.779 $\pm$ 0.007 & 0.926 $\pm$ 0.015 \\
CMSMLC &\textbf{0.731 $\pm$ 0.007} & 0.789 $\pm$ 0.020 & 0.839 $\pm$ 0.007 & 0.709 $\pm$ 0.013 & 0.791 $\pm$ 0.007 & 0.943 $\pm$ 0.004 \\
\bottomrule \end{tabular}}
\end{subtable}
\end{table}

\clearpage

\subsection{Pre-training, Fine-tuning, and Evaluating using 12 Leads}
\label{sec:finetune_12_leads}

We present Tables~\ref{table:test_auc_finetuning_L_12_E_32} - \ref{table:test_auc_finetuning_L_12_E_256} which illustrate the test AUC on downstream datasets after having pre-trained on Chapman or PhysioNet 2020 using all 12 leads. These are shown for a range of embedding dimensions, $E=(32,64,128,256)$, and available labelled training data, $F=(0.25,0.50,0.75,1.00)$. Overall, we find that encouraging the representations of a large and diverse set of leads to be similar to one another might be detrimental. This is shown in the subsequent sections by the consistently poorer performance ($\downarrow \mathrm{AUC}$) of CMLC and CMSMLC relative to CMSC where the latter method does not enforce the aforementioned similarity. 

\subsubsection{Embedding Dimension, $E=32$}

In this section, we show that in 22/24 (92\%) of all experiments conducted, CMSC outperforms the state-of-the-art method, SimCLR. This can be seen by the bold test AUC results in Table~\ref{table:test_auc_finetuning_L_12_E_32}. Such a finding illustrates the robustness of our methods to the pre-training and downstream dataset used for evaluation, especially given the diversity of the tasks at hand. 

The performance gap between CMSC and SimCLR widens as the fraction of available labelled training data decreases. For instance, when evaluating on the Cardiology dataset, as $F = 1 \xrightarrow{} 0.25$, CMSC's $\mathrm{AUC} = 0.723 \xrightarrow{} 0.689$ whereas SimCLR's $\mathrm{AUC} = 0.694 \xrightarrow{} 0.636$. Therefore, the performance gap widens by almost a factor of 2 from $2.9\%$ to $5.3\%$. This suggests that CMSC is better equipped to deal with downstream tasks that lack a sufficient amount of labelled data. 

\begin{table}[!h]
\centering
\caption{Comparison of self-supervised methods when used as parameter initializations before fine-tuning on downstream datasets. Pre-training, fine-tuning, and evaluating multi-lead datasets* using all 12 leads. Mean and standard deviation are shown across 5 seeds.}
\label{table:test_auc_finetuning_L_12_E_32}
\vskip 0.1in 
\begin{subtable}{\textwidth}
\centering
\caption{$F=0.25$}
\label{table:linear_64_0.25}
\resizebox{\linewidth}{!}{%
\begin{tabular}{c | c c c | c c c }
\toprule
\multirow{1}{*}{Pre-training Dataset}&\multicolumn{3}{c}{Chapman*}&\multicolumn{3}{c}{PhysioNet 2020*}\\
\midrule
Downstream Dataset& Cardiology & PhysioNet 2017 & PhysioNet 2020* &Cardiology & PhysioNet 2017 & Chapman*\\
\midrule
Random Init. &0.631 $\pm$ 0.006 & 0.738 $\pm$ 0.014 & 0.823 $\pm$ 0.007 & 0.631 $\pm$ 0.006 & 0.738 $\pm$ 0.014 & 0.907 $\pm$ 0.006 \\
SimCLR &0.636 $\pm$ 0.019 & 0.724 $\pm$ 0.016 & 0.826 $\pm$ 0.011 & 0.616 $\pm$ 0.011 & 0.727 $\pm$ 0.020& 0.921 $\pm$ 0.011 \\
CMSC &\textbf{0.689 $\pm$ 0.017} & \textbf{0.782 $\pm$ 0.005} & \textbf{0.833 $\pm$ 0.002} & \textbf{0.681 $\pm$ 0.017} & \textbf{0.769 $\pm$ 0.015} & \textbf{0.936 $\pm$ 0.011} \\
CMLC &0.639 $\pm$ 0.023 & 0.744 $\pm$ 0.018 & 0.827 $\pm$ 0.003 & 0.630 $\pm$ 0.022 & 0.744 $\pm$ 0.022 & 0.912 $\pm$ 0.007 \\
CMSMLC &0.644 $\pm$ 0.026 & 0.740 $\pm$ 0.019 & 0.818 $\pm$ 0.015 & 0.647 $\pm$ 0.022 & 0.745 $\pm$ 0.015 & 0.920 $\pm$ 0.011 \\
\bottomrule \end{tabular}}
\end{subtable}

\vskip 0.1in \begin{subtable}{\textwidth}
\centering
\caption{$F=0.5$}
\label{table:linear_64_0.25}
\resizebox{\linewidth}{!}{%
\begin{tabular}{c | c c c | c c c }
\toprule

\multirow{1}{*}{Pre-training Dataset}&\multicolumn{3}{c}{Chapman*}&\multicolumn{3}{c}{PhysioNet 2020*}\\
\midrule
Downstream Dataset& Cardiology & PhysioNet 2017 & PhysioNet 2020* &Cardiology & PhysioNet 2017 & Chapman*\\
\midrule
Random Init. &0.669 $\pm$ 0.007 & 0.782 $\pm$ 0.011 & 0.814 $\pm$ 0.009 & 0.669 $\pm$ 0.007 & 0.782 $\pm$ 0.011 & 0.938 $\pm$ 0.009 \\
SimCLR &0.659 $\pm$ 0.011 & 0.764 $\pm$ 0.003 & 0.820 $\pm$ 0.032 & 0.669 $\pm$ 0.023 & 0.766 $\pm$ 0.015 & 0.936 $\pm$ 0.014 \\
CMSC &0.686 $\pm$ 0.024 & \textbf{0.800 $\pm$ 0.013} & \textbf{0.836 $\pm$ 0.004} & \textbf{0.719 $\pm$ 0.014} & \textbf{0.778 $\pm$ 0.019} & \textbf{0.951 $\pm$ 0.003} \\
CMLC &0.674 $\pm$ 0.012 & 0.773 $\pm$ 0.018 & 0.831 $\pm$ 0.002 & 0.667 $\pm$ 0.011 & 0.758 $\pm$ 0.017 & 0.933 $\pm$ 0.008 \\
CMSMLC &\textbf{0.691 $\pm$ 0.007} & 0.759 $\pm$ 0.020& 0.831 $\pm$ 0.009 & 0.684 $\pm$ 0.028 & 0.763 $\pm$ 0.024 & 0.942 $\pm$ 0.005 \\
\bottomrule \end{tabular}}
\end{subtable}

\vskip 0.1in \begin{subtable}{\textwidth}
\centering
\caption{$F=0.75$}
\label{table:linear_64_0.25}
\resizebox{\linewidth}{!}{%
\begin{tabular}{c | c c c | c c c }
\toprule

\multirow{1}{*}{Pre-training Dataset}&\multicolumn{3}{c}{Chapman*}&\multicolumn{3}{c}{PhysioNet 2020*}\\
\midrule
Downstream Dataset& Cardiology & PhysioNet 2017 & PhysioNet 2020* &Cardiology & PhysioNet 2017 & Chapman*\\
\midrule
Random Init. &0.682 $\pm$ 0.016 & 0.764 $\pm$ 0.011 & 0.845 $\pm$ 0.001 & 0.682 $\pm$ 0.016 & 0.764 $\pm$ 0.011 & 0.937 $\pm$ 0.016 \\
SimCLR &0.690 $\pm$ 0.023 & 0.786 $\pm$ 0.023 & 0.840 $\pm$ 0.006 & 0.668 $\pm$ 0.013 & 0.782 $\pm$ 0.007 & 0.945 $\pm$ 0.009 \\
CMSC &0.702 $\pm$ 0.013 & \textbf{0.809 $\pm$ 0.009} & \textbf{0.847 $\pm$ 0.001} & \textbf{0.709 $\pm$ 0.010} & \textbf{0.806 $\pm$ 0.005} & \textbf{0.952 $\pm$ 0.010} \\
CMLC &0.684 $\pm$ 0.027 & 0.774 $\pm$ 0.019 & 0.841 $\pm$ 0.015 & 0.680 $\pm$ 0.021 & 0.783 $\pm$ 0.018 & 0.933 $\pm$ 0.014 \\
CMSMLC &\textbf{0.719 $\pm$ 0.007} & 0.757 $\pm$ 0.025 & 0.843 $\pm$ 0.005 & 0.708 $\pm$ 0.011 & 0.787 $\pm$ 0.015 & 0.944 $\pm$ 0.006 \\
\bottomrule \end{tabular}}
\end{subtable}

\vskip 0.1in \begin{subtable}{\textwidth}
\centering
\caption{$F=1$}
\label{table:linear_64_0.25}
\resizebox{\linewidth}{!}{%
\begin{tabular}{c | c c c | c c c }
\toprule

\multirow{1}{*}{Pre-training Dataset}&\multicolumn{3}{c}{Chapman*}&\multicolumn{3}{c}{PhysioNet 2020*}\\
\midrule
Downstream Dataset& Cardiology & PhysioNet 2017 & PhysioNet 2020* &Cardiology & PhysioNet 2017 & Chapman*\\
\midrule
Random Init. &0.700 $\pm$ 0.019 & 0.771 $\pm$ 0.018 & 0.825 $\pm$ 0.016 & 0.700 $\pm$ 0.019 & 0.771 $\pm$ 0.018 & 0.945 $\pm$ 0.003 \\
SimCLR &0.694 $\pm$ 0.010& 0.790 $\pm$ 0.022 & 0.839 $\pm$ 0.008 & 0.691 $\pm$ 0.009 & 0.790 $\pm$ 0.020& 0.942 $\pm$ 0.014 \\
CMSC &\textbf{0.723 $\pm$ 0.011} & \textbf{0.821 $\pm$ 0.013} & \textbf{0.845 $\pm$ 0.003} & \textbf{0.725 $\pm$ 0.017} & \textbf{0.798 $\pm$ 0.008} & \textbf{0.954 $\pm$ 0.007} \\
CMLC &0.702 $\pm$ 0.011 & 0.762 $\pm$ 0.014 & 0.844 $\pm$ 0.003 & 0.708 $\pm$ 0.026 & 0.777 $\pm$ 0.019 & 0.948 $\pm$ 0.005 \\
CMSMLC &0.722 $\pm$ 0.007 & 0.782 $\pm$ 0.013 & 0.845 $\pm$ 0.005 & 0.710 $\pm$ 0.020& 0.768 $\pm$ 0.033 & 0.946 $\pm$ 0.005 \\
\bottomrule \end{tabular}}
\end{subtable}
\end{table}

\clearpage

\subsubsection{Embedding Dimension, $E=64$}

In this section, we show that in 21/24 (88\%) of all experiments conducted, CMSC outperforms the state-of-the-art method, SimCLR. This can be seen by the bold test AUC results in Table~\ref{table:test_auc_finetuning_L_12_E_64}.

\begin{table}[!h]
\centering
\caption{Comparison of self-supervised methods when used as parameter initializations before fine-tuning on downstream datasets. Pre-training, fine-tuning, and evaluating multi-lead datasets* using all 12 leads. Mean and standard deviation are shown across 5 seeds.}
\label{table:test_auc_finetuning_L_12_E_64}
\vskip 0.1in 
\begin{subtable}{\textwidth}
\centering
\caption{$F=0.25$}
\label{table:linear_64_0.25}
\resizebox{\linewidth}{!}{%
\begin{tabular}{c | c c c | c c c }
\toprule
\multirow{1}{*}{Pre-training Dataset}&\multicolumn{3}{c}{Chapman*}&\multicolumn{3}{c}{PhysioNet 2020*}\\
\midrule
Downstream Dataset& Cardiology & PhysioNet 2017 & PhysioNet 2020* &Cardiology & PhysioNet 2017 & Chapman*\\
\midrule
Random Init. &0.632 $\pm$ 0.018 & 0.746 $\pm$ 0.009 & 0.822 $\pm$ 0.011 & 0.632 $\pm$ 0.018 & 0.746 $\pm$ 0.009 & 0.901 $\pm$ 0.004 \\
SimCLR &0.632 $\pm$ 0.021 & 0.736 $\pm$ 0.019 & 0.833 $\pm$ 0.008 & 0.626 $\pm$ 0.008 & 0.734 $\pm$ 0.018 & 0.925 $\pm$ 0.013 \\
CMSC &\textbf{0.681 $\pm$ 0.024} & \textbf{0.798 $\pm$ 0.008} & \textbf{0.834 $\pm$ 0.006} & \textbf{0.658 $\pm$ 0.026} & \textbf{0.779 $\pm$ 0.012} & \textbf{0.942 $\pm$ 0.011} \\
CMLC &0.626 $\pm$ 0.025 & 0.735 $\pm$ 0.011 & 0.825 $\pm$ 0.004 & 0.627 $\pm$ 0.016 & 0.739 $\pm$ 0.014 & 0.910 $\pm$ 0.007 \\
CMSMLC &0.659 $\pm$ 0.024 & 0.738 $\pm$ 0.013 & 0.820 $\pm$ 0.016 & 0.647 $\pm$ 0.023 & 0.743 $\pm$ 0.012 & 0.912 $\pm$ 0.009 \\
\bottomrule \end{tabular}}
\end{subtable}

\vskip 0.1in \begin{subtable}{\textwidth}
\centering
\caption{$F=0.5$}
\label{table:linear_64_0.25}
\resizebox{\linewidth}{!}{%
\begin{tabular}{c | c c c | c c c }
\toprule

\multirow{1}{*}{Pre-training Dataset}&\multicolumn{3}{c}{Chapman*}&\multicolumn{3}{c}{PhysioNet 2020*}\\
\midrule
Downstream Dataset& Cardiology & PhysioNet 2017 & PhysioNet 2020* &Cardiology & PhysioNet 2017 & Chapman*\\
\midrule
Random Init. &0.685 $\pm$ 0.004 & 0.768 $\pm$ 0.010 & 0.831 $\pm$ 0.007 & 0.685 $\pm$ 0.004 & 0.768 $\pm$ 0.01 & 0.931 $\pm$ 0.016 \\
SimCLR &0.672 $\pm$ 0.023 & 0.762 $\pm$ 0.021 & 0.833 $\pm$ 0.011 & 0.681 $\pm$ 0.011 & 0.767 $\pm$ 0.012 & 0.943 $\pm$ 0.006 \\
CMSC & \textbf{0.708 $\pm$ 0.010} & \textbf{0.804 $\pm$ 0.011} & \textbf{0.834 $\pm$ 0.010} & \textbf{0.709 $\pm$ 0.013} & \textbf{0.792 $\pm$ 0.015} & \textbf{0.954 $\pm$ 0.005} \\
CMLC &0.680 $\pm$ 0.017 & 0.763 $\pm$ 0.010& 0.832 $\pm$ 0.005 & 0.694 $\pm$ 0.019 & 0.748 $\pm$ 0.023 & 0.933 $\pm$ 0.009 \\
CMSMLC &0.706 $\pm$ 0.007 & 0.759 $\pm$ 0.014 & 0.815 $\pm$ 0.025 & 0.699 $\pm$ 0.023 & 0.753 $\pm$ 0.017 & 0.940 $\pm$ 0.008 \\
\bottomrule \end{tabular}}
\end{subtable}

\vskip 0.1in \begin{subtable}{\textwidth}
\centering
\caption{$F=0.75$}
\label{table:linear_64_0.25}
\resizebox{\linewidth}{!}{%
\begin{tabular}{c | c c c | c c c }
\toprule

\multirow{1}{*}{Pre-training Dataset}&\multicolumn{3}{c}{Chapman*}&\multicolumn{3}{c}{PhysioNet 2020*}\\
\midrule
Downstream Dataset& Cardiology & PhysioNet 2017 & PhysioNet 2020* &Cardiology & PhysioNet 2017 & Chapman*\\
\midrule
Random Init. &0.68 $\pm$ 0.015 & 0.76 $\pm$ 0.011 & 0.841 $\pm$ 0.008 & 0.680 $\pm$ 0.015 & 0.760 $\pm$ 0.011 & 0.937 $\pm$ 0.009 \\
SimCLR &0.695 $\pm$ 0.023 & 0.779 $\pm$ 0.012 & 0.844 $\pm$ 0.007 & 0.674 $\pm$ 0.017 & 0.775 $\pm$ 0.011 & 0.948 $\pm$ 0.009 \\
CMSC &0.709 $\pm$ 0.014 & \textbf{0.809 $\pm$ 0.014} & 0.844 $\pm$ 0.007 & \textbf{0.714 $\pm$ 0.017} & \textbf{0.802 $\pm$ 0.012} & \textbf{0.953 $\pm$ 0.006} \\
CMLC &0.690$\pm$ 0.007 & 0.778 $\pm$ 0.010& \textbf{0.844 $\pm$ 0.002} & 0.704 $\pm$ 0.021 & 0.768 $\pm$ 0.018 & 0.946 $\pm$ 0.003 \\
CMSMLC &\textbf{0.711 $\pm$ 0.011} & 0.763 $\pm$ 0.016 & 0.838 $\pm$ 0.006 & 0.689 $\pm$ 0.022 & 0.762 $\pm$ 0.019 & 0.946 $\pm$ 0.008 \\
\bottomrule \end{tabular}}
\end{subtable}

\vskip 0.1in \begin{subtable}{\textwidth}
\centering
\caption{$F=1$}
\label{table:linear_64_0.25}
\resizebox{\linewidth}{!}{%
\begin{tabular}{c | c c c | c c c }

\toprule
\multirow{1}{*}{Pre-training Dataset}&\multicolumn{3}{c}{Chapman*}&\multicolumn{3}{c}{PhysioNet 2020*}\\
\midrule
Downstream Dataset& Cardiology & PhysioNet 2017 & PhysioNet 2020* &Cardiology & PhysioNet 2017 & Chapman*\\
\midrule
Random Init. &0.692 $\pm$ 0.023 & 0.778 $\pm$ 0.017 & 0.846 $\pm$ 0.003 & 0.692 $\pm$ 0.023 & 0.778 $\pm$ 0.017 & 0.946 $\pm$ 0.005 \\
SimCLR &0.715 $\pm$ 0.011 & 0.808 $\pm$ 0.009 & 0.842 $\pm$ 0.007 & 0.703 $\pm$ 0.006 & 0.797 $\pm$ 0.018 & 0.952 $\pm$ 0.008 \\
CMSC & \textbf{0.736 $\pm$ 0.016} & \textbf{0.810 $\pm$ 0.005} & 0.843 $\pm$ 0.005 & \textbf{0.731 $\pm$ 0.010} & \textbf{0.810 $\pm$ 0.015} & \textbf{0.958 $\pm$ 0.007} \\
CMLC &0.706 $\pm$ 0.012 & 0.777 $\pm$ 0.017 & \textbf{0.846 $\pm$ 0.002} & 0.709 $\pm$ 0.012 & 0.779 $\pm$ 0.018 & 0.947 $\pm$ 0.005 \\
CMSMLC &0.722 $\pm$ 0.008 & 0.780 $\pm$ 0.015 & 0.842 $\pm$ 0.008 & 0.701 $\pm$ 0.023 & 0.779 $\pm$ 0.015 & 0.943 $\pm$ 0.009 \\
\bottomrule \end{tabular}}
\end{subtable}
\end{table}

\clearpage

\subsubsection{Embedding Dimension, $E=128$}

In this section, we show that in 24/24 (100\%) of all experiments conducted, CMSC outperforms the the state-of-the-art method, SimCLR. This can be seen by the bold test AUC results in Table~\ref{table:test_auc_finetuning_L_12_E_128}. 

\begin{table}[!h]
\centering
\caption{Comparison of self-supervised methods when used as parameter initializations before fine-tuning on downstream datasets. Pre-training, fine-tuning, and evaluating multi-lead datasets* using all 12 leads. Mean and standard deviation are shown across 5 seeds.}
\label{table:test_auc_finetuning_L_12_E_128}
\vskip 0.1in 
\begin{subtable}{\textwidth}
\centering
\caption{$F=0.25$}
\label{table:linear_64_0.25}
\resizebox{\linewidth}{!}{%
\begin{tabular}{c | c c c | c c c }
\toprule

\multirow{1}{*}{Pre-training Dataset}&\multicolumn{3}{c}{Chapman*}&\multicolumn{3}{c}{PhysioNet 2020*}\\
\midrule
Downstream Dataset& Cardiology & PhysioNet 2017 & PhysioNet 2020* &Cardiology & PhysioNet 2017 & Chapman*\\
\midrule
Random Init. &0.625 $\pm$ 0.015 & 0.746 $\pm$ 0.006 & 0.819 $\pm$ 0.008 & 0.625 $\pm$ 0.015 & 0.746 $\pm$ 0.006 & 0.909 $\pm$ 0.006 \\
SimCLR &0.630 $\pm$ 0.011 & 0.735 $\pm$ 0.012 & 0.833 $\pm$ 0.008 & 0.624 $\pm$ 0.007 & 0.729 $\pm$ 0.018 & 0.918 $\pm$ 0.015 \\
CMSC &\textbf{0.678 $\pm$ 0.010} & \textbf{0.790 $\pm$ 0.012} & \textbf{0.833 $\pm$ 0.008} & \textbf{0.680 $\pm$ 0.011} & \textbf{0.777 $\pm$ 0.027} & \textbf{0.940 $\pm$ 0.007} \\
CMLC &0.639 $\pm$ 0.012 & 0.740 $\pm$ 0.007 & 0.831 $\pm$ 0.003 & 0.639 $\pm$ 0.019 & 0.743 $\pm$ 0.016 & 0.913 $\pm$ 0.012 \\
CMSMLC &0.661 $\pm$ 0.029 & 0.748 $\pm$ 0.005 & 0.813 $\pm$ 0.024 & 0.646 $\pm$ 0.023 & 0.736 $\pm$ 0.007 & 0.918 $\pm$ 0.012 \\
\bottomrule \end{tabular}}
\end{subtable}

\vskip 0.1in \begin{subtable}{\textwidth}
\centering
\caption{$F=0.5$}
\label{table:linear_64_0.25}
\resizebox{\linewidth}{!}{%
\begin{tabular}{c | c c c | c c c }
\toprule

\multirow{1}{*}{Pre-training Dataset}&\multicolumn{3}{c}{Chapman*}&\multicolumn{3}{c}{PhysioNet 2020*}\\
\midrule
Downstream Dataset& Cardiology & PhysioNet 2017 & PhysioNet 2020* &Cardiology & PhysioNet 2017 & Chapman*\\
\midrule
Random Init.&0.678 $\pm$ 0.011 & 0.763 $\pm$ 0.005 & 0.832 $\pm$ 0.003 & 0.678 $\pm$ 0.011 & 0.763 $\pm$ 0.005 & 0.931 $\pm$ 0.014 \\
SimCLR &0.667 $\pm$ 0.021 & 0.768 $\pm$ 0.012 & 0.835 $\pm$ 0.010& 0.659 $\pm$ 0.012 & 0.754 $\pm$ 0.024 & 0.939 $\pm$ 0.007 \\
CMSC &\textbf{0.716 $\pm$ 0.010} & \textbf{0.802 $\pm$ 0.007} & \textbf{0.840 $\pm$ 0.003} & \textbf{0.718 $\pm$ 0.005} & \textbf{0.791 $\pm$ 0.025} & \textbf{0.944 $\pm$ 0.008} \\
CMLC &0.690 $\pm$ 0.012 & 0.763 $\pm$ 0.009 & 0.840 $\pm$ 0.003 & 0.663 $\pm$ 0.040 & 0.752 $\pm$ 0.016 & 0.927 $\pm$ 0.013 \\
CMSMLC &0.699 $\pm$ 0.013 & 0.751 $\pm$ 0.013 & 0.815 $\pm$ 0.014 & 0.695 $\pm$ 0.020& 0.748 $\pm$ 0.013 & 0.931 $\pm$ 0.011 \\
\bottomrule \end{tabular}}
\end{subtable}

\vskip 0.1in \begin{subtable}{\textwidth}
\centering
\caption{$F=0.75$}
\label{table:linear_64_0.25}
\resizebox{\linewidth}{!}{%
\begin{tabular}{c | c c c | c c c }
\toprule

\multirow{1}{*}{Pre-training Dataset}&\multicolumn{3}{c}{Chapman*}&\multicolumn{3}{c}{PhysioNet 2020*}\\
\midrule
Downstream Dataset& Cardiology & PhysioNet 2017 & PhysioNet 2020* &Cardiology & PhysioNet 2017 & Chapman*\\
\midrule
Random Init. &0.675 $\pm$ 0.020 & 0.775 $\pm$ 0.005 & 0.844 $\pm$ 0.006 & 0.675 $\pm$ 0.020 & 0.775 $\pm$ 0.005 & 0.945 $\pm$ 0.004 \\
SimCLR &0.682 $\pm$ 0.023 & 0.775 $\pm$ 0.009 & 0.843 $\pm$ 0.007 & 0.681 $\pm$ 0.020& 0.764 $\pm$ 0.019 & 0.946 $\pm$ 0.010\\
CMSC &\textbf{0.719 $\pm$ 0.008} & \textbf{0.813 $\pm$ 0.006} & \textbf{0.847 $\pm$ 0.002} & \textbf{0.711 $\pm$ 0.004} & \textbf{0.810 $\pm$ 0.020} & \textbf{0.955 $\pm$ 0.005} \\
CMLC &0.684 $\pm$ 0.008 & 0.777 $\pm$ 0.021 & 0.846 $\pm$ 0.001 & 0.700 $\pm$ 0.016 & 0.755 $\pm$ 0.016 & 0.942 $\pm$ 0.005 \\
CMSMLC &0.711 $\pm$ 0.011 & 0.782 $\pm$ 0.006 & 0.839 $\pm$ 0.007 & 0.694 $\pm$ 0.028 & 0.769 $\pm$ 0.014 & 0.941 $\pm$ 0.007 \\
\bottomrule \end{tabular}}
\end{subtable}

\vskip 0.1in \begin{subtable}{\textwidth}
\centering
\caption{$F=1$}
\label{table:linear_64_0.25}
\resizebox{\linewidth}{!}{%
\begin{tabular}{c | c c c | c c c }
\toprule

\multirow{1}{*}{Pre-training Dataset}&\multicolumn{3}{c}{Chapman*}&\multicolumn{3}{c}{PhysioNet 2020*}\\
\midrule
Downstream Dataset& Cardiology & PhysioNet 2017 & PhysioNet 2020* &Cardiology & PhysioNet 2017 & Chapman*\\
\midrule
Random Init. &0.702 $\pm$ 0.016 & 0.773 $\pm$ 0.01 & 0.842 $\pm$ 0.008 & 0.702 $\pm$ 0.016 & 0.773 $\pm$ 0.01 & 0.942 $\pm$ 0.006 \\
SimCLR &0.703 $\pm$ 0.020& 0.801 $\pm$ 0.014 & 0.845 $\pm$ 0.009 & 0.703 $\pm$ 0.014 & 0.784 $\pm$ 0.009 & 0.948 $\pm$ 0.008 \\
CMSC & \textbf{0.731 $\pm$ 0.022} & \textbf{0.819 $\pm$ 0.004} & \textbf{0.847 $\pm$ 0.003} & \textbf{0.718 $\pm$ 0.012} & \textbf{0.809 $\pm$ 0.021} & \textbf{0.959 $\pm$ 0.004} \\
CMLC &0.705 $\pm$ 0.010& 0.777 $\pm$ 0.011 & 0.845 $\pm$ 0.002 & 0.713 $\pm$ 0.023 & 0.789 $\pm$ 0.012 & 0.946 $\pm$ 0.005 \\
CMSMLC &0.719 $\pm$ 0.005 & 0.764 $\pm$ 0.010& 0.837 $\pm$ 0.007 & 0.711 $\pm$ 0.013 & 0.779 $\pm$ 0.013 & 0.947 $\pm$ 0.003 \\
\bottomrule \end{tabular}}
\end{subtable}
\end{table}

\clearpage

\subsubsection{Embedding Dimension, $E=256$}

In this section, we show that in 22/24 (92\%) of all experiments conducted, CMSC outperforms the state-of-the-art method, SimCLR. This can be seen by the bold test AUC results in Table~\ref{table:test_auc_finetuning_L_12_E_256}.

\begin{table}[!h]
\centering
\caption{Comparison of self-supervised methods when used as parameter initializations before fine-tuning on downstream datasets. Pre-training, fine-tuning, and evaluating multi-lead datasets* using all 12 leads. Mean and standard deviation are shown across 5 seeds.}
\label{table:test_auc_finetuning_L_12_E_256}
\vskip 0.1in 
\begin{subtable}{\textwidth}
\centering
\caption{$F=0.25$}
\label{table:linear_64_0.25}
\resizebox{\linewidth}{!}{%
\begin{tabular}{c | c c c | c c c }
\toprule

\multirow{1}{*}{Pre-training Dataset}&\multicolumn{3}{c}{Chapman*}&\multicolumn{3}{c}{PhysioNet 2020*}\\
\midrule
Downstream Dataset& Cardiology & PhysioNet 2017 & PhysioNet 2020* &Cardiology & PhysioNet 2017 & Chapman*\\
\midrule
Random Init.&0.630 $\pm$ 0.014 & 0.737 $\pm$ 0.008 & 0.809 $\pm$ 0.023 & 0.630 $\pm$ 0.014 & 0.737 $\pm$ 0.008 & 0.903 $\pm$ 0.005 \\
SimCLR &0.620 $\pm$ 0.028 & 0.729 $\pm$ 0.013 & 0.830 $\pm$ 0.007 & 0.621 $\pm$ 0.016 & 0.726 $\pm$ 0.008 & 0.933 $\pm$ 0.007 \\
CMSC & \textbf{0.692 $\pm$ 0.007} & \textbf{0.792 $\pm$ 0.014} & \textbf{0.832 $\pm$ 0.009} & \textbf{0.689 $\pm$ 0.013} & \textbf{0.782 $\pm$ 0.010} & \textbf{0.940 $\pm$ 0.010} \\
CMLC &0.618 $\pm$ 0.004 & 0.733 $\pm$ 0.006 & 0.831 $\pm$ 0.009 & 0.648 $\pm$ 0.018 & 0.743 $\pm$ 0.010& 0.912 $\pm$ 0.006 \\
CMSMLC &0.666 $\pm$ 0.012 & 0.741 $\pm$ 0.010& 0.820 $\pm$ 0.013 & 0.666 $\pm$ 0.008 & 0.736 $\pm$ 0.012 & 0.922 $\pm$ 0.011 \\
\bottomrule \end{tabular}}
\end{subtable}

\vskip 0.1in \begin{subtable}{\textwidth}
\centering
\caption{$F=0.5$}
\label{table:linear_64_0.25}
\resizebox{\linewidth}{!}{%
\begin{tabular}{c | c c c | c c c }
\toprule

\multirow{1}{*}{Pre-training Dataset}&\multicolumn{3}{c}{Chapman*}&\multicolumn{3}{c}{PhysioNet 2020*}\\
\midrule
Downstream Dataset& Cardiology & PhysioNet 2017 & PhysioNet 2020* &Cardiology & PhysioNet 2017 & Chapman*\\
\midrule
Random Init.&0.659 $\pm$ 0.012 & 0.758 $\pm$ 0.021 & 0.831 $\pm$ 0.011 & 0.659 $\pm$ 0.012 & 0.758 $\pm$ 0.021 & 0.929 $\pm$ 0.010 \\
SimCLR &0.670 $\pm$ 0.021 & 0.764 $\pm$ 0.008 & 0.830 $\pm$ 0.011 & 0.663 $\pm$ 0.007 & 0.762 $\pm$ 0.009 & 0.942 $\pm$ 0.005 \\
CMSC & \textbf{0.706 $\pm$ 0.024} & \textbf{0.809 $\pm$ 0.004} & 0.835 $\pm$ 0.009 & \textbf{0.714 $\pm$ 0.006} & \textbf{0.798 $\pm$ 0.009} & \textbf{0.953 $\pm$ 0.007} \\
CMLC &0.668 $\pm$ 0.006 & 0.762 $\pm$ 0.005 & \textbf{0.837 $\pm$ 0.007} & 0.700 $\pm$ 0.013 & 0.768 $\pm$ 0.011 & 0.935 $\pm$ 0.010\\
CMSMLC &0.704 $\pm$ 0.012 & 0.763 $\pm$ 0.009 & 0.829 $\pm$ 0.009 & 0.713 $\pm$ 0.006 & 0.748 $\pm$ 0.011 & 0.940 $\pm$ 0.003 \\
\bottomrule \end{tabular}}
\end{subtable}

\vskip 0.1in \begin{subtable}{\textwidth}
\centering
\caption{$F=0.75$}
\label{table:linear_64_0.25}
\resizebox{\linewidth}{!}{%
\begin{tabular}{c | c c c | c c c }
\toprule

\multirow{1}{*}{Pre-training Dataset}&\multicolumn{3}{c}{Chapman*}&\multicolumn{3}{c}{PhysioNet 2020*}\\
\midrule
Downstream Dataset& Cardiology & PhysioNet 2017 & PhysioNet 2020* &Cardiology & PhysioNet 2017 & Chapman*\\
\midrule
Random Init.&0.680 $\pm$ 0.018 & 0.764 $\pm$ 0.006 & 0.844 $\pm$ 0.004 & 0.68 $\pm$ 0.018 & 0.764 $\pm$ 0.006 & 0.936 $\pm$ 0.014 \\
SimCLR &0.675 $\pm$ 0.016 & 0.782 $\pm$ 0.014 & 0.842 $\pm$ 0.007 & 0.678 $\pm$ 0.007 & 0.786 $\pm$ 0.010& 0.953 $\pm$ 0.002 \\
CMSC &\textbf{0.714 $\pm$ 0.013} & \textbf{0.816 $\pm$ 0.003} & \textbf{0.843 $\pm$ 0.006} & \textbf{0.722 $\pm$ 0.015} & \textbf{0.805 $\pm$ 0.011} & \textbf{0.958 $\pm$ 0.003} \\
CMLC &0.678 $\pm$ 0.011 & 0.777 $\pm$ 0.009 & 0.841 $\pm$ 0.006 & 0.705 $\pm$ 0.013 & 0.775 $\pm$ 0.013 & 0.936 $\pm$ 0.008 \\
CMSMLC &0.701 $\pm$ 0.014 & 0.775 $\pm$ 0.009 & 0.842 $\pm$ 0.007 & 0.705 $\pm$ 0.004 & 0.763 $\pm$ 0.009 & 0.946 $\pm$ 0.005 \\
\bottomrule \end{tabular}}
\end{subtable}

\vskip 0.1in \begin{subtable}{\textwidth}
\centering
\caption{$F=1$}
\label{table:linear_64_0.25}
\resizebox{\linewidth}{!}{%
\begin{tabular}{c | c c c | c c c }
\toprule

\multirow{1}{*}{Pre-training Dataset}&\multicolumn{3}{c}{Chapman*}&\multicolumn{3}{c}{PhysioNet 2020*}\\
\midrule
Downstream Dataset& Cardiology & PhysioNet 2017 & PhysioNet 2020* &Cardiology & PhysioNet 2017 & Chapman*\\
\midrule
Random Init.&0.696 $\pm$ 0.015 & 0.763 $\pm$ 0.012 & 0.839 $\pm$ 0.009 & 0.696 $\pm$ 0.015 & 0.763 $\pm$ 0.012 & 0.943 $\pm$ 0.002 \\
SimCLR &0.708 $\pm$ 0.019 & 0.789 $\pm$ 0.006 & \textbf{0.844 $\pm$ 0.007} & 0.707 $\pm$ 0.009 & 0.792 $\pm$ 0.008 & 0.951 $\pm$ 0.005 \\
CMSC & \textbf{0.735 $\pm$ 0.006} & \textbf{0.822 $\pm$ 0.004} & 0.843 $\pm$ 0.006 & \textbf{0.729 $\pm$ 0.010} & \textbf{0.807 $\pm$ 0.012} & \textbf{0.957 $\pm$ 0.004} \\
CMLC &0.705 $\pm$ 0.006 & 0.795 $\pm$ 0.014 & 0.843 $\pm$ 0.007 & 0.719 $\pm$ 0.003 & 0.793 $\pm$ 0.016 & 0.942 $\pm$ 0.006 \\
CMSMLC &0.722 $\pm$ 0.008 & 0.778 $\pm$ 0.013 & 0.842 $\pm$ 0.008 & 0.722 $\pm$ 0.003 & 0.767 $\pm$ 0.013 & 0.946 $\pm$ 0.003 \\
\bottomrule \end{tabular}}
\end{subtable}
\end{table}

\clearpage

\section{Effect of $\tau_{d}$ on BYOL Implementation}
\label{appendix:effect_of_tau}

In the BYOL implementation, two networks exist; an online network and a target network. The latter is a delayed version of the former where its parameters are an exponential moving average of the parameters of the online network. This exponential moving average is a function of the hyperparameter, $\tau_{d}$. In this section, we outline the effect of $\tau_{d}$ on the downstream generalization performance of networks both in the linear and transfer evaluation scenarios. This can be found in Figs~\ref{table:effect_of_tau_linear} and \ref{table:effect_of_tau_finetuning}, respectively. We find that the results associated with $\tau_{d}=0.900$ lead to the best performance, and are thus quoted in the main manuscript. 

\subsection{Linear Evaluation of Representations}

\begin{table}[!h]
\centering
\caption{Effect of the value of $\tau_{d}$ during BYOL pre-training on the downstream generalization performance of a linear evaluation scenario. Pre-training, fine-tuning, and evaluating multi-lead datasets* using 4 leads. Mean and standard deviation are shown across 5 seeds.}
\label{table:effect_of_tau_linear}
\vskip 0.1in 
\vskip 0.1in \begin{subtable}{\textwidth}
\centering
\caption{$F=0.25$}
\label{table:linear_64_0.25}
\begin{tabular}{c | c c}
\toprule
Dataset&\multicolumn{1}{c}{Chapman*}&\multicolumn{1}{c}{PhysioNet 2020*}\\
\midrule
$\tau_{d} = 0.500$ & 0.602 $\pm$ 0.072 & 0.581 $\pm$ 0.010 \\
$\tau_{d} = 0.900$ & \textbf{0.671 $\pm$ 0.042} & \textbf{0.587 $\pm$ 0.021} \\ 
$\tau_{d} = 0.990$ & 0.597 $\pm$ 0.068 & 0.571 $\pm$ 0.028 \\
\bottomrule \end{tabular}
\end{subtable}

\vskip 0.1in \begin{subtable}{\textwidth}
\centering
\caption{$F=0.5$}
\label{table:linear_64_0.25}
\begin{tabular}{c | c c c | c c c }
\toprule
Dataset&\multicolumn{1}{c}{Chapman*}&\multicolumn{1}{c}{PhysioNet 2020*}\\
\midrule
$\tau_{d} = 0.500$ & 0.618 $\pm$ 0.087 & 0.590 $\pm$ 0.010 \\
$\tau_{d} = 0.900$ & \textbf{0.643 $\pm$ 0.043} & \textbf{0.595 $\pm$ 0.018} \\
$\tau_{d} = 0.990$ & 0.604 $\pm$ 0.079 & 0.578 $\pm$ 0.033 \\
\bottomrule \end{tabular}
\end{subtable}

\vskip 0.1in \begin{subtable}{\textwidth}
\centering
\caption{$F=0.75$}
\label{table:linear_64_0.25}
\begin{tabular}{c | c c c | c c c }
\toprule

Dataset&\multicolumn{1}{c}{Chapman*}&\multicolumn{1}{c}{PhysioNet 2020*}\\
\midrule
$\tau_{d} = 0.500$ & 0.635 $\pm$ 0.075 & 0.597 $\pm$ 0.008 \\
$\tau_{d} = 0.900$ & \textbf{0.666 $\pm$ 0.032} & \textbf{0.598 $\pm$ 0.022} \\
$\tau_{d} = 0.990$ & 0.613 $\pm$ 0.085 & 0.586 $\pm$ 0.026 \\
\bottomrule \end{tabular}
\end{subtable}

\vskip 0.1in \begin{subtable}{\textwidth}
\centering
\caption{$F=1$}
\label{table:linear_64_0.25}
\begin{tabular}{c | c c c | c c c }
\toprule

Dataset&\multicolumn{1}{c}{Chapman*}&\multicolumn{1}{c}{PhysioNet 2020*}\\
\midrule
$\tau_{d} = 0.500$ & 0.637 $\pm$ 0.082 & 0.601 $\pm$ 0.008 \\
$\tau_{d} = 0.900$ & \textbf{0.653 $\pm$ 0.026} & \textbf{0.602 $\pm$ 0.015} \\ 
$\tau_{d} = 0.990$ & 0.619 $\pm$ 0.088 & 0.592 $\pm$ 0.026 \\
\bottomrule \end{tabular}
\end{subtable}
\end{table}

\clearpage

\subsection{Transfer Capabilities of Representations}

\begin{table}[!h]
\centering
\caption{Effect of the value of $\tau_{d}$ during BYOL pre-training on the downstream generalization performance in the fine-tuning evaluation scenario. Pre-training, fine-tuning, and evaluating multi-lead datasets* using 4 leads. Mean and standard deviation are shown across 5 seeds.}
\label{table:effect_of_tau_finetuning}
\vskip 0.1in 
\vskip 0.1in \begin{subtable}{\textwidth}
\centering
\caption{$F=0.25$}
\label{table:linear_64_0.25}
\resizebox{\linewidth}{!}{%
\begin{tabular}{c | c c c | c c c }
\toprule

\multirow{1}{*}{Pre-training Dataset}&\multicolumn{3}{c}{Chapman*}&\multicolumn{3}{c}{PhysioNet 2020*}\\
\midrule
Downstream Dataset& Cardiology & PhysioNet 2017 & PhysioNet 2020* &Cardiology & PhysioNet 2017 & Chapman*\\
\midrule
$\tau_{d} = 0.500$ &0.614 $\pm$ 0.026 & 0.738 $\pm$ 0.023 & 0.765 $\pm$ 0.015 & 0.609 $\pm$ 0.015 & 0.724 $\pm$ 0.027 & 0.900 $\pm$ 0.003 \\
$\tau_{d} = 0.900$ &0.620 $\pm$ 0.013 & 0.726 $\pm$ 0.013 & 0.764 $\pm$ 0.013 & 0.624 $\pm$ 0.021 & 0.752 $\pm$ 0.011 & 0.904 $\pm$ 0.006 \\
$\tau_{d} = 0.990$ &0.612 $\pm$ 0.009 & 0.732 $\pm$ 0.022 & 0.767 $\pm$ 0.018 & 0.617 $\pm$ 0.022 & 0.729 $\pm$ 0.015 & 0.901 $\pm$ 0.003 \\ 
\bottomrule \end{tabular}}
\end{subtable}

\vskip 0.1in \begin{subtable}{\textwidth}
\centering
\caption{$F=0.5$}
\label{table:linear_64_0.25}
\resizebox{\linewidth}{!}{%
\begin{tabular}{c | c c c | c c c }
\toprule

\multirow{1}{*}{Pre-training Dataset}&\multicolumn{3}{c}{Chapman*}&\multicolumn{3}{c}{PhysioNet 2020*}\\
\midrule
Downstream Dataset& Cardiology & PhysioNet 2017 & PhysioNet 2020* &Cardiology & PhysioNet 2017 & Chapman*\\
\midrule
$\tau_{d} = 0.500$ & 0.685 $\pm$ 0.015 & 0.763 $\pm$ 0.011 & 0.797 $\pm$ 0.019 & 0.658 $\pm$ 0.046 & 0.739 $\pm$ 0.027 & 0.913 $\pm$ 0.009 \\ 
$\tau_{d} = 0.900$ & 0.678 $\pm$ 0.021 & 0.748 $\pm$ 0.014 & 0.802 $\pm$ 0.013 & 0.674 $\pm$ 0.022 & 0.757 $\pm$ 0.010 & 0.916 $\pm$ 0.009 \\
$\tau_{d} = 0.990$ & 0.671 $\pm$ 0.013 & 0.748 $\pm$ 0.014 & 0.802 $\pm$ 0.017 & 0.658 $\pm$ 0.017 & 0.755 $\pm$ 0.021 & 0.910 $\pm$ 0.009 \\
\bottomrule \end{tabular}}
\end{subtable}

\vskip 0.1in \begin{subtable}{\textwidth}
\centering
\caption{$F=0.75$}
\label{table:linear_64_0.25}
\resizebox{\linewidth}{!}{%
\begin{tabular}{c | c c c | c c c }
\toprule

\multirow{1}{*}{Pre-training Dataset}&\multicolumn{3}{c}{Chapman*}&\multicolumn{3}{c}{PhysioNet 2020*}\\
\midrule
Downstream Dataset& Cardiology & PhysioNet 2017 & PhysioNet 2020* &Cardiology & PhysioNet 2017 & Chapman*\\
\midrule
$\tau_{d} = 0.500$ & 0.689 $\pm$ 0.007 & 0.766 $\pm$ 0.014 & 0.824 $\pm$ 0.014 & 0.693 $\pm$ 0.015 & 0.758 $\pm$ 0.030 & 0.919 $\pm$ 0.013 \\
$\tau_{d} = 0.900$ & 0.671 $\pm$ 0.022 & 0.754 $\pm$ 0.009 & 0.825 $\pm$ 0.009 & 0.700 $\pm$ 0.020 & 0.751 $\pm$ 0.033 & 0.930 $\pm$ 0.005 \\
$\tau_{d} = 0.990$ & 0.678 $\pm$ 0.019 & 0.764 $\pm$ 0.009 & 0.822 $\pm$ 0.011 & 0.662 $\pm$ 0.026 & 0.763 $\pm$ 0.01 & 0.925 $\pm$ 0.010 \\
\bottomrule \end{tabular}}
\end{subtable}

\vskip 0.1in \begin{subtable}{\textwidth}
\centering
\caption{$F=1$}
\label{table:linear_64_0.25}
\resizebox{\linewidth}{!}{%
\begin{tabular}{c | c c c | c c c }
\toprule

\multirow{1}{*}{Pre-training Dataset}&\multicolumn{3}{c}{Chapman*}&\multicolumn{3}{c}{PhysioNet 2020*}\\
\midrule
Downstream Dataset& Cardiology & PhysioNet 2017 & PhysioNet 2020* &Cardiology & PhysioNet 2017 & Chapman*\\
\midrule
$\tau_{d} = 0.500$ & 0.709 $\pm$ 0.013 & 0.754 $\pm$ 0.008 & 0.826 $\pm$ 0.015 & 0.691 $\pm$ 0.038 & 0.770 $\pm$ 0.017 & 0.931 $\pm$ 0.007 \\
$\tau_{d} = 0.900$ & 0.697 $\pm$ 0.006 & 0.774 $\pm$ 0.017 & 0.834 $\pm$ 0.011 & 0.709 $\pm$ 0.017 & 0.771 $\pm$ 0.022 & 0.935 $\pm$ 0.008 \\
$\tau_{d} = 0.990$ & 0.701 $\pm$ 0.014 & 0.761 $\pm$ 0.020 & 0.833 $\pm$ 0.008 & 0.679 $\pm$ 0.042 & 0.756 $\pm$ 0.013 & 0.936 $\pm$ 0.011 \\
\bottomrule \end{tabular}}
\end{subtable}
\end{table}

\clearpage

\section{Intra and Inter-Patient Representation Distances}
\label{appendix:patient_distances}

\subsection{Effect of Embedding Dimension, $E$, on Learning Patient-specific Representations}

\begin{figure}[!h]
\centering
\begin{subfigure}{0.45\textwidth}
	\centering
	\includegraphics[width=\textwidth]{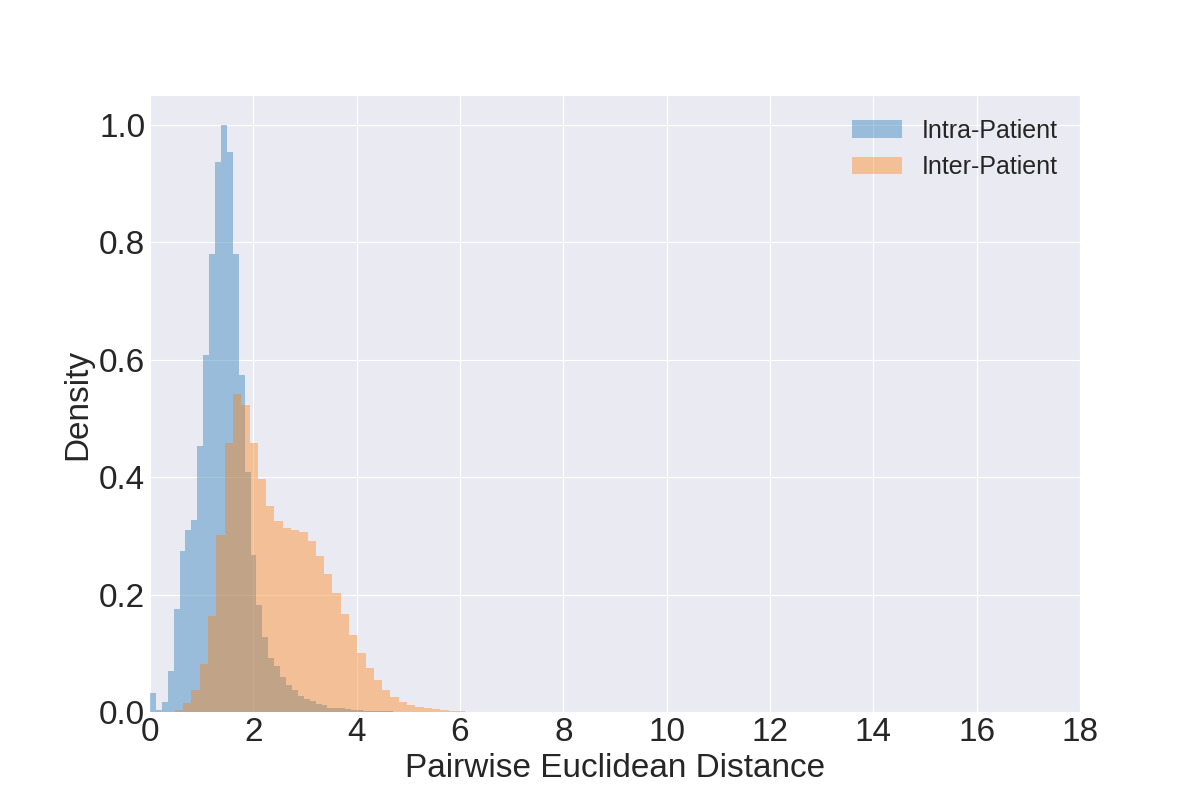}
	\caption{$E=32$}
	\label{fig:patient_distances_32}
\end{subfigure}
\begin{subfigure}{0.45\textwidth}
	\centering
	\includegraphics[width=\textwidth]{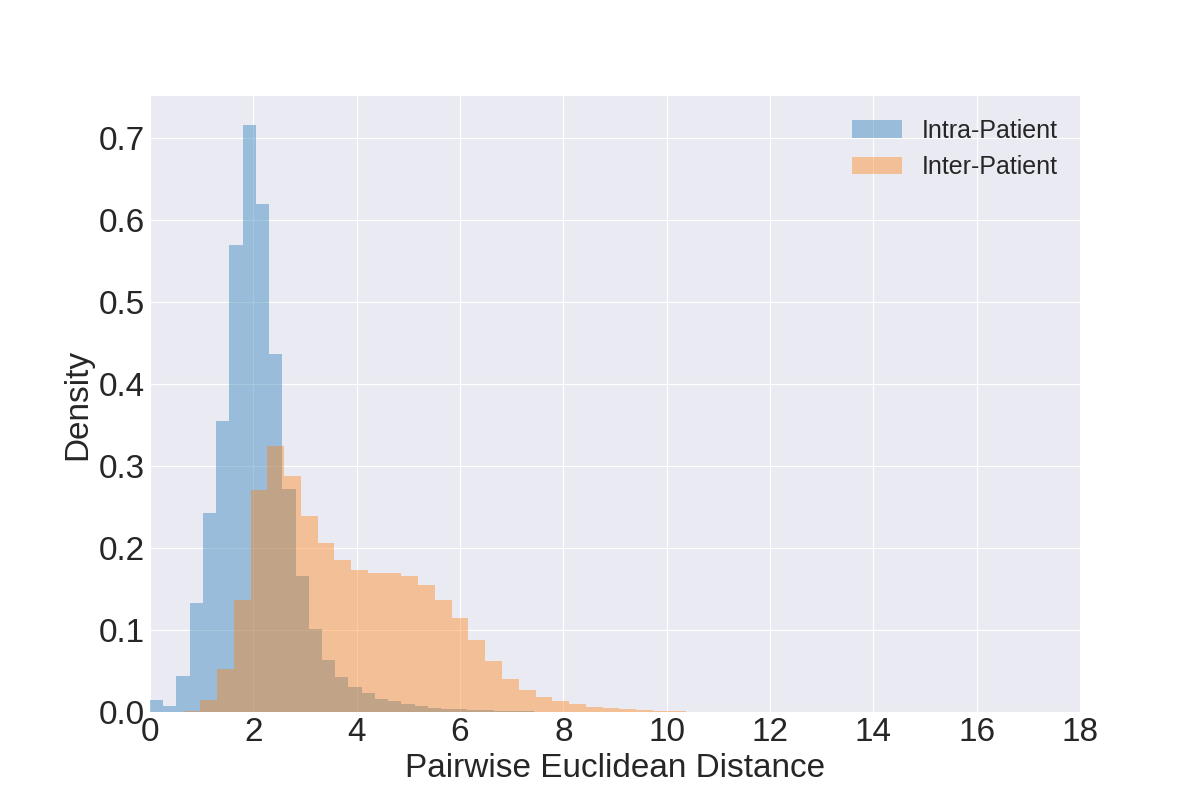}
	\caption{$E=64$}
	\label{fig:patient_distances_64}
\end{subfigure}
\begin{subfigure}{0.45\textwidth}
	\centering
	\includegraphics[width=\textwidth]{intra_vs_inter_patient_distances_E_128.png}
	\caption{$E=128$}
	\label{fig:patient_distances_128}
\end{subfigure}
\begin{subfigure}{0.45\textwidth}
	\centering
	\includegraphics[width=\textwidth]{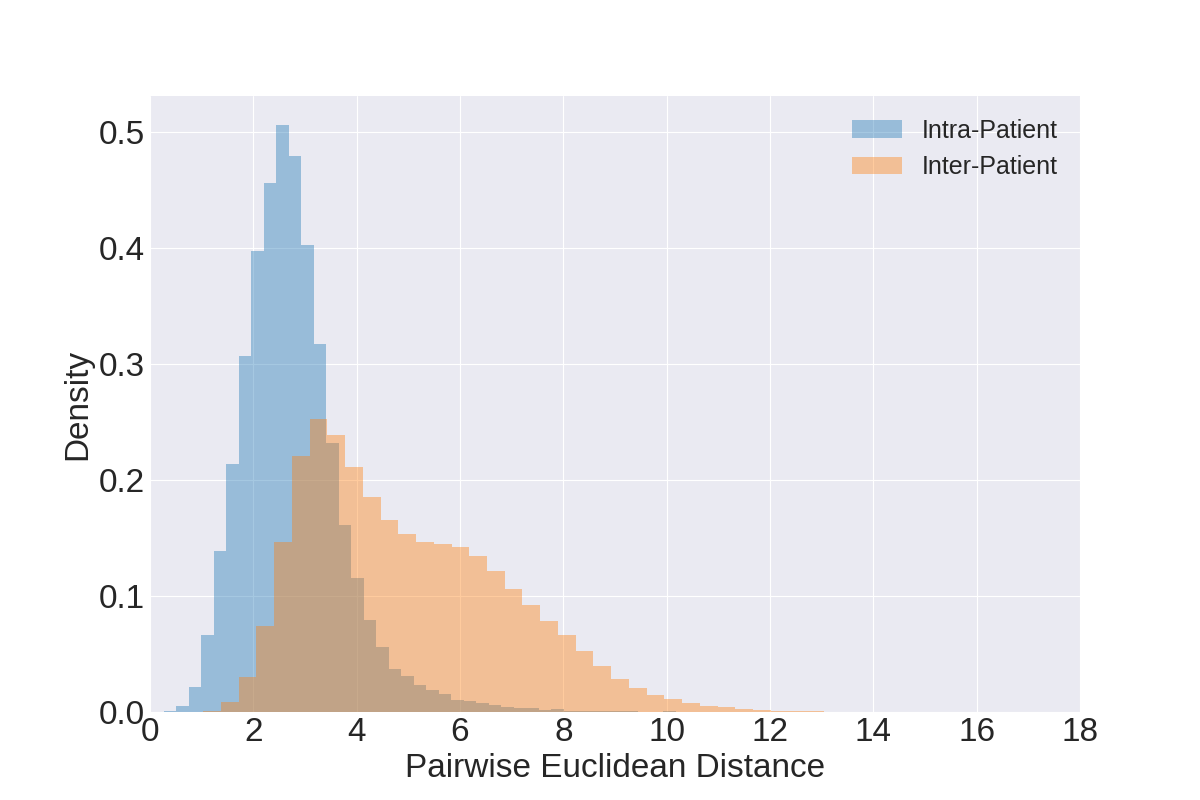}
	\caption{$E=256$}
	\label{fig:patient_distances_256}
\end{subfigure}
	\caption{Distribution of pairwise Euclidean distance between representations belonging to the same patient (Intra-Patient) and those belonging to different patients (Inter-Patient). Representations are of instances present in the validation set of PhysioNet 2020. Self-supervision was performed with CMSC on PhysioNet 2020 using 4 leads.}
	\label{fig:patient_distances}
\end{figure}

In Fig.~\ref{fig:patient_distances}, we show that, when using a low embedding dimension ($E=32$), the intra-patient distances are the lowest with a mean of around 1. As $E=32 \xrightarrow{} 256$, the distributions begin to shift to higher values. Such high pairwise distances imply that maintaining similar representations at higher dimensions is more difficult. Moreover, we clearly see two distinct distributions belonging to intra-patient and inter-patient distances. This suggests that the training procedure worked as expected, leading to representations that are more similar within patients than across patients. 

\end{subappendices}

\end{document}